%% file: neurips_2026.tex
\documentclass{article}

\usepackage[preprint]{neurips_2026}

\usepackage[utf8]{inputenc} 
\usepackage[T1]{fontenc}    
\usepackage{url}            
\usepackage{booktabs}       
\usepackage{amsfonts}       
\usepackage{nicefrac}       
\usepackage{microtype}      
\usepackage{xcolor}         
\usepackage{graphicx}
\usepackage{subfigure}

\usepackage{amsmath}
\usepackage{amssymb}
\usepackage{mathtools}
\usepackage{amsthm}

\usepackage[capitalize,noabbrev]{cleveref}

\theoremstyle{plain}

\theoremstyle{definition}

\theoremstyle{remark}

\usepackage[textsize=tiny]{todonotes}

\title{Distributional Spectral Diagnostics for Localizing Grokking Transitions}

\author{
Ziyue Wang\textsuperscript{1} \quad
Yufeng Ying\textsuperscript{2} \quad
Takafumi Kanamori\textsuperscript{1} \\
{\small
\textsuperscript{1}Institute of Science Tokyo \quad
\textsuperscript{2}University of Science and Technology of China
}
}

\begin{document}

\maketitle

\begin{abstract}
In grokking, a model first fits the training data while test accuracy remains low, and only later begins to generalize. We ask whether this transition can be localized from observed training trajectories before the test accuracy rises, and formulate grokking transition localization as a diagnostic problem with an explicit threshold/FPR/lead-time trade-off. Task-dependent observables are summarized as empirical distributions, mapped to Wasserstein/quantile coordinates, and analyzed by Hankel dynamic mode decomposition (DMD); the resulting reconstruction residual, together with spectrum and effective rank, forms the diagnostic output. On held-out modular-addition Transformer runs, the residual achieves AUROC \(\approx \) 0.93 for grokking-vs-non-grokking discrimination at the run level; under a fixed sustained-threshold operating rule, true-positive alarms can precede onset, with lead time reported jointly with false-alarm rate and uncertainty intervals. Perturbation experiments show that, in the tested \(wd=1\) pool, high-residual windows exhibit about \(3\times\) larger short-horizon perturbation deviation than low-residual windows. In a same-data norm-window control, perturbation sensitivity aligns with the residual ordering rather than total-parameter-norm ordering, suggesting that the residual is not merely a total-norm proxy at the window level in the studied \(wd=1\) dynamics. Norm signals remain strong run-level regime indicators, and log-probability performs best among the observables tested under the current protocol. We position the residual as a window-level monitoring and localization signal in the studied modular-arithmetic Transformer settings, not a universal early-warning predictor or an intervention rule.
\end{abstract}

\input{Chapters/Introduction}

\input{Chapters/distributional}

\input{Chapters/Results}

\input{Chapters/Discussion}

\begin{ack}
\end{ack}

\bibliography{Chapters/ref}
\bibliographystyle{abbrvnat}

\newpage
\appendix
\input{Chapters/Appendix}



\end{document}

%% file: Chapters/Introduction.tex
\section{Introduction}

Grokking is a striking failure mode of standard training summaries: on certain algorithmic tasks, a model fits the training set early yet generalizes only after a long delay, with test-accuracy curves remaining nearly flat throughout the intervening period \citep{power2022grokkinggeneralizationoverfittingsmall}. The flatness is the difficulty---loss and accuracy alone do not directly indicate when the transition will occur, even in runs that eventually grok. 
We address a narrow question motivated by this gap: can transition windows be localized from observed training trajectories, using generic distributional observables and an explicit threshold/FPR/lead-time diagnostic protocol?
This is a different question from explaining \emph{why} grokking occurs.

A growing body of work explains grokking through mechanism. Mechanistic interpretability identifies emergent computational circuits \citep{nanda2023progressmeasuresgrokkingmechanistic,varma2023explaininggrokkingcircuitefficiency}; implicit-bias analyses attribute the transition to late-phase norm minimization on the zero-loss manifold \citep{liu2022understandinggrokkingeffectivetheory,lyu2024dichotomyearlylatephase,musat2026geometrygrokkingnormminimization}; stability-based accounts link grokking to logit scaling and softmax collapse \citep{thilak2022slingshotmechanismempiricalstudy,prieto2025grokkingedgenumericalstability}. Norm-growth, logit-scaling, AGOP-based feature emergence \citep{radhakrishnan2024agop}, and circuit-formation views are the natural reference points for our work. 
Our goal is complementary: a thresholded localization signal computed from task-dependent observables and reported with explicit false-alarm/lead-time trade-offs, intended to flag transition windows rather than explain them.

We summarize a chosen task-dependent observable $o_t$ at each training step as an empirical distribution $\mu_t$. Because the diagnostic is computed from the chosen output distribution rather than from hidden-unit coordinates, it does not depend on hidden-unit indexing. Wasserstein/quantile coordinates convert $\mu_t$ into a vector observation $z_t$, and windowed Hankel dynamic mode decomposition (DMD) provides a local dynamical approximation \citep{SCHMID_2010,H_Tu_2014,Arbabi_2017}. The reconstruction residual $\mathrm{Res}^{(r)}$ is the primary diagnostic; the spectrum and effective rank $r_{0.99}$ are auxiliary descriptors interpretable mainly in low-residual windows. Our primary setting takes the observable from the empirical distribution of correct-answer log-probabilities on a fixed probe set in Transformers trained on modular addition; secondary observables and FCN comparisons appear later as scope checks.

Empirically, the residual rises around grokking transitions in the modular-addition Transformer setting. On a held-out test split with fresh seeds, the residual gives nontrivial run-level detection behavior under an explicit threshold/FPR/lead-time trade-off (precise numbers in \S\ref{subsec:detection} and Table~\ref{tab:detection}). A perturbation experiment shows that high-residual windows exhibit larger short-horizon deviation than low-residual windows under matched noise. A same-data norm-window control re-labels the same runs by total-parameter-norm percentile and produces an opposite fragility ordering, suggesting that the residual is not merely a total-norm proxy at the window level in the studied wd$=1$ dynamics; norm signals nevertheless remain strong run-level regime indicators. 
An observable ablation finds log-probability to be the best-performing observable among those tested under the current protocol.

\paragraph{Contributions.}
\begin{itemize}
    \item[(i)] We propose a windowed distributional diagnostic for training dynamics. The method maps task-dependent observables to empirical distributions, represents them by Wasserstein/quantile coordinates, and applies Hankel-DMD to compute spectrum, effective rank, and reconstruction residual.
    \item[(ii)] We evaluate the reconstruction residual as a transition-localization signal for grokking. Paired with a sustained-threshold operating rule, it gives held-out detection behavior under an explicit threshold/FPR/lead-time trade-off; log-probability is the best-performing observable among those we tested under the current protocol.
    \item[(iii)] We provide perturbation-based evidence that high-residual windows correspond to fragile training periods, supporting the residual as a monitoring/localization signal rather than an intervention rule.
    \item[(iv)] We evaluate scope and boundaries through model-scale, task-family, norm-baseline, AGOP, intervention, CIFAR-10, and FCN checks; these are presented as scope checks, not universal robustness claims.
\end{itemize}

\paragraph{Scope of claims.}
We do not claim a universal predictor of grokking, an architecture-independent diagnostic, or an automatic intervention rule. We do not claim the residual replaces norm-based regime classifiers, nor that perturbation alignment establishes causal mechanism. 
Our claim is narrower: in the studied modular-arithmetic Transformer settings, the reconstruction residual is a window-level cue for transition localization and fragility monitoring, evaluated under an explicit threshold/FPR/lead-time trade-off.
Extended comparisons to mechanistic, spectral, Koopman/DMD, and Wasserstein training-dynamics diagnostics are deferred to Appendix~\ref{app:related}.

%% file: Chapters/distributional.tex
\section{Method}\label{sec:method}

The pipeline has three stages: (i) summarize a chosen task-dependent observable as an empirical distribution; (ii) embed each distribution in a Hilbert coordinate via the Wasserstein--quantile representation; (iii) analyze the resulting vector-valued trajectory over fixed step windows by Hankel-DMD and read off a small set of windowed quantities. The pipeline is observable-dependent: the choice of $o_t$ determines what the diagnostic can detect.
Figure~\ref{fig:pipeline} summarizes the overall
diagnostic pipeline, and Table~\ref{tab:notation} lists the main notation.
\begin{table}[ht]
\centering
\caption{Notation used in the distributional Hankel-DMD diagnostic.}
\label{tab:notation}
\small
\begin{tabular}{ll}
\hline
Symbol & Meaning \\
\hline
\(\theta_t\) & Model parameters at training step \(t\). \\
\(o_t\) & Task-dependent observable selected at step \(t\). \\
\(\mu_t\) & Empirical distribution induced by \(o_t\). \\
\(z_t \in \mathbb{R}^d\) & Fixed-quantile feature vector representing \(\mu_t\). \\
\(\xi_t \in \mathbb{R}^{qd}\) & Hankel delay state \((z_t,\ldots,z_{t+q-1})\). \\
\(A_H^\star\) & Hankel-DMD least-squares estimator on a window. \\
\(\{\lambda_j\}\) & Retained DMD eigenvalues, i.e., the local spectrum. \\
\(r_{0.99}\) & Effective rank explaining 99\% of Hankel snapshot energy. \\
\(\mathrm{Res}^{(r)}\) & Rank-\(r\) reconstruction residual. \\
\(b_{\mathrm{run}}\) & Per-run residual baseline used by the alarm rule. \\
\(\tau, K\) & Threshold multiplier and sustainment length. \\
\hline
\end{tabular}
\end{table}


\subsection{Observable and Wasserstein--quantile coordinates}\label{subsec:wass}


\paragraph{Observable and distributional state.}
For Transformers on modular addition (primary setting), we fix a probe set $\mathcal{P}=\{(x_i,y_i^\star)\}_{i=1}^{M}$ ($M=100$ examples), where $y_i^\star$ is the correct-answer token for input $x_i$. At training step $t$ the per-sample observable is the scalar correct-answer log-probability $o_{t,i}=\log p_{\theta_t}(y_i^\star\mid x_i)$, and the distributional state is the empirical distribution of these $M$ scalars:
\begin{equation}\label{eq:mu-tr}
\mu_t \;:=\; \frac{1}{M}\sum_{i=1}^{M} \delta_{o_{t,i}}.
\end{equation}
The diagnostic therefore tracks the empirical distribution of correct-answer log-probabilities over a fixed probe set; an averaged loss or accuracy collapses this distribution to a single scalar. Because the construction uses the chosen output distribution rather than hidden-unit coordinates, it does not depend on hidden-unit indexing. Wasserstein/quantile coordinates provide the representation: they convert distribution-valued states into vector-valued observations. Hankel-DMD provides the local dynamical approximation: it analyzes how these vectors evolve over short training windows. FCN observables, used as a secondary low-residual descriptor, are defined in Appendix~\ref{app:fcn-secondary}.

\paragraph{Wasserstein--quantile coordinate.}
For one-dimensional measures with finite second moment, the quantile map is a global isometry between $(\mathcal{W}_2(\mathbb{R}),d_W)$ and a closed convex subset of $L^2(0,1)$ \citep{villani2009optimal}: at a fixed reference $\mu^\star$, $\log_{\mu^\star}(\mu) = F_\mu^{-1}\circ F_{\mu^\star} - \mathrm{id}$ identifies the Wasserstein tangent space with a Hilbert subspace. We evaluate $F^{-1}_{\mu_t}$ on a fixed quantile grid $p_1,\dots,p_d$ ($d=19$ levels, $0.05$--$0.95$):
\begin{equation}\label{eq:z-coord}
z_t \;=\; \bigl(F_{\mu_t}^{-1}(p_1),\,\dots,\,F_{\mu_t}^{-1}(p_d)\bigr) \in \mathbb{R}^d.
\end{equation}
Multi-dimensional analogues require embeddings (kernel mean embeddings, MDS) that do not preserve Wasserstein geometry; we therefore restrict to one-dimensional task-dependent observables. Full Wasserstein background, including the Hadamard structure of $\mathcal{W}_2(D)$, is in Appendix~\ref{app:wass-detail}.

\subsection{Windowed Hankel-DMD diagnostics}\label{subsec:hankel-rrr}

\paragraph{Delay state and snapshot matrices.}
Over a step window of length $m+q$ we form delay-embedded vectors
\begin{equation}\label{eq:xi}
\xi_t \;=\; \bigl[z_t^\top,\, z_{t+1}^\top,\, \ldots,\, z_{t+q-1}^\top\bigr]^\top \in \mathbb{R}^{qd},
\end{equation}
and Hankel snapshot matrices
$
H_- = [\xi_0\,\cdots\,\xi_{m-1}],\ H_+ = [\xi_1\,\cdots\,\xi_m]\in\mathbb{R}^{qd\times m}.
$

\paragraph{Hankel-DMD estimator.}
A Koopman/DMD approximation \citep{SCHMID_2010,H_Tu_2014,Arbabi_2017,drmač2017datadrivenmodaldecompositions} solves the ordinary least-squares problem
\begin{equation}\label{eq:dmd-est}
A_H^\star \;=\; \arg\min_{A}\,\bigl\|H_+ - A H_-\bigr\|_F^2,
\end{equation}
and we then truncate the fitted operator to rank $r$ via the leading $r$ eigenpairs $(\lambda_j, w_j)_{j=1}^{r}$ of $A_H^\star$. The rank-$r$ DMD reconstruction is $\widehat{\xi}_t^{(r)} = W\Lambda^t b$, where $b = W^\dagger \xi_0$. Snapshot construction details, the reduced-rank projection, and a discussion of non-normal sensitivity are in Appendix~\ref{app:dmd-detail}.

\paragraph{Reconstruction residual.}
\begin{equation}\label{eq:rr}
\mathrm{Res}^{(r)} \;:=\; \frac{\bigl(\sum_{t}\|\xi_t-\widehat{\xi}_t^{(r)}\|_2^2\bigr)^{1/2}}{\bigl(\sum_{t}\|\xi_t\|_2^2\bigr)^{1/2}}.
\end{equation}
A small $\mathrm{Res}^{(r)}$ indicates the windowed trajectory admits an accurate low-rank linear description in the chosen coordinates; a large value indicates a departure from that regime.

\paragraph{Effective rank.}
With $H_- = U\Sigma V^\top$ and singular values $\sigma_1\ge\sigma_2\ge\cdots$,
\begin{equation}\label{eq:reff}
r_{0.99} \;:=\; \min\Bigl\{r\ge 1:\ \tfrac{\sum_{i=1}^{r}\sigma_i^2}{\sum_i \sigma_i^2}\ge 0.99\Bigr\}.
\end{equation}

\paragraph{Validity gate.}
As summarized in Figure~\ref{fig:pipeline}, the residual serves as a validity gate for the auxiliary descriptors. In low-residual windows, the spectrum of $A_H^\star$ and $r_{0.99}$ can be read as empirical summaries of the local linear-evolution regime. In high-residual windows, the low-rank linear approximation is poor; the residual itself should then be interpreted as a transition or fragility signal, and spectral points and effective rank should not be over-interpreted as stable regime descriptors.

\begin{figure}[t]
\centering
\[
\underbrace{o_t}_{\substack{\text{selected}\\ \text{observable}}}
\;\longrightarrow\;
\underbrace{\mu_t}_{\substack{\text{distributional}\\ \text{state}}}
\;\xrightarrow{\;\text{fixed quantile map}\;}\;
\underbrace{z_t\in\mathbb{R}^d}_{\substack{\text{Wasserstein}\\ \text{coordinate}}}
\;\xrightarrow{\;\text{windowed Hankel-DMD}\;}\;
\underbrace{\bigl(\{\lambda_j\},\, r_{0.99},\, \mathrm{Res}^{(r)}\bigr)}_{\substack{\text{spectrum / effective rank}\\ \text{/ residual}}}
\]

\begin{minipage}{0.96\linewidth}
\small
\emph{How to read the diagnostics.}
Low \(\mathrm{Res}^{(r)}\): interpret the spectrum and \(r_{0.99}\) as local regime descriptors;
high \(\mathrm{Res}^{(r)}\): interpret the residual itself as a transition / fragility signal.
\end{minipage}

\caption{\textbf{Pipeline of the proposed diagnostic.}
The selected task-dependent observable \(o_t\) at each training step is summarized as an empirical distribution \(\mu_t\).
Wasserstein/quantile coordinates convert each \(\mu_t\) into a vector observation \(z_t\in\mathbb{R}^d\).
Windowed Hankel-DMD then analyzes the local temporal evolution of \(\{z_t\}\) over fixed step windows and returns spectrum, effective rank, and reconstruction residual.
Low residual supports interpreting the spectrum and effective rank as local descriptors; high residual is treated as a transition/fragility signal.}
\label{fig:pipeline}
\end{figure}

\paragraph{AGOP as a parallel route.}
The Average Gradient Outer Product (AGOP) \citep{radhakrishnan2024agop} summarizes input sensitivity through averaged input-gradient outer products and has been used to study feature emergence and grokking-related transitions. AGOP-based diagnostics provide a parallel route under sufficient checkpoint coverage; in our setup, sparse checkpoint coverage prevents a fair quantitative comparison, so we treat AGOP as corroborative rather than competitive evidence (Appendix~\ref{app:agop}).

%% file: Chapters/Results.tex
\section{Experiments}\label{sec:experiments}

We evaluate the residual as a transition-localization signal under an explicit threshold/FPR/lead-time protocol in modular-addition Transformers and report scope checks for related settings. 
Section~\ref{subsec:protocol} fixes the diagnostic protocol and observable choice; Section~\ref{subsec:detection} reports detection performance on a held-out test fold; Section~\ref{subsec:fragility} reports perturbation fragility, norm baselines, and architecture scale; Section~\ref{subsec:scope} summarizes scope checks. All experiments run on a single workstation (NVIDIA RTX 4070 Laptop, 8\,GB VRAM; PyTorch 2.7.1, CUDA 12.8).

\subsection{Diagnostic protocol and observable choice}\label{subsec:protocol}

This section addresses two design questions: which observable should be used, and how the detection protocol is defined.

\paragraph{Setup.}
We train a decoder-only Transformer ($d_{\mathrm{model}}{=}128$, $n_{\mathrm{layers}}{=}2$, $n_{\mathrm{heads}}{=}4$, AdamW, training fraction $0.4$, $6{,}000$ max steps) on the modular-addition task of \citet{power2022grokkinggeneralizationoverfittingsmall}. The base pool consists of $41$ unique runs ($8$ seeds $\times$ $5$ weight-decay settings $+\,1$ smoke run); full hyperparameters are in Appendix~\ref{app:exp}. The default observable is the empirical distribution of correct-answer log-probabilities on a fixed probe set, defined precisely below and summarized by $19$ fixed quantile coordinates ($0.05$--$0.95$).

\paragraph{Transformer observable.}
We instantiate the distributional state of \S\ref{subsec:wass} as follows (see Eq.~\ref{eq:mu-tr}). At each training step $t$ we evaluate the current model on a fixed probe set of $M=100$ training examples and record the scalar correct-answer log-probability $o_{t,i}=\log p_{\theta_t}(y_i^\star\mid x_i)$ for each example $i$. The distributional state $\mu_t$ is the empirical distribution of these $M$ scalars, and $z_t$ is its quantile coordinate on the same $19$-level grid as in \S\ref{subsec:wass}. This definition also explains the observable ablation below: the empirical distribution of correct-answer log-probabilities retains distributional information about the output dynamics, whereas logits, top-$k$ summaries, and hidden-state observables provide weaker or less stable signals under the current DMD configuration.

\paragraph{Onset and labeling.}
Grokking onset is defined as the first step at which test accuracy crosses $99\%$. In the base pool, mean onset for wd$=1$ runs is $3398 \pm 361$ steps, while wd$=2$ runs reach the same threshold by step $1528$ on average; the bimodal gap between these distributions motivates a step threshold separating grokking from early generalization (justification and onset histogram in Appendix~\ref{app:onset}).

\paragraph{Operating rule.}
A run-level alarm fires when $\mathrm{Res}^{(r)}$ exceeds $\tau\!\times\! b_{\mathrm{run}}$ for $K$ consecutive windows, where $b_{\mathrm{run}}$ is a per-run residual baseline. We treat $K{=}2,\tau{=}10$ as a fixed heuristic operating point: $\tau$ is set on the order of $10\times$ baseline (residual-multiplier heuristic) and $K{=}2$ enforces sustainment. Held-out fresh seeds are used only for evaluation, and we report the full threshold sweep to expose the sensitivity--specificity trade-off (Appendix~\ref{app:threshold}). We do not select $\tau$ by AUROC or AUPRC optimization on either fold.

\paragraph{Observable ablation.}
Replacing log-probability with alternative observables under the same DMD configuration (Appendix~\ref{app:n3-obs}) gives, on the $4$ grok runs of the ablation sub-pool: log-probability fires \texttt{sustained\_K2\_tau10} alarms on $4/4$ runs, of which $2/4$ land before onset (TP) with median lead $\approx 601$ steps; the matched FPR is the rule's test-fold value $0.50$ from \S\ref{subsec:detection}. Logits fire alarms on $3/4$; top-$k$ ($\sim 95$-dim) and hidden ($\sim 1900$-dim) trigger on $0/4$. Among the observables we tested under the current protocol, log-probability is the best-performing default. The present implementation is best suited to scalar empirical observables represented by one-dimensional quantile coordinates, so the failure of top-$k$ and hidden-state variants under the same DMD configuration should be interpreted as a limitation of this scalar-distribution implementation and fixed DMD setup, not as evidence that those observables lack useful information.

\paragraph{DMD-vs-persistence calibration.}
To check whether the Hankel-DMD fit captures temporal structure beyond a trivial identity predictor, we compare one-step holdout prediction within each window against a persistence baseline (predicting that the next state equals the current state). Across the tested weight-decay regimes, Hankel-DMD improves over persistence (Table~\ref{tab:dmd_persistence}). We use this as a calibration check that the residual is not merely a noise-fitting artifact, not as evidence that DMD is a uniformly reliable forecaster (full setup in Appendix~\ref{app:dmd-quality}).

\begin{table}[h]
\centering
\small
\caption{DMD-vs-persistence calibration. Holdout error and persistence error are relative residuals on within-window holdout splits, aggregated across seeds and segment sizes per weight-decay setting. Positive gain (persistence minus holdout) means Hankel-DMD improves over the persistence baseline.}
\label{tab:dmd_persistence}
\begin{tabular}{c|ccc}
\toprule
Weight decay & DMD holdout error & Persistence error & Gain \\
\midrule
$0$ & $0.308$ & $0.654$ & $+0.346$ \\
$1$ & $0.276$ & $0.378$ & $+0.102$ \\
$2$ & $0.252$ & $0.457$ & $+0.205$ \\
\bottomrule
\end{tabular}
\end{table}

\subsection{Residual localizes grokking transitions}\label{subsec:detection}

We next evaluate the residual as a transition-localization signal, reporting false alarms and lead time jointly.

\textit{Finding 1.} Paired with the sustained-threshold operating rule of \S\ref{subsec:protocol}, the reconstruction residual yields a held-out detector for grokking-vs-non-grokking. Lead time is threshold-dependent and is reported jointly with FPR and CI.

\paragraph{Protocol.}
Test fold: fresh seeds $46$--$49$, $n_{\mathrm{grok}}{=}5$, $n_{\mathrm{non}}{=}12$ (base rate $\approx 0.29$); wd$=2$ early-generalization runs are excluded from numerator and denominator. Onset is the first step crossing $99\%$ test accuracy. Operating rule: \texttt{sustained\_K2\_tau10} fixed by the heuristic of \S\ref{subsec:protocol}. Lead time is computed only over true-positive alarms. Unless otherwise stated, AUROC and AUPRC reported in this section are run-level ranking metrics computed from residual-based alarm scores on held-out runs; the temporal alarm metrics (TPR, FPR, lead, CI) describe the first alarm time and lead relative to grokking onset under a thresholded rule. Window-level evidence appears in \S\ref{subsec:fragility} (perturbation and norm-window controls).

\begin{figure}[h]
  \centering
  \includegraphics[width=0.5\linewidth]{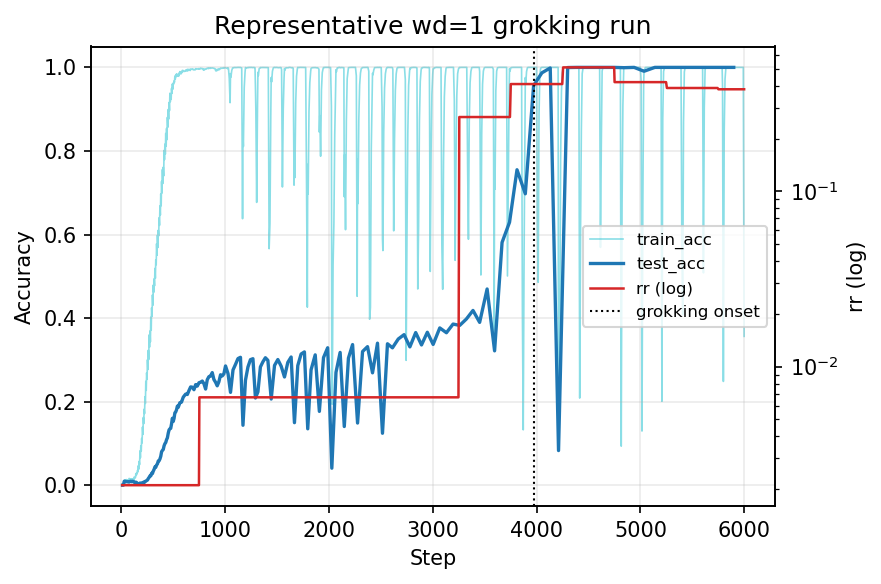}
  \caption{\textbf{Representative wd$=1$ grokking run.} 
The dark-blue curve shows test accuracy (left axis), the light-cyan 
curve in the background shows training accuracy (left axis, included 
as a reference for the memorization phase), and the red curve shows 
the reconstruction residual on a log scale (right axis). The vertical 
dotted line marks grokking onset, defined as the first step at which 
test accuracy exceeds $99\%$. In this run, the rising edge of the 
residual precedes the onset.}
  \label{fig:case-success}
\end{figure}

\paragraph{Evidence.}
On this fold, the residual achieves AUROC $\approx 0.93$ (AUPRC $\approx 0.91$). At the selected operating point, TPR $=0.80$ and FPR $=0.50$; the median lead computed only over true-positive alarms is $1068$ steps ($95\%$ bootstrap CI $[142,\,2426]$; Appendix~\ref{app:uncertainty}). Threshold trade-off (Table~\ref{tab:detection}): an instantaneous $\tau{=}5\times$ rule recovers all grok runs (TPR $1.00$) at the cost of FPR $0.917$ and longer lead ($1774$ steps); a stricter instantaneous $\tau{=}20\times$ rule reduces FPR to $0.250$ but recalls only $40\%$ of grok runs. Figure~\ref{fig:case-success} shows a representative wd$=1$ run with the residual rising before onset; full ROC and lead-time distributions on the test fold are in Appendix~\ref{app:threshold}.

\paragraph{Limitation.}
On the reused-seed split (seeds $42$--$45$), the same fixed \texttt{sustained\_K2\_tau10} operating point fires almost no alarms, yielding TPR $=0$ and FPR $=0$ on that split (Appendix~\ref{app:threshold}). We therefore report this asymmetry as seed-split sensitivity rather than as evidence of calibrated threshold selection: the reused-seed split does not validate threshold calibration. High absolute residual can occur in non-grokking runs, motivating the relative sustained-threshold rule rather than an absolute cut. Lead time must always be read together with FPR and CI. The selected operating rule is one specific point on a sensitivity--specificity trade-off.

\begin{table}[h]
\centering
\caption{Detection threshold trade-off on the held-out test fold (fresh seeds $46$--$49$; $n_{\mathrm{grok}}{=}5$, $n_{\mathrm{non}}{=}12$, base rate $\approx 0.29$; wd$=2$ early-generalization runs excluded). Lead is computed only over true-positive alarms; the bracketed range for the selected sustained rule is the $95\%$ bootstrap CI of the median (Appendix~\ref{app:uncertainty}). AUROC $\approx 0.93$ (AUPRC $\approx 0.91$) is computed on this same fold and applies to all rows (single detector at different cuts). The instantaneous $\tau{=}5\times$ rule gives high recall at high FPR; the instantaneous $\tau{=}20\times$ rule gives high specificity at low recall; the selected sustained rule \texttt{sustained\_K2\_tau10} sits at a moderate recall/FPR point on the same curve.}
\label{tab:detection}
\small
\begin{tabular}{lcccc}
\toprule
Rule & TPR $(\uparrow)$ & FPR $(\downarrow)$  & Median lead & $95\%$ CI  \\
\midrule
\multicolumn{5}{l}{\emph{Instantaneous threshold rules}} \\
$\tau{=}5\times$              & $1.00$ & $0.917$ & $1774$ & ---  \\
$\tau{=}20\times$             & $0.40$ & $0.250$ & ---    & ---  \\
\midrule
\multicolumn{5}{l}{\emph{Sustained operating rule}} \\
\texttt{sustained\_K2\_tau10} (selected) & $0.80$ & $0.500$ & $1068$ & $[142,\,2426]$  \\
\bottomrule
\end{tabular}
\end{table}

\paragraph{AUROC/AUPRC vs.\ operating point.}
The AUROC and AUPRC summarize ranking performance across all thresholds; Table~\ref{tab:detection} reports concrete operating points on the corresponding sensitivity--specificity curve. Lower thresholds give earlier alarms at higher false-positive rates, while higher thresholds reduce false positives at the cost of missed or delayed alarms. The selected operating rule is one configuration on this curve, intended to localize candidate transition windows for further inspection rather than to serve as a single decision rule in isolation. Because the held-out split is small, we report run-level bootstrap confidence intervals and operating-point uncertainty estimates in Appendix~\ref{app:uncertainty}; these intervals support the run-level ranking value of RR while making the uncertainty of the fixed operating point explicit.

\subsection{Fragility, norm baselines, and architecture scale}\label{subsec:fragility}

We now test whether high-residual windows have functional meaning through perturbations, compare the residual against norm-based baselines, and check sensitivity to model scale.

\paragraph{Perturbation fragility.}

\textit{Finding 2.} In the wd$=1$ pool we tested, identical perturbations applied at high-residual windows produce roughly $3\times$ larger short-horizon deviation than at low-residual windows. The result quantifies functional fragility under matched noise; it does not establish a causal mechanism.

\begin{figure}[h]
  \centering
  \includegraphics[width=0.55\linewidth]{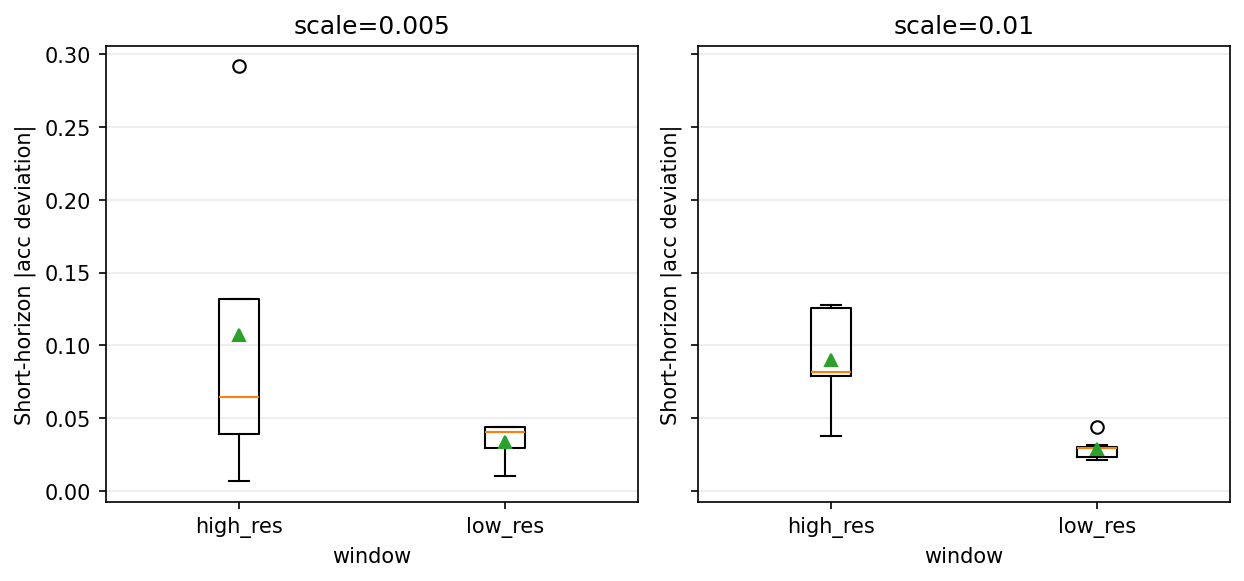}
  \caption{\textbf{Perturbation fragility.} Short-horizon deviation by RR window and noise scale on the wd$=1$ pool. High-RR windows show larger short-horizon deviation under matched noise than low-RR windows; the low-RR early-training failure is a boundary case discussed in Appendix~\ref{app:sensitivity-window}.}
  \label{fig:e5}
\end{figure}

Protocol: identical multiplicative perturbations at scales $\{0.005, 0.01\}$ are applied at high-RR vs.\ low-RR windows on wd$=1$ grokking baselines ($4$ high-RR runs and $5$ low-RR runs at each scale; details in Appendix~\ref{app:sensitivity-window}). Evidence: at scale $0.01$, mean short-horizon deviation is $0.090$ (high-RR) vs.\ $0.029$ (low-RR), giving high/low ratio $\approx 3.1\times$; at scale $0.005$, $0.107$ vs.\ $0.034$, ratio $\approx 3.2\times$. Figure~\ref{fig:e5} shows the deviation distribution. Limitation: one unrecoverable failure occurs in a high-RR window near the transition; one low-RR early-training failure indicates a separate instability unrelated to transition fragility and is treated as a boundary case in Appendix~\ref{app:sensitivity-window}.

\paragraph{Norm baselines and a same-data window control.}

\textit{Finding 3.} Norm-derived signals are strong run-level regime indicators on the shared $22$-run pool, confirming that scale dynamics are informative for grokking. In the same-data norm-window control, perturbation sensitivity aligns with the residual ordering rather than total-norm ordering, suggesting that the residual is not merely a total-norm proxy at the window level in the studied wd$=1$ dynamics.

Protocol: (i) the same $21$ perturbation runs ($17$ reused above plus $4$ new low-norm $s{=}5250$ perturbations on seeds $46$/$49$) are re-labeled by total-parameter-norm percentile rather than RR percentile (window-control protocol); (ii) run-level (max-of-trajectory) AUROC for several signals is computed on a shared $22$-run pool; (iii) a fair-FPR temporal alarm matches FPR to $0.50$ on the reused-seed split for each signal independently. Note that the fair-FPR protocol in (iii) is a baseline-comparison protocol used to set comparison thresholds and is \emph{not} the \texttt{sustained\_K2\_tau10} operating rule used in \S\ref{subsec:detection}; details in Appendix~\ref{app:norm-baseline}.

Evidence:
\begin{itemize}
\setlength\itemsep{0pt}
\item Window-control deviation (scale $0.01$): under RR-window framing the high/low ratio is $\approx 3.82\times$; under norm-window framing the same runs give $\approx 0.41\times$, reversing the ordering.
\item Run-level AUROCs under run-level max-of-trajectory scoring on the shared norm-baseline pool ($22$ runs): total parameter norm $0.9829$, $\lvert\Delta\mathrm{norm}\rvert$ $0.9915$, RR (run-level max) $0.4103$. This run-level RR AUROC should not be compared directly with the selected sustained-threshold result in \S\ref{subsec:detection}; the protocols differ.
\item Under the fair-FPR temporal alarm (FPR target $0.50$ on the reused-seed split), RR triggers TPR $=0.75$ on the test fold with median lead $273.5$ steps; \texttt{norm\_N\_total} triggers TPR $=0$ (no alarms fire) under the same fair-FPR protocol. This is a different protocol from the selected sustained-threshold operating rule of \S\ref{subsec:detection}.
\end{itemize}

Limitation: norm and RR are temporally anti-correlated in wd$=1$ dynamics, so the two framings select temporally adjacent windows; the control distinguishes which signal aligns with perturbation outcome direction but does not establish causal mechanism. Norm signals' high run-level AUROC and the residual's window-level perturbation alignment are complementary use cases. We do not claim RR universally outperforms norm baselines, and a fully fair head-to-head detection comparison requires per-signal threshold recalibration. Appendix pointer: Appendix~\ref{app:norm-baseline}.

\paragraph{Architecture scale.}

\textit{Finding 4.} The residual signal is observed across the three tested Transformer scales, with onset and amplitude shifting with capacity. This evaluates model-scale sensitivity, not standard-vs-mean-field parameterization.

Protocol and evidence: small ($d_{\mathrm{model}}{=}64$, $1$ layer), baseline ($d_{\mathrm{model}}{=}128$, $2$ layers), and large ($d_{\mathrm{model}}{=}256$, $3$ layers) variants are trained at wd$\in\{0,1,2\}$ with $2$ seeds per scale; the per-scale aggregated table is in Appendix~\ref{app:arch-ablation}. Limitation: residual amplitude shrinks with capacity, so the relative-threshold rule's behavior at small scale requires further investigation; this ablation does not test parameterization scaling.

\subsection{Scope and remaining checks}\label{subsec:scope}

The remaining checks are summarized in Table~\ref{tab:scope}; full plots and per-row protocol details are in the cited appendices. Lead times reported in this table use the protocol of the corresponding appendix and are paired with the FPR of that appendix's protocol.

\begin{table}[h]
\centering
\caption{\textbf{Compact summary of scope and robustness checks.} ``$\rho$'' denotes the lead-lag Spearman correlation between $\mathrm{Res}^{(r)}$ and subsequent test-accuracy change. Each appendix gives full plots and the per-row protocol; lead-time figures (where applicable) are paired with the FPR of that protocol.}
\label{tab:scope}
\small
\begin{tabular}{p{1.7cm}p{2.6cm}p{3.4cm}p{2.7cm}c}
\toprule
Check & Evidence & Main takeaway & Limitation & App.\ \\
\midrule
Segment size & wd$=1$, $\{250,500,1000\}$ steps & lead-lag $\rho > 0.80$ for grokking runs at all tested sizes; peak-RR and onset orderings preserved across sizes & RR amplitude shifts with size; quantitative cell values vary & \ref{app:partition} \\
Cross-task transfer & mod-mult; mod-add fraction/prime sweep & the same rule fires on mod-mult wd$=1$ runs & shorter lead than mod-add; explicit failure cases (e.g.\ frac.\ $=0.5$) & \ref{app:task-transfer} \\
Baseline-battle pool & broader $18$-run pool ($5$ grok / $13$ non-grok) & RR detection holds (AUROC $\approx 0.92$) on this pool & AGOP coverage $=1$ run; no fair head-to-head & \ref{app:agop} \\
Triggered intervention & $10$ wd$=1$ runs, RR-triggered lr halving & no reliable speedup vs.\ fixed/random schedules & $0.10$ failure rate (vs.\ $0$ for fixed/random) & \ref{app:intervention} \\
CIFAR-10 portability & Tiny CNN, $8$ runs, $5$ epochs & pipeline runs end-to-end in a different setting & no detection metric reported; not a grokking benchmark & \ref{app:cifar} \\
FCN secondary & MNIST, $h\in\{5,40,256,1024\}$ & RR/$r_{0.99}$ as low-RR regime descriptors & no grokking-diagnostic claim & \ref{app:fcn-secondary} \\
\bottomrule
\end{tabular}
\end{table}

%% file: Chapters/Discussion.tex
\section{Discussion}

We formulated grokking transition localization as a diagnostic problem with an explicit threshold/FPR/lead-time trade-off on observed training trajectories and instantiated it via a windowed distributional pipeline: task-dependent observables are summarized as empirical distributions, mapped to Wasserstein/quantile coordinates, and analyzed by Hankel-DMD, with the reconstruction residual as the primary diagnostic and the spectrum and effective rank as auxiliary low-residual descriptors. The framing is deliberately narrow: the residual is a window-level monitoring and localization signal in the studied modular-arithmetic Transformer settings, not a universal predictor or an intervention rule.

\paragraph{What the evidence supports.}
Paired with a fixed sustained-threshold operating rule of \S\ref{subsec:protocol}, the reconstruction residual yields nontrivial held-out detection behavior (numbers in \S\ref{subsec:detection} and Table~\ref{tab:detection}), with lead time, false-alarm rate, and bootstrap CI reported jointly. In the wd$=1$ pool we tested, identical perturbations applied at high-RR windows produce larger short-horizon deviations than at low-RR windows, and a same-data norm-window control reverses the fragility ordering (\S\ref{subsec:fragility}). Log-probability performs best among the observables tested under the current protocol
(Appendix~\ref{app:n3-obs}).

\paragraph{What the diagnostic adds.}
The main value of the reconstruction residual is not to replace mechanistic, norm-based, or gradient-based analyses, but to provide a lightweight window-level readout of when the observed output distribution is being reconfigured. Unlike run-level norm scores, which are strong regime indicators, the residual is evaluated locally on fixed step windows and highlights candidate transition periods for further inspection. The perturbation and norm-window controls suggest that these windows have functional meaning: matched noise produces larger deviations in high-RR windows, and the same runs do not show the same ordering when re-labeled by total parameter norm. Thus, the residual provides a complementary output-distribution view of training dynamics: it can be computed post hoc from a fixed probe set, does not rely on hidden-unit indexing, and helps identify where mechanistic or gradient-based analyses should focus.

\paragraph{What it does not support.}
Norm-derived signals are strong run-level regime indicators on the shared pool (run-level AUROC $\approx 0.98$--$0.99$ for total norm and $\lvert\Delta\mathrm{norm}\rvert$ vs.\ $0.41$ for run-level RR); we do not claim the residual replaces them as a regime classifier. Lead times are threshold-dependent and must be paired with FPR. Perturbation alignment between residual ordering and short-horizon deviation quantifies sensitivity, not a causal mechanism. AGOP coverage in our setup is one run on the baseline-battle pool, so a head-to-head AGOP comparison is not feasible; AGOP is treated as a parallel route under sufficient checkpoint coverage (Appendix~\ref{app:agop}). An RR-triggered learning-rate intervention introduces a non-zero failure rate without a reliable speedup, so monitoring does not directly imply beneficial control (Appendix~\ref{app:intervention}).

\paragraph{Boundaries.}
The current implementation assumes a scalar empirical observable represented by one-dimensional quantile coordinates; the failure of top-$k$ and hidden-state observables to trigger any alarm under the current DMD configuration is best interpreted as a limitation of this scalar-distribution implementation and fixed DMD setup rather than as a verdict on those observables' information content. Extending the diagnostic to genuinely multidimensional observables would require separate geometric embeddings, such as sliced Wasserstein representations, kernel mean embeddings, or layer-wise scalar summaries, and is outside the scope of the present study. The signal is observed across the three tested Transformer scales but with shrinking residual amplitude at small scale (Appendix~\ref{app:arch-ablation}); coarse conclusions are stable across segment sizes $\{250,500,1000\}$ while fine ordering may vary (Appendix~\ref{app:partition}). Cross-task transfer is partial, with shorter lead on modular multiplication and explicit failure cases on the fraction/prime sweep (Appendix~\ref{app:task-transfer}).

\paragraph{Future work.}
Layer-wise observables and multi-dimensional Wasserstein embeddings such as kernel mean embeddings \citep{kme10.1007/978-3-540-75225-7_5} that respect distributional geometry beyond the one-dimensional case; per-signal threshold recalibration for fully fair head-to-head detection comparisons against norm-based baselines; non-autonomous extensions accounting for schedule-induced drift \citep{mandt2018stochasticgradientdescentapproximate}; and large-scale evaluation across architectures and task families. Mechanistic accounts remain complementary: they explain \emph{why} grokking occurs, while the residual addresses \emph{where in training} the transition can be located from chosen task-dependent observables.

%% file: Chapters/Appendix.tex
\section{Why one-dimensional distributions}\label{app:why-1d}

Section~\ref{sec:method} introduces how to project a one-dimensional distribution onto a tangent space and obtain a finite-dimensional coordinate representation via fixed quantile levels. However, for multi-dimensional distributions, there is generally no global isometric isomorphism that maps the Wasserstein space into a linear space \citep{villani2009optimal}, which creates a practical obstacle to constructing the finite-dimensional snapshot coordinates required in Section~\ref{subsec:hankel-rrr}. One may instead use kernel mean embeddings (KME) \citep{kme10.1007/978-3-540-75225-7_5} or multidimensional scaling (MDS) to obtain a finite-dimensional representation, but this typically introduces systematic distortion and does not preserve Wasserstein geometry, limiting the fidelity of the resulting dynamical characterization in the chosen observation space.

\section{Wasserstein geometry: full background}\label{app:wass-detail}

This appendix expands on the compact Wasserstein-quantile material in \S\ref{subsec:wass}. Let $(\mathcal{X},\|\cdot\|)$ be a Polish space and let $\mathcal{P}_2(\mathcal{X})$ denote the set of Borel probability measures on $\mathcal{X}$ with finite second moments,
\[
\mathcal{P}_2(\mathcal{X}) := \Bigl\{\mu\in\mathcal{P}(\mathcal{X}):\ \int_{\mathcal{X}}\|x\|^2\,d\mu(x) < \infty\Bigr\}.
\]
For $\mu,\nu\in\mathcal{P}_2(\mathcal{X})$, the $2$-Wasserstein distance is
\[
W_2(\mu,\nu) := \Bigl(\inf_{\pi\in\Pi(\mu,\nu)}\int_{\mathcal{X}\times\mathcal{X}}\|x-y\|^2\,d\pi(x,y)\Bigr)^{1/2},
\]
where $\Pi(\mu,\nu)$ is the set of couplings of $\mu$ and $\nu$. This is the minimum quadratic transport cost from $\mu$ to $\nu$ \citep{villani2009optimal}.

For $D\subseteq\mathbb{R}$ closed and $\mathcal{W}_2(D)$ the corresponding $1$-D Wasserstein space,
\[
d_W^2(\mu,\nu) = \int_0^1 \bigl(F_\mu^{-1}(p) - F_\nu^{-1}(p)\bigr)^2\,dp,
\]
with $F_\mu^{-1}$ the quantile function. If $\mu\in\mathcal{W}_2(D)$ is atomless, the map $\mu\mapsto F_\mu^{-1}\in L^2(0,1)$ is a global isometric embedding, identifying $\mathcal{W}_2(D)$ with a closed convex subset of $L^2(0,1)$. Consequently $(\mathcal{W}_2(D),d_W)$ is a Hadamard space (complete, geodesic, non-positively curved), Fréchet means are uniquely defined, and at any atomless reference $\mu^\star$ the tangent space $T_{\mu^\star}$ is identified with a closed subspace of $L^2(\mu^\star)$ via the logarithmic map
\[
\log_{\mu^\star}(\mu) = F_\mu^{-1}\circ F_{\mu^\star} - \mathrm{id}.
\]
This isometry justifies treating the windowed dynamics on a linear coordinate (the quantile function evaluated on a fixed grid) as in \S\ref{subsec:wass}.

\section{Hankel-DMD: full construction}\label{app:dmd-detail}

This appendix expands on \S\ref{subsec:hankel-rrr}. Koopman-operator theory analyzes a nonlinear system through an infinite-dimensional linear operator acting on a chosen observation space \citep{SCHMID_2010,rowley2009}. In practice, dynamic mode decomposition (DMD) \citep{SCHMID_2010,H_Tu_2014,Arbabi_2017,Brunton_2017,drmač2017datadrivenmodaldecompositions} approximates the Koopman operator and its spectrum from snapshots.

\paragraph{Random-system interpretation.}
Following \citet{2019koopmanrds,H_Tu_2014}, let $(X_t)_{t\ge 0}$ be a stochastic process on $\mathcal{X}$ and $\psi:\mathcal{X}\to\mathbb{R}^d$ a fixed observable vector (here, the quantiles of the distributional state); $z_t = \psi(X_t)\in\mathbb{R}^d$. The Koopman operator $K$ acts on observables $g:\mathcal{X}\to\mathbb{R}$ by $(Kg)(x)=\mathbb{E}[g(X_{t+1})\mid X_t=x]$. The vectors $z_t$ are evaluations of a finite collection of observables evolving under $K$.

\paragraph{Raw DMD on a window.}
Collecting $m+1$ successive observations into snapshot matrices
\[
Z_- = [z_0\ \cdots\ z_{m-1}]\in\mathbb{R}^{d\times m},\qquad
Z_+ = [z_1\ \cdots\ z_m]\in\mathbb{R}^{d\times m},
\]
raw DMD seeks $A\in\mathbb{R}^{d\times d}$ minimising $\|Z_+ - A Z_-\|_F^2$, with closed form $A^\star = Z_+ Z_-^{\dagger}$ when $Z_-$ has full row rank. $A^\star$ is the orthogonal projection (under the empirical inner product) of the action of $K$ onto the Krylov subspace $\mathcal{K}_m=\mathrm{span}\{z_0,\dots,z_{m-1}\}$; its eigenvalues (Ritz values) give a finite-dimensional approximation of the spectral properties of $K$ restricted to that subspace.

\paragraph{Non-normal sensitivity.}
With noise and finite samples, $A^\star$'s eigenvalues are sensitive to small perturbations of $(Z_-,Z_+)$ when the underlying operator is non-normal (large pseudospectrum). Raw DMD spectra may then be unstable or contain spurious modes.

\paragraph{Hankel time-delay extension.}
Fix $q\ge 1$ and form delay-stacked vectors $\xi_t = [z_t^\top, z_{t+1}^\top,\ldots,z_{t+q-1}^\top]^\top\in\mathbb{R}^{qd}$. The block Hankel snapshots are
\[
H_- = [\xi_0\ \cdots\ \xi_{m-q-1}],\qquad H_+ = [\xi_1\ \cdots\ \xi_{m-q}].
\]
Hankel-DMD solves the ordinary least-squares problem $A_H^\star = \arg\min_A\|H_+ - A H_-\|_F^2$ and studies its eigenvalues; the time-delay embedding enlarges the observable space and encodes temporal correlations over $q$ lags, typically improving spectral robustness and allowing continuous spectral components to be approximated by clusters of discrete modes \citep{Arbabi_2017,H_Tu_2014}. We then truncate the fitted operator to a low-rank approximation by retaining the leading $r$ eigenpairs of $A_H^\star$ (after a stabilizing SVD projection onto the dominant snapshot subspace) \citep{drmač2017datadrivenmodaldecompositions}.

\paragraph{Spectral linear approximation.}
Let $A_H^\star W = W\Lambda$ with $\Lambda=\mathrm{diag}(\lambda_1,\dots,\lambda_r)$ and $W=[w_1,\dots,w_r]$. For a delay state $\xi_0$, modal amplitudes are $b = W^{\dagger}\xi_0$, and the rank-$r$ reconstruction is
\[
\widehat{\xi}_t^{(r)} := W\Lambda^t b = \sum_{j=1}^r b_j\,\lambda_j^t\, w_j,\qquad t=0,\dots,m-q.
\]
This is a spectral, low-rank linear approximation of the windowed evolution within the data-driven subspace. Each eigenpair represents a direction; real eigenvalues correspond to non-oscillatory modes and complex eigenvalues to oscillatory ones \citep{rowley2009}. The directions are not in general orthogonal; reading $\widehat{\xi}_t^{(r)}$ as a complete orthogonal decomposition requires substantially stronger assumptions. The reconstruction residual~(\ref{eq:rr}) and effective rank~(\ref{eq:reff}) are the two windowed scalars we use to summarize a window's dynamics.

\section{Training experiment details}\label{app:exp}
\subsection{Fully connected networks (FCNs)}
The FCN experiment considers a single-hidden-layer fully connected network trained on MNIST. This part is mainly based on \citet{redman2024identifyingequivalenttrainingdynamics}'s open-source repository (\url{https://github.com/william-redman/Identifying_Equivalent_Training_Dynamics}), and we keep closely matched hyperparameters. We use the standard PyTorch initialization scheme, but instead of the repository’s multiplicative perturbation protocol, we obtain different initializations directly by changing the random seed. This choice is consistent with our Wasserstein-geometry perspective, under which seed-wise initializations are treated as independent samples from the same underlying initialization distribution and hence correspond to a common starting point on the Wasserstein manifold. We additionally report the same spectral diagnostics for FCNs trained with AdamW.

Concretely, the FCN hyperparameters used in this work are summarized in Table~\ref{tab:fcn_hparams}.
\begin{table}[htbp]
  \caption{FCN hyperparameters}
  \label{tab:fcn_hparams}
  \begin{center}
    \begin{small}
      \begin{sc}
        \begin{tabular}{cc}
          \toprule
          Hyperparameters         & Values  \\
          \midrule
          Learning rate ($\eta$)& 0.1  \\
          Batch size ($b$) & 60 \\
          Optimizer & SGD, AdamW  \\
          Epochs & 1           \\
          Activation functions & ReLU, GeLU \\
          
          \bottomrule
        \end{tabular}
      \end{sc}
    \end{small}
  \end{center}
  \vskip -0.1in
\end{table}

We use 25 different random seeds and, following Sections~\ref{sec:method}, construct distributional-state snapshots, using 19 quantile levels (0.05--0.95) as a finite-dimensional coordinate representation of the empirical distribution, collect snapshots every step, define stages by non-overlapping 500-step windows 
 and compute DMD. For all experiments, we fix the delay length to $4$, and in DMD-RRR we retain $10$ Koopman modes. Note that in a small number of windows the effective rank can be below 10; however, to enable consistent visual comparisons of spectral plots and take advantage of the randomized shuffle control \citet{redman2024identifyingequivalenttrainingdynamics}, we report 10 spectral points in all cases. 

\paragraph{Randomized shuffle control }

Let $\Omega_1=\{\lambda^{(1)}_j\}_{j=1}^k$ and $\Omega_2=\{\lambda^{(2)}_j\}_{j=1}^k$ be two estimated spectral point sets, and let
$\omega := d(\Omega_1,\Omega_2)$ denote their spectral distance (e.g., the 2-Wasserstein / optimal-matching distance).
Following \citet{redman2024identifyingequivalenttrainingdynamics}, we construct a ``shuffle'' baseline by first computing an optimal matching
$\sigma\in S_k$ that minimizes $\sum_{j=1}^k \|\lambda^{(1)}_j-\lambda^{(2)}_{\sigma(j)}\|$.
Then, for each matched pair $\big(\lambda^{(1)}_j,\lambda^{(2)}_{\sigma(j)}\big)$, we independently swap the two elements with probability $1/2$,
forming a randomized partition $(\Omega_1',\Omega_2')$ that preserves pairwise proximity while removing systematic group structure.
Repeating this procedure $n_{\mathrm{shuff}}$ times yields distances $\{\omega'_i=d(\Omega'_{1,i},\Omega'_{2,i})\}_{i=1}^{n_{\mathrm{shuff}}}$.
We report the empirical exceedance rate
\[
\hat p \;:=\; \frac{1}{n_{\mathrm{shuff}}}\sum_{i=1}^{n_{\mathrm{shuff}}} \mathbf{1}\{\omega'_i \ge \omega\},
\]
which quantifies how often a ``naturally shuffled'' pair is at least as separated as the observed spectra.

\begin{figure*}[t]
  \centering
  \subfigure[h=40]{%
    \includegraphics[width=0.5\linewidth]{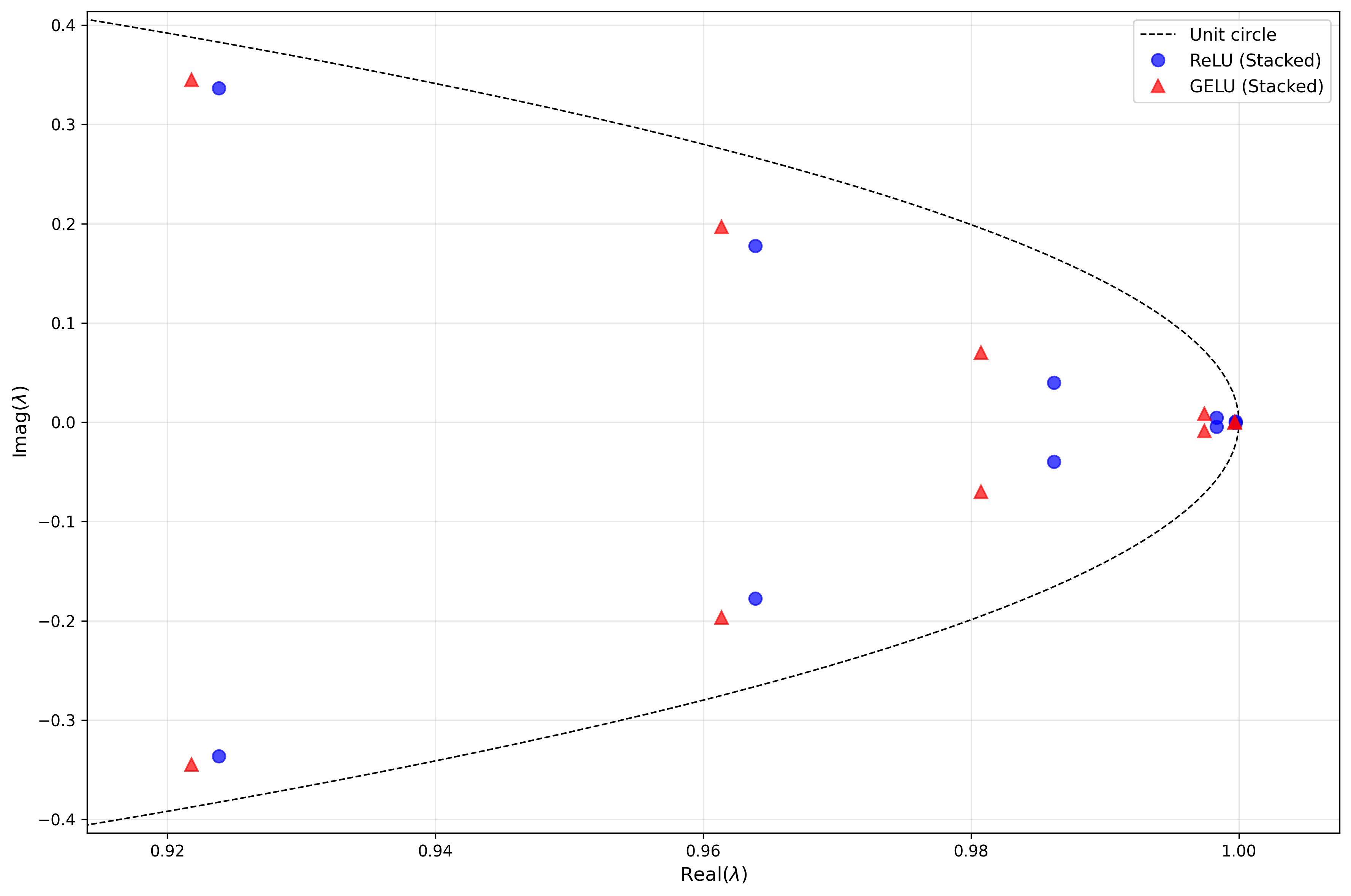}%
    \label{fig:gelu40}%
  }\hfill
  \subfigure[h=256]{%
    \includegraphics[width=0.5\linewidth]{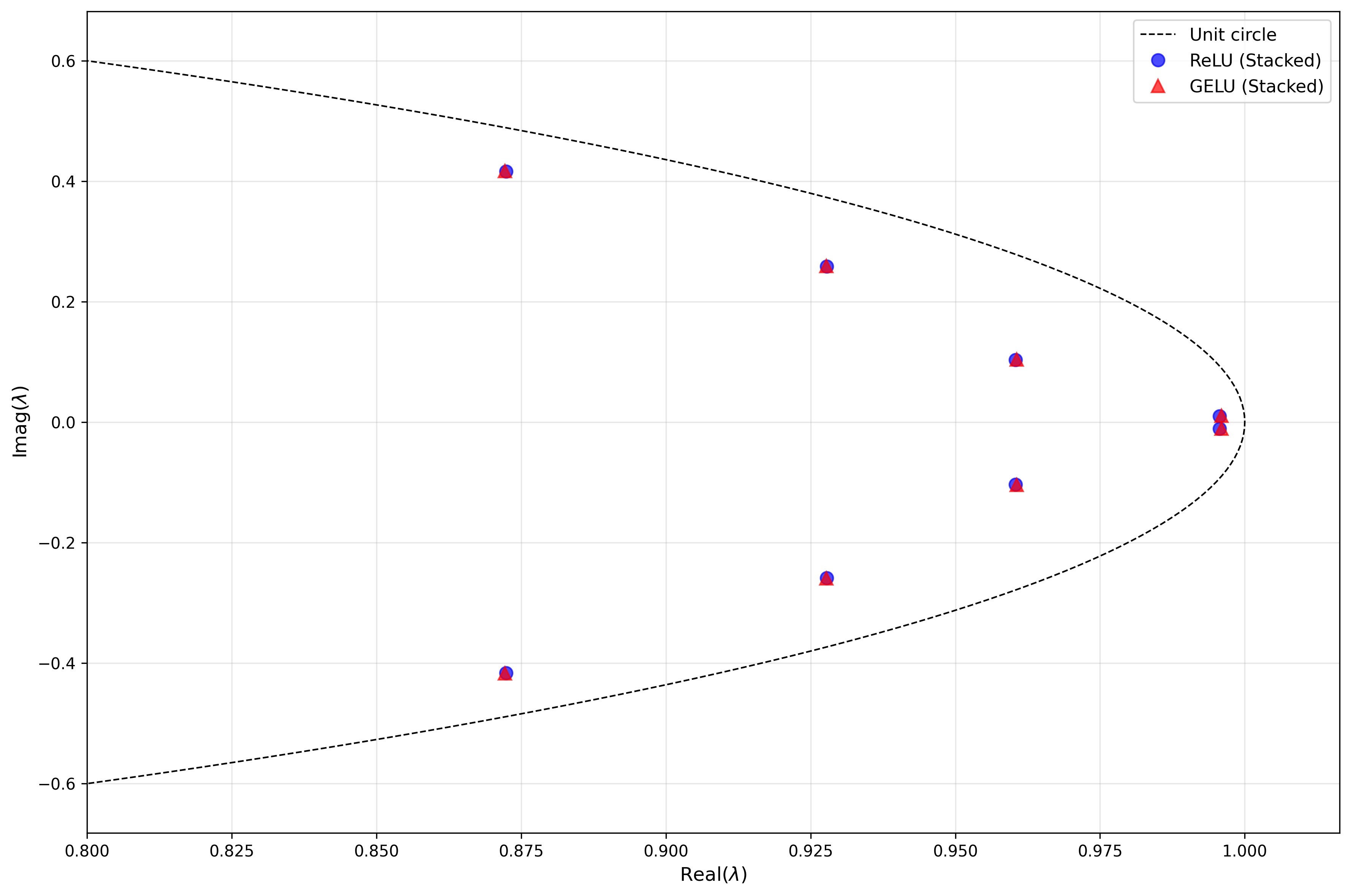}%
    \label{fig:gelu256}%
  }\hfill
  
  \caption{Training-dynamics spectral points under SGD for ReLU and GeLU activations at widths h=40 and h=256. To better visualize differences in the locations of the spectral points, we slightly warp the axis scale.
}
  \label{fig:gelu}
\end{figure*}

\paragraph{Qualitative spectral robustness checks.}
We ran three qualitative robustness checks on the FCN spectra under the shuffle control above; the corresponding raw experiment outputs are not part of the released claim index, so we report direction only and omit numerical tables.
(i) \emph{Multiplicative initialization perturbations.} For $h=40$, applying $0.1\%$ relative multiplicative perturbations elementwise to the weights and computing the Wasserstein distance between spectra over the resulting initialization pairs gives small distances under both SGD and AdamW; the shuffle control does not reject the null that the observed distances are within the shuffle baseline.
(ii) \emph{Seed variability.} For each width $h\in\{5,40,256,1024\}$ we draw $25$ independent seeds and compute pairwise spectral Wasserstein distances; within-width seed-to-seed distances are small at every tested width and again do not exceed the shuffle baseline. Across-width distances are visibly larger than within-width seed distances, indicating that in this setting width changes the spectrum more than seed choice does.
(iii) \emph{Activation choice (ReLU vs.\ GeLU).} Under SGD at $h\in\{40,256\}$, the ReLU-vs-GeLU spectral Wasserstein distance is small at $h=256$ and larger at $h=40$, but in neither case exceeds the shuffle baseline; Figure~\ref{fig:gelu} shows the corresponding spectral points.

\subsection{Transformers}

We follow the modular addition (mod-$97$) task and training codebase introduced by \citet{power2022grokkinggeneralizationoverfittingsmall} and use their open-source implementation {\url{https://github.com/openai/grok}}. The task maps pairs $(a,b)\in\{0,\dots,96\}^2$ to the label $y=(a+b)\bmod 97$. We modify only the training-set fraction, setting it to $0.4$, and keep the remaining data generation and evaluation protocol unchanged.

We consider three weight decay settings, $\mathrm{wd}\in\{0,1,2\}$, and train one model per setting under otherwise matched hyperparameters. Unless stated otherwise, all architectural choices and optimization hyperparameters follow the repository defaults. The detailed values (optimizer, learning rate, batch size, number of training steps, model dimensions, random seeds, and any learning-rate schedule) are reported in Table~\ref{tab:tr_modadd_setup}.

As described in Section~\ref{sec:method} (Eq.~\ref{eq:mu-tr}), we construct the distributional state variable from a fixed probe set of $M=100$ training examples: at training step $t$ we record the scalar correct-answer log-probability $o_{t,i}=\log p_{\theta_t}(y_i^\star\mid x_i)$ for each probe sample $i$, and $\mu_t$ is their empirical distribution. The quantile coordinate $z_t$ uses the same $19$-level grid as the FCN setting. We then compute windowed Hankel-DMD spectra and diagnostics (reconstruction residual and effective rank) on step-based segments, using the same DMD settings with FCN experiments across all weight decay values.


\begin{table}[t]
\centering
\caption{Transformer modular addition (mod $97$) experimental setup.}
\label{tab:tr_modadd_setup}
\begin{tabular}{ll}
\hline
\textbf{Component} & \textbf{Setting} \\
\hline
Task & Modular addition (mod $97$), operator ``$+$'' \\
Data format & Equation-style text: ``$a + b = c \ (\mathrm{mod}\ 97)$'' \\
Training fraction & $40\%$ (train $\approx 3764$ equations; val $\approx 5645$ equations) \\

\hline
Model & Decoder-only Transformer \\
\# layers & $n_{\mathrm{layers}}=2$ \\
\# heads & $n_{\mathrm{heads}}=4$ \\
Model width & $d_{\mathrm{model}}=128$ \\
FFN width & $d_{\mathrm{ff}}=512$ (multiplier $=4$) \\
Activation & ReLU \\
Dropout & $0.1$ \\
Positional encoding & Sinusoidal \\
Attention mask & Causal (lower-triangular) \\
Vocabulary size & $\approx 2000$ tokens \\
\hline
Optimizer & AdamW \\
Learning rate & $\max\_lr = 10^{-3}$ \\
AdamW betas & $(0.9,\,0.98)$ \\
AdamW eps & $10^{-8}$ \\
Weight decay & $\mathrm{wd}\in\{0,1,2\}$ with 10 seeds \\

Training steps & $\max\_steps=6000$ \\
Early stopping & stop after val acc $\ge 99\%$ and $1000$ additional steps \\
Batch size & $\min(512,\lceil N_{\mathrm{train}}/2\rceil)$, here $\approx 1882$ \\

\hline
Fingerprint source & log\_prob (log-probability of the correct answer) \\
Probe size & 100 samples \\
Record interval & record\_every $=2$ steps \\
Quantile grid & 19 quantile levels \\
\hline
\end{tabular}
\end{table}

\begin{figure*}[t]
  \centering
  \subfigure[wd=0]{%
    \includegraphics[width=0.32\textwidth]{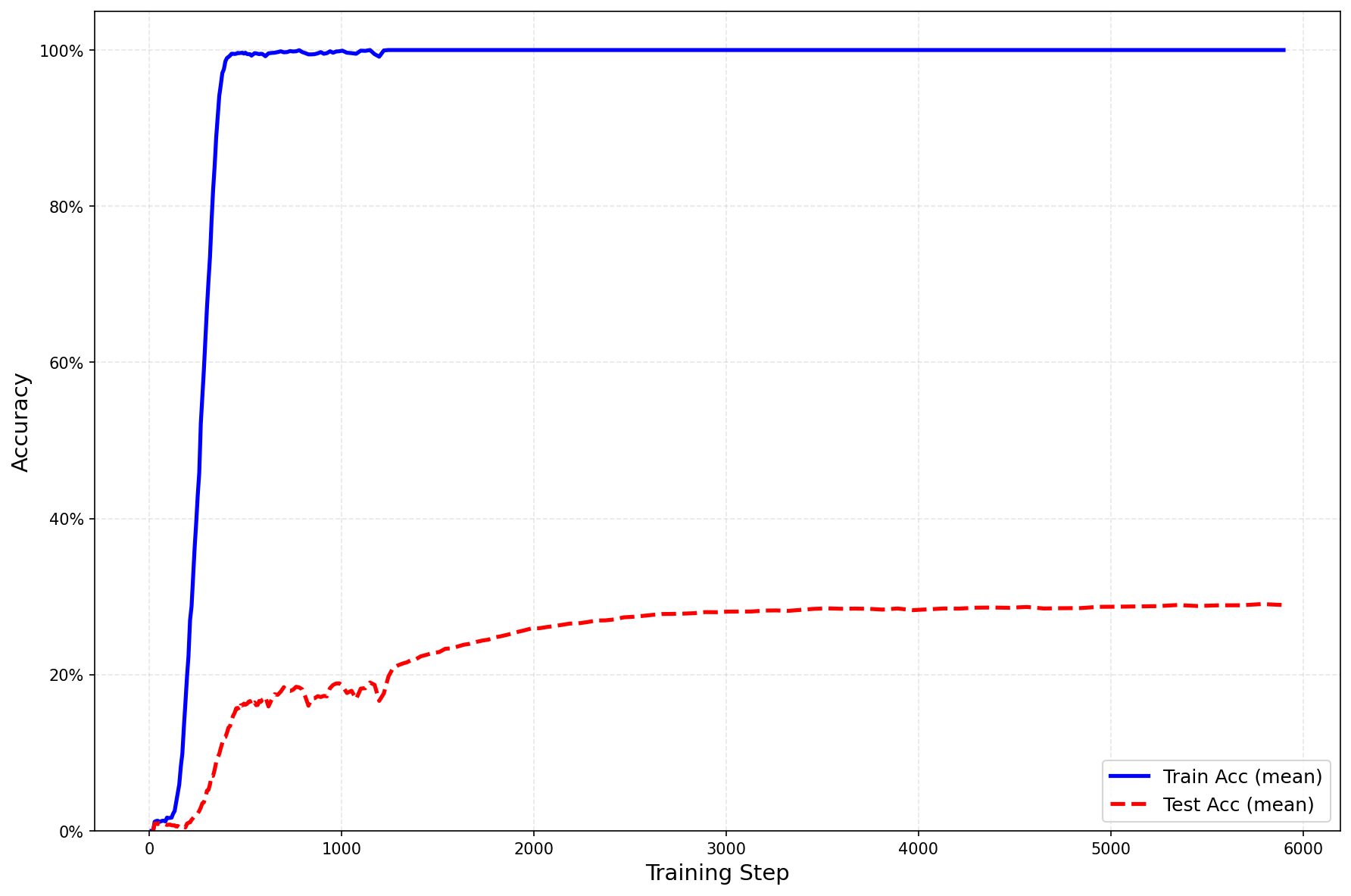}
    \label{fig:wd0acc}
  }\hfill
  \subfigure[wd=1]{%
    \includegraphics[width=0.32\textwidth]{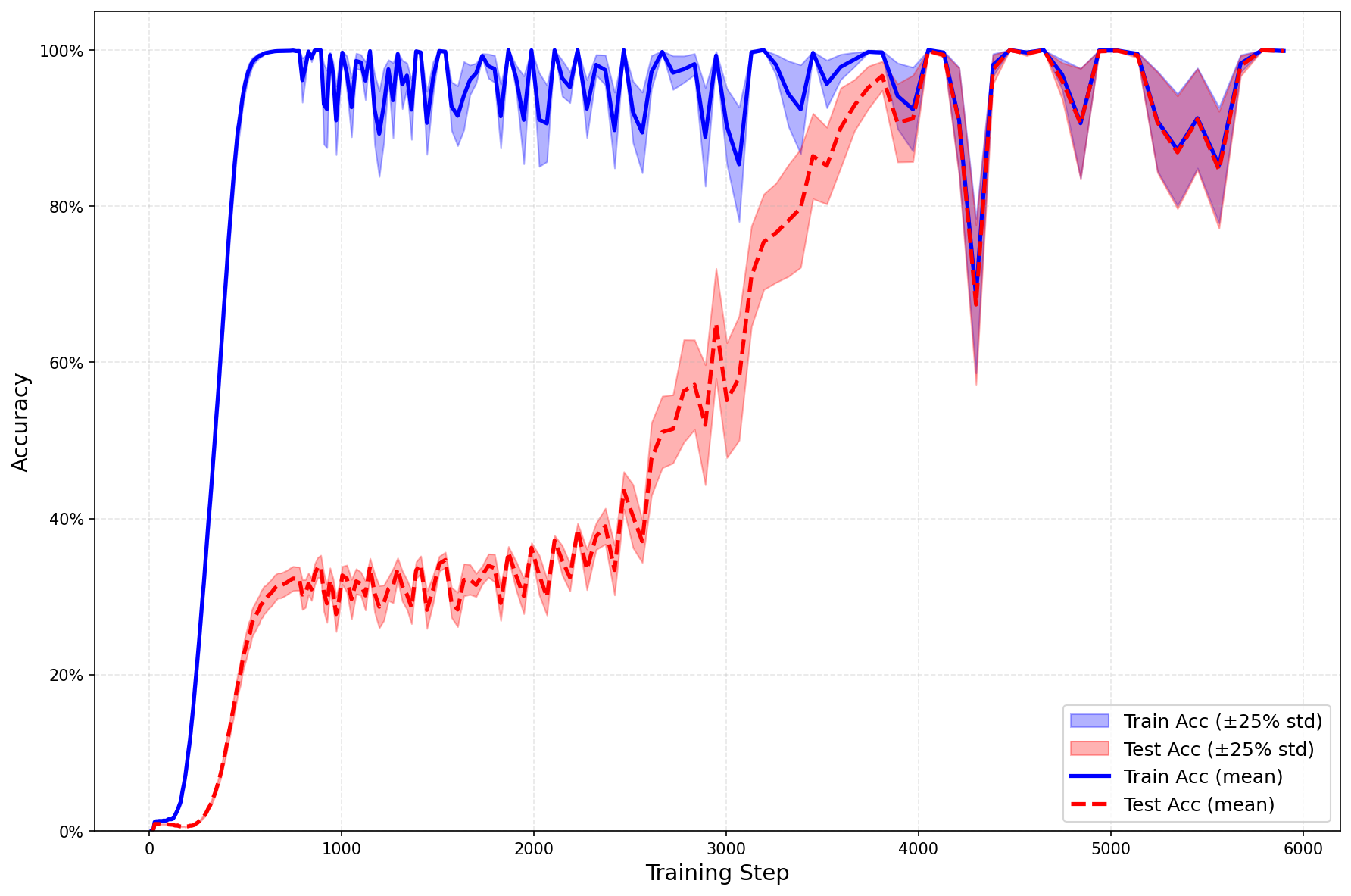}
    \label{fig:wd1acc}
  }\hfill
  \subfigure[wd=2]{%
    \includegraphics[width=0.32\textwidth]{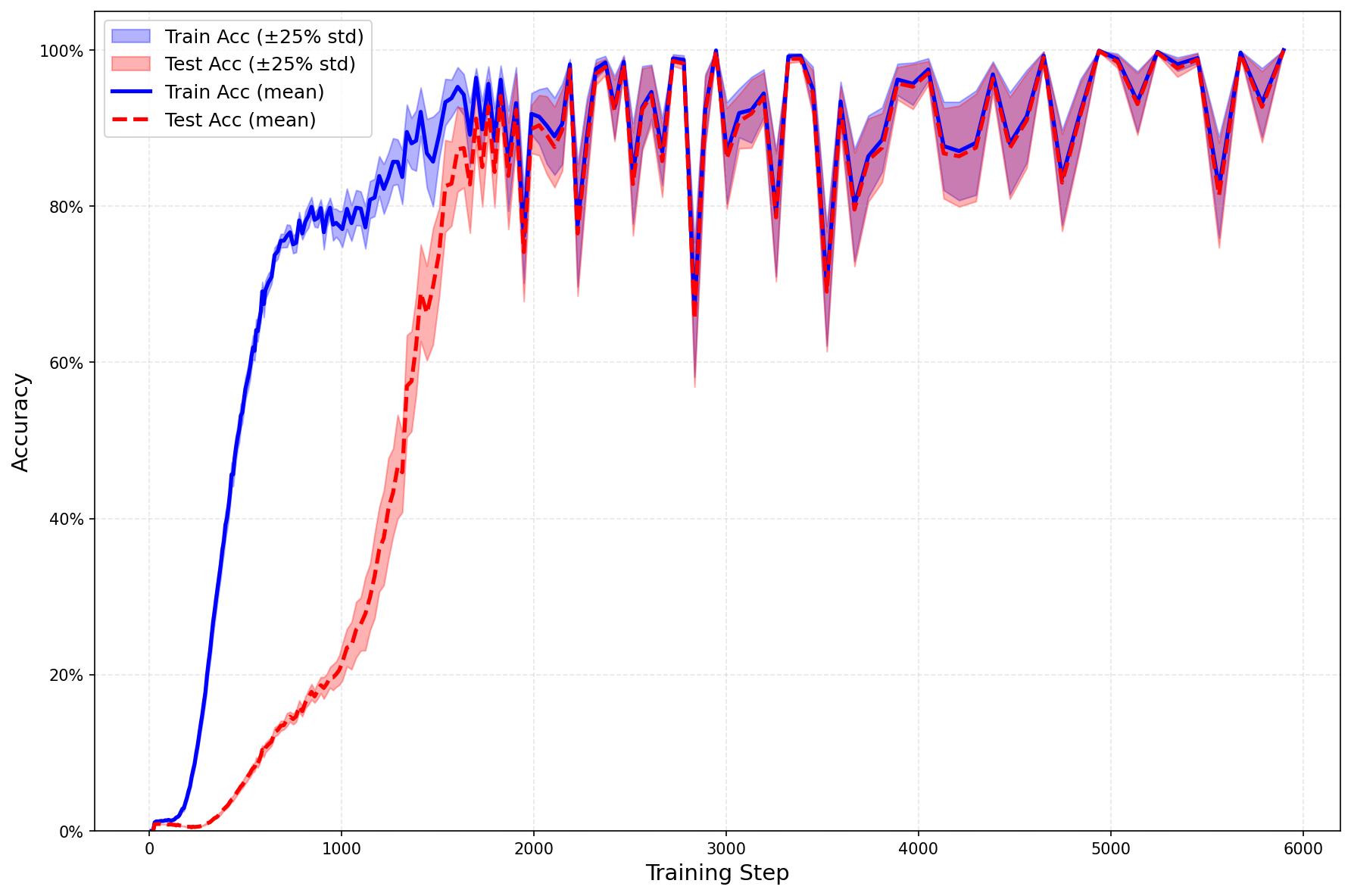}
    \label{fig:wd2acc}
  }
  \caption{(a) An example of training and test accuracy curves without weight decay. The blue solid line denotes training accuracy and the red dashed line denotes test accuracy; the run remains overfitted for an extended period. (b) Mean training and test accuracy curves with weight decay $\mathrm{wd}=1$, with a $\pm 25\%$ standard-deviation band. The blue solid line denotes training accuracy, the red dashed line denotes test accuracy, and the shaded region indicates the $\pm 25\%$ standard-deviation range; after a period of overfitting, the model undergoes rapid generalization. (c) Mean training and test accuracy curves with weight decay $\mathrm{wd}=2$, with a $\pm 25\%$ standard-deviation band. The blue solid line denotes training accuracy, the red dashed line denotes test accuracy, and the shaded region indicates the $\pm 25\%$ standard-deviation range; generalization occurs without a pronounced overfitting period.
}
  \label{fig:wdacc}
\end{figure*}

\begin{figure*}[t]
  \centering
  \subfigure[stage 1]{%
    \includegraphics[width=0.30\textwidth]{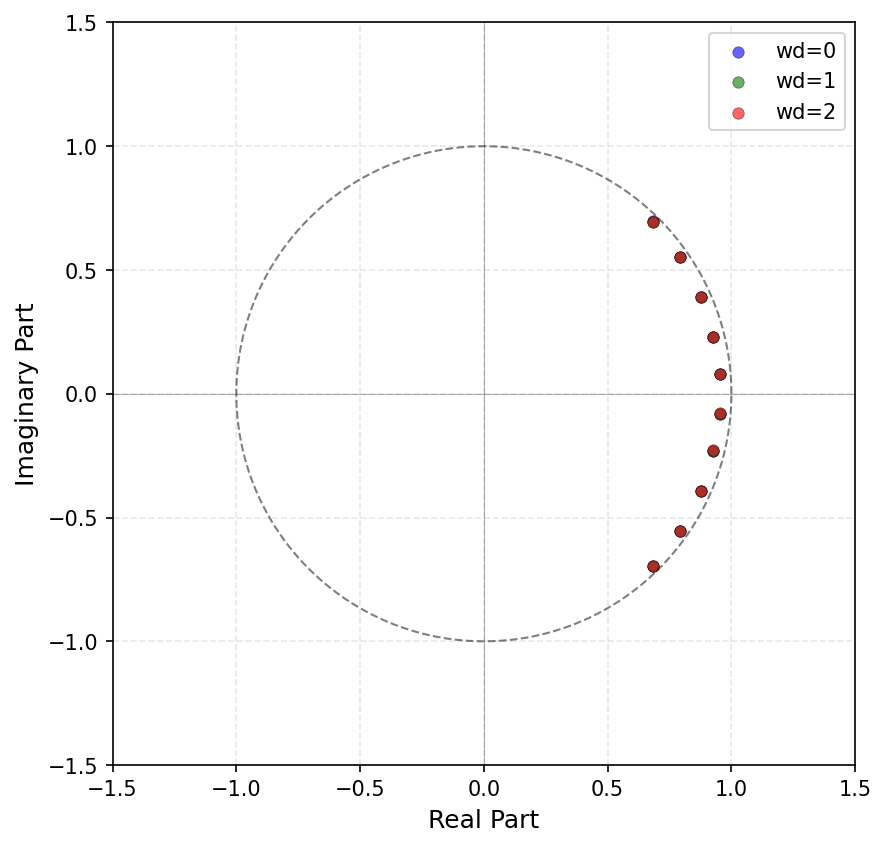}
  }
  \subfigure[stage 2]{%
    \includegraphics[width=0.30\textwidth]{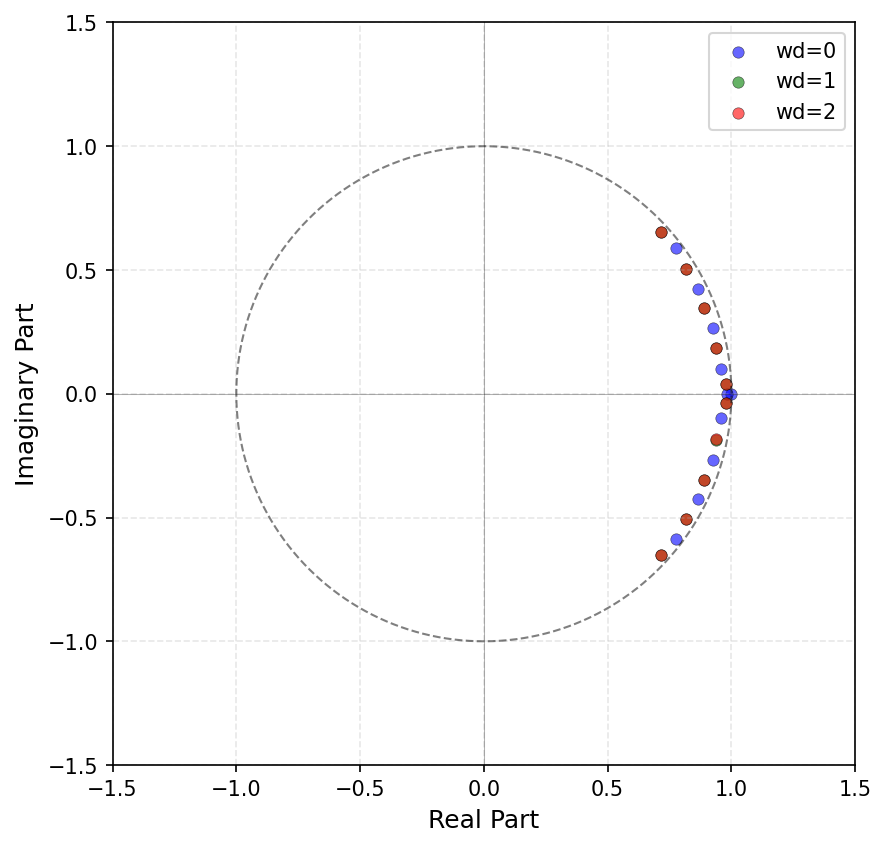}
  }
  \subfigure[stage 3]{%
    \includegraphics[width=0.30\textwidth]{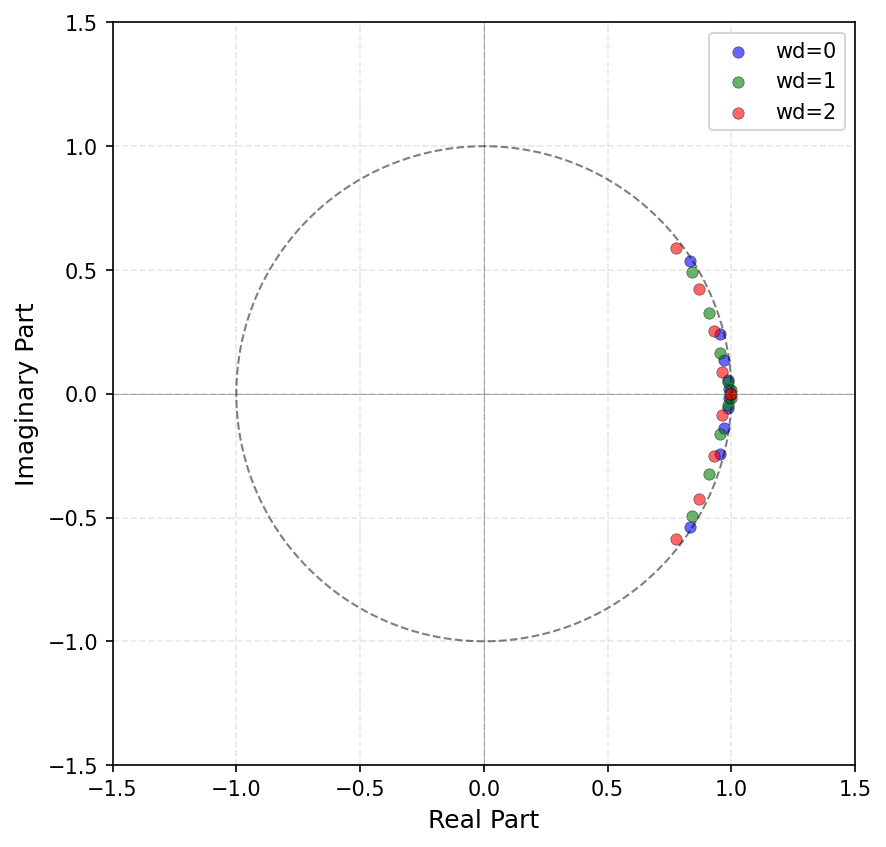}
  }\\[-1mm]

  \subfigure[stage 4]{%
    \includegraphics[width=0.30\textwidth]{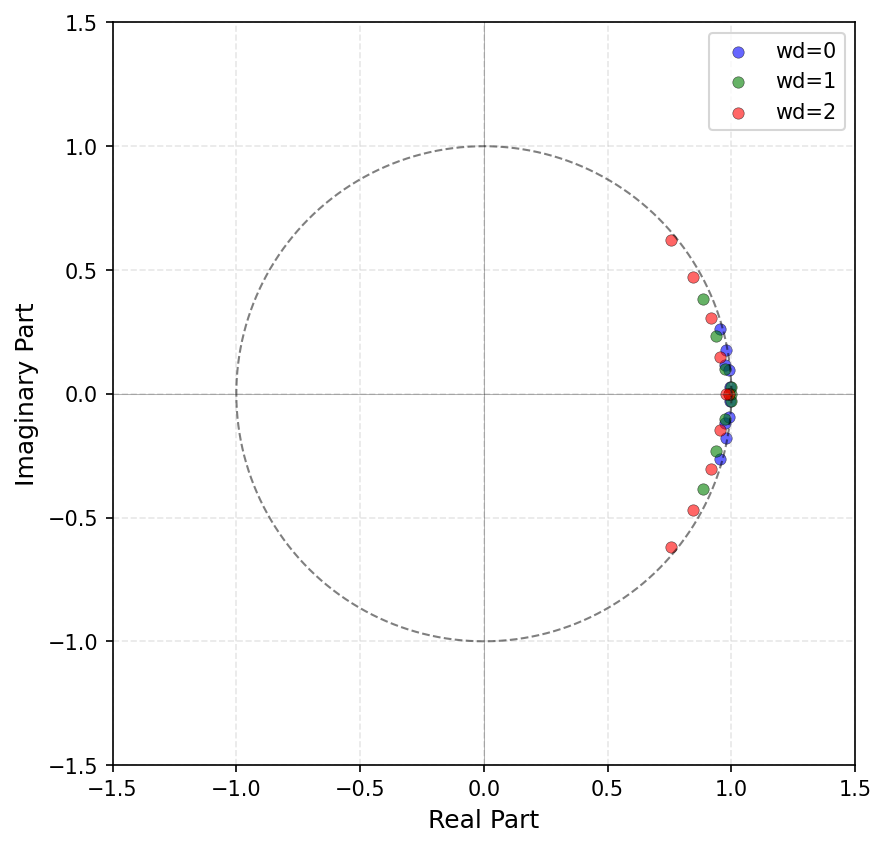}
  }
  \subfigure[stage 5]{%
    \includegraphics[width=0.30\textwidth]{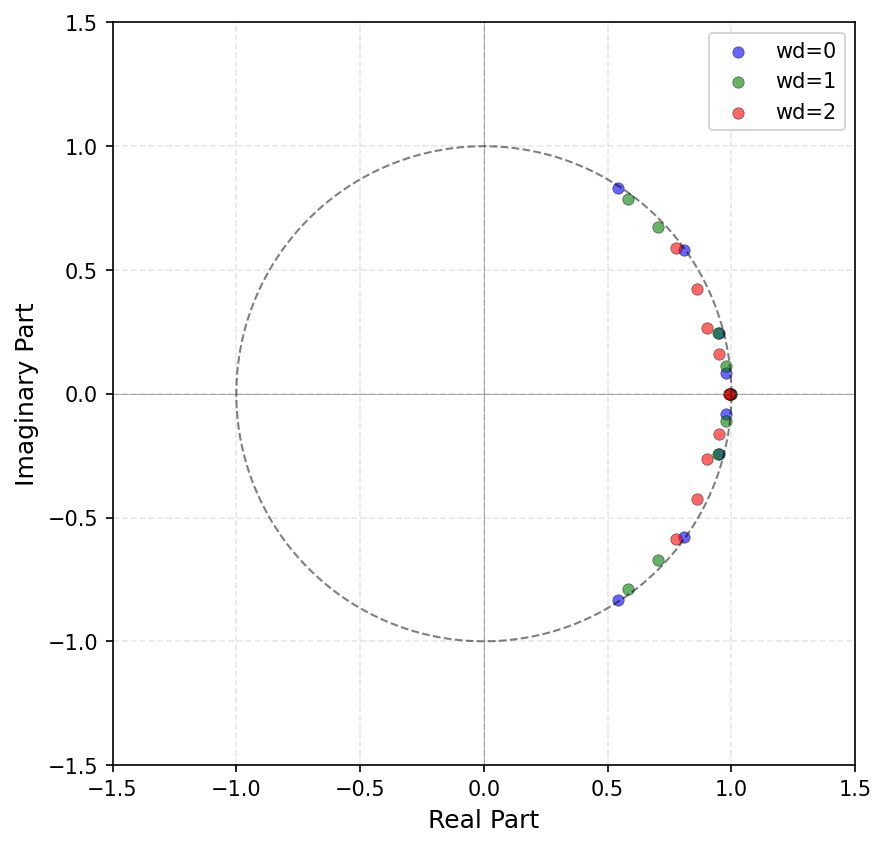}
  }
  \subfigure[stage 6]{%
    \includegraphics[width=0.30\textwidth]{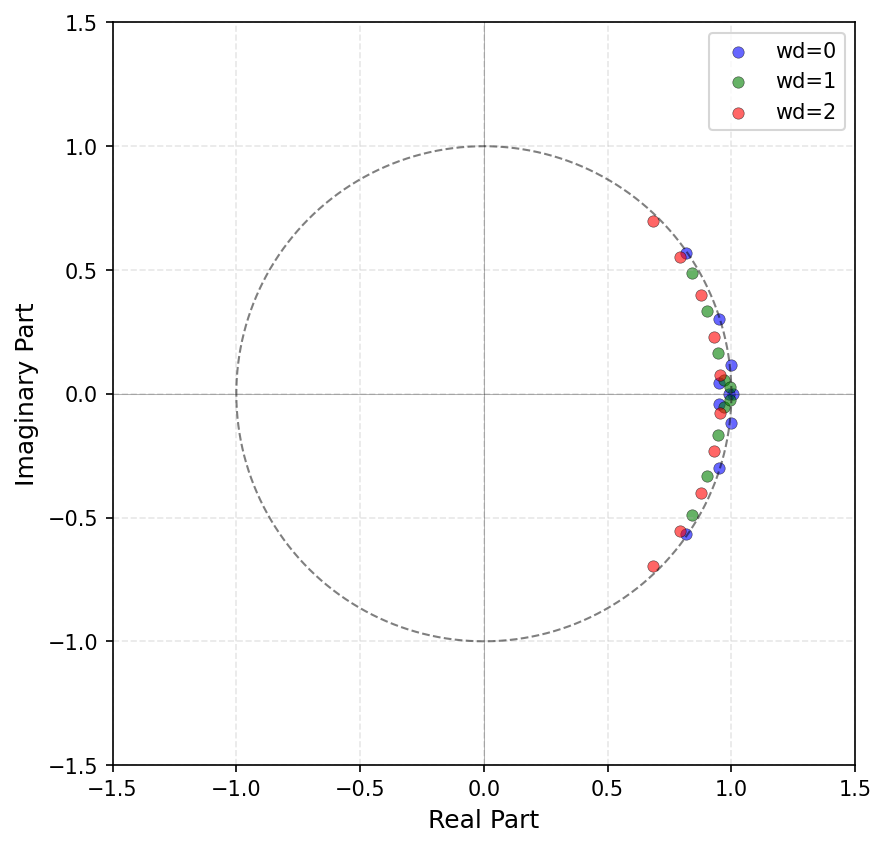}
  }\\[-1mm]

  \subfigure[stage 7]{%
    \includegraphics[width=0.30\textwidth]{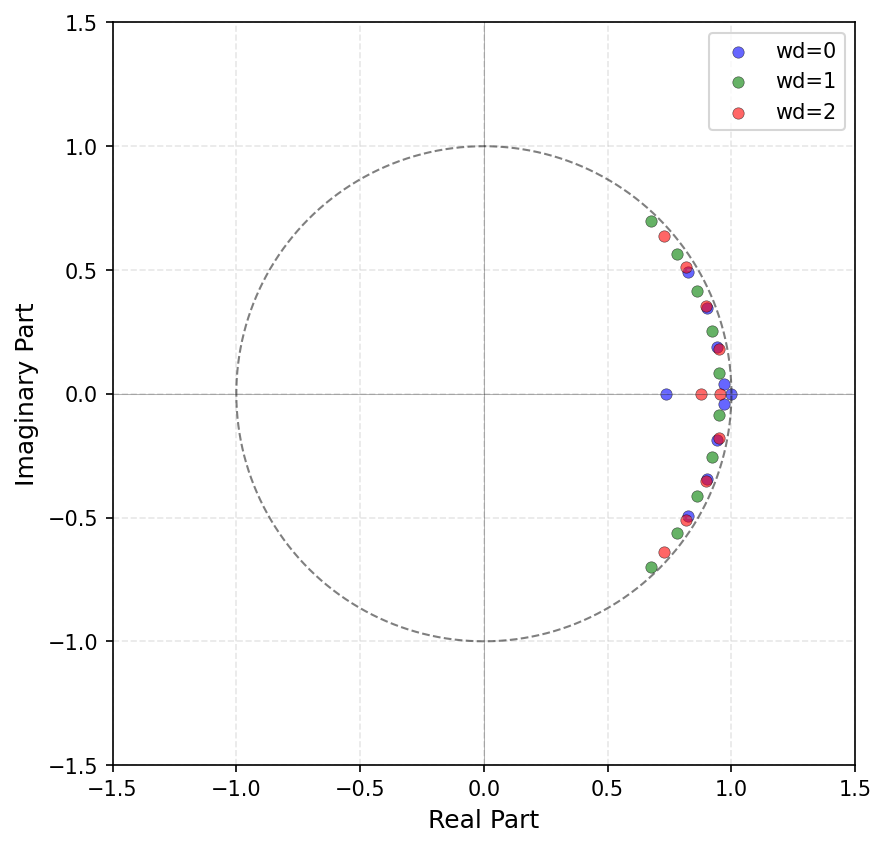}
  }
  \subfigure[stage 8]{%
    \includegraphics[width=0.30\textwidth]{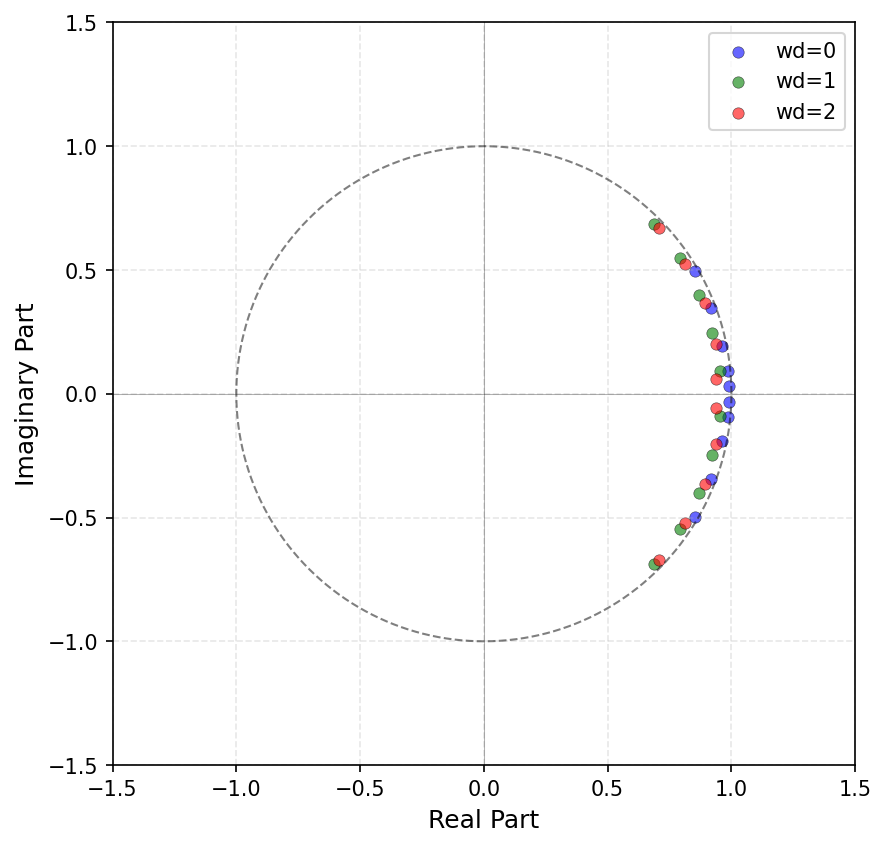}
  }
  \subfigure[stage 9]{%
    \includegraphics[width=0.30\textwidth]{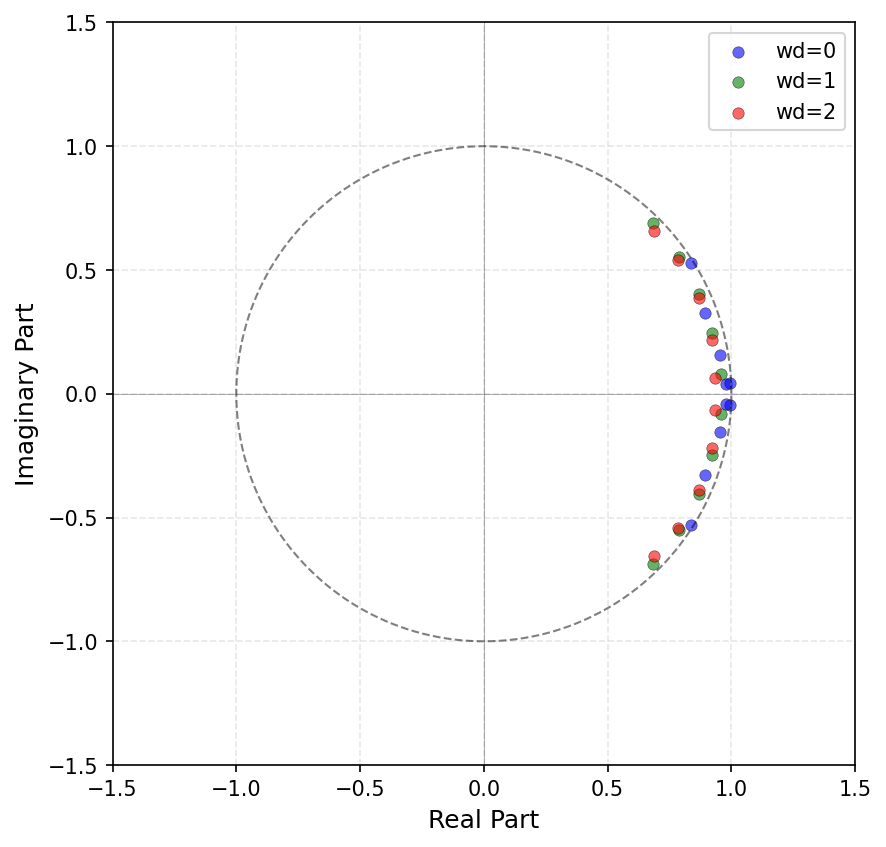}
  }\\[-1mm]

  \subfigure[stage 10]{%
    \includegraphics[width=0.30\textwidth]{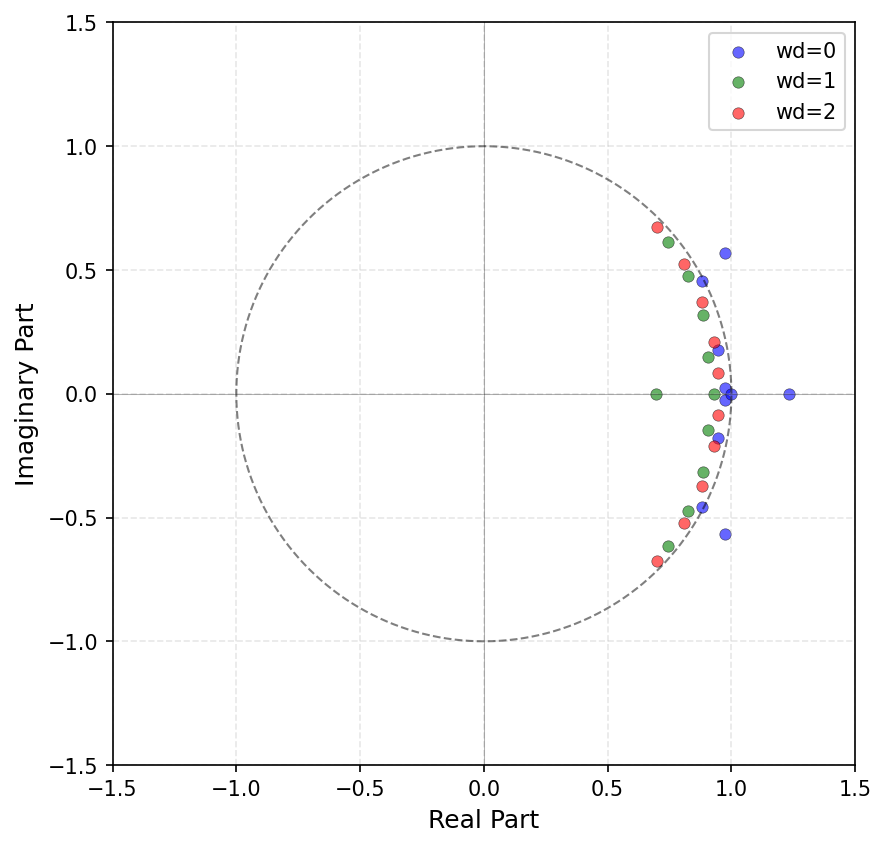}
  }
  \subfigure[stage 11]{%
    \includegraphics[width=0.30\textwidth]{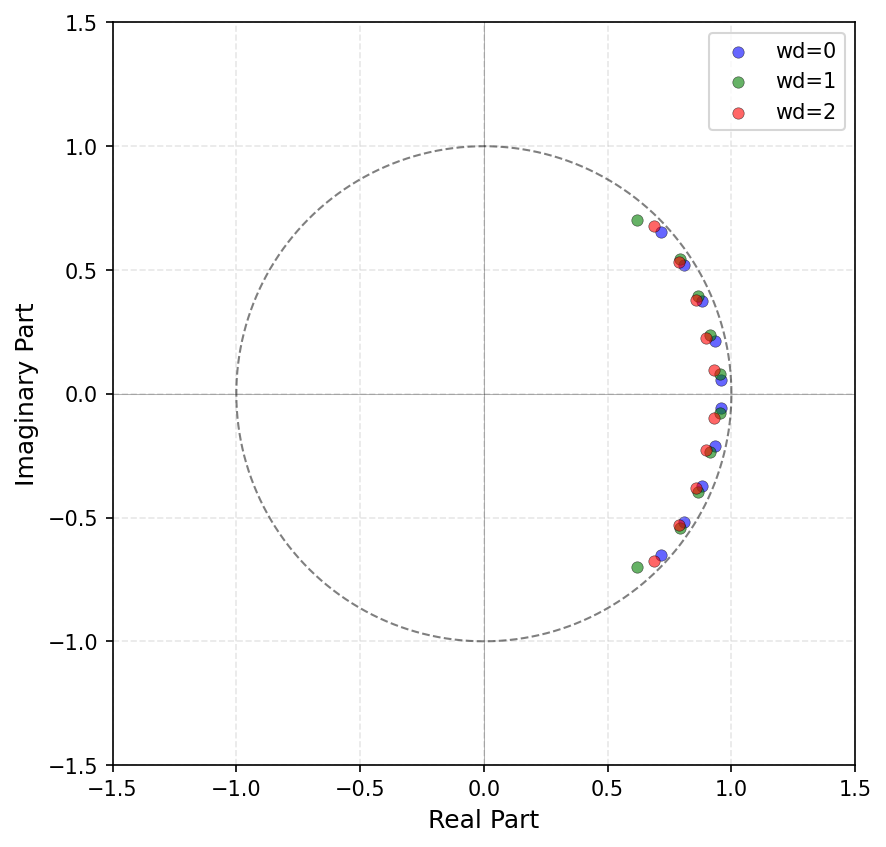}
  }
  \subfigure[stage 12]{%
    \includegraphics[width=0.30\textwidth]{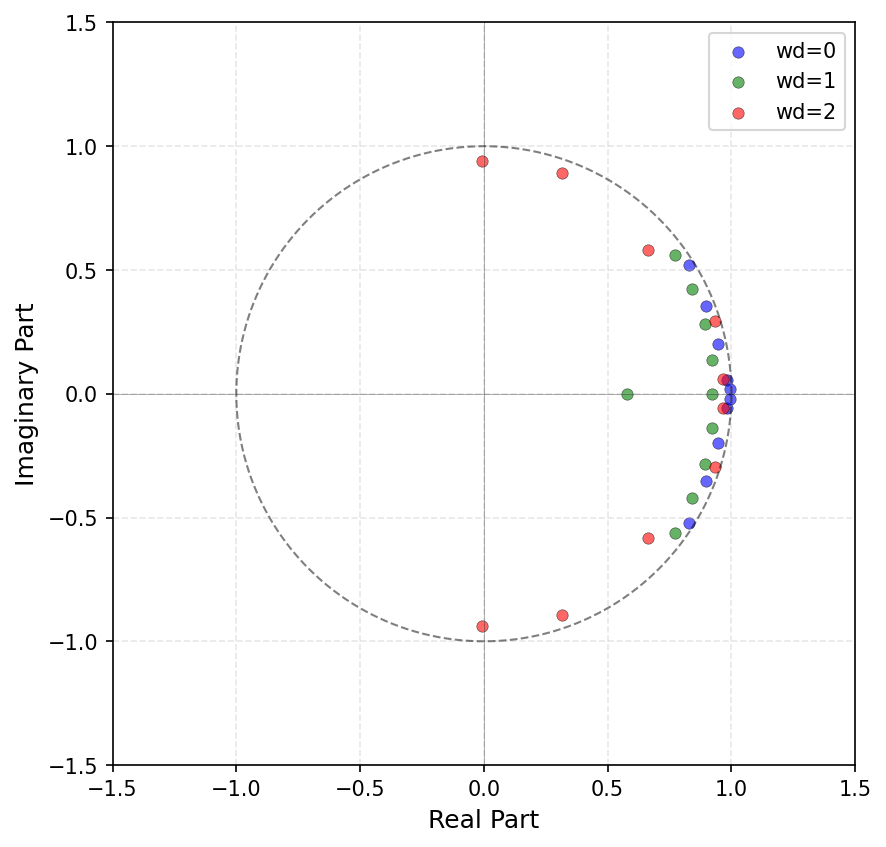}
  }

  \caption{Local Koopman spectra across successive training stages.
  Each panel corresponds to a distinct step interval.}
  \label{fig:stages}
\end{figure*}

Across our three weight decay settings, the no-weight-decay runs exhibit prolonged overfitting followed by slow generalization. Even with the training fraction set to 40\%, some seeds do not generalize until beyond $5\times 10^{5}$ steps. In contrast, under our experimental setup, all runs with weight decay generalize within 6000 steps, as shown in Figure~\ref{fig:wdacc}. We therefore analyze spectral features only over the training trajectory from steps 1 to 6000: we record the log-probability every two steps and define stages by non-overlapping 500-step windows. The complete spectral plots are provided in Figure~\ref{fig:stages}.

\section{Extended Weight Decay Robustness}\label{app:wd-robust}

To address concerns regarding the sensitivity of our Transformer results to the choice of weight decay, we expand the original sweep from three values ($\mathrm{wd}\in\{0,1,2\}$) to five values ($\mathrm{wd}\in\{0,0.25,0.5,1,2\}$) with four random seeds per setting, yielding 20 independent runs. All other hyperparameters remain identical to Table~\ref{tab:tr_modadd_setup}.

\begin{figure}[h]
  \centering
  \includegraphics[width=0.75\linewidth]{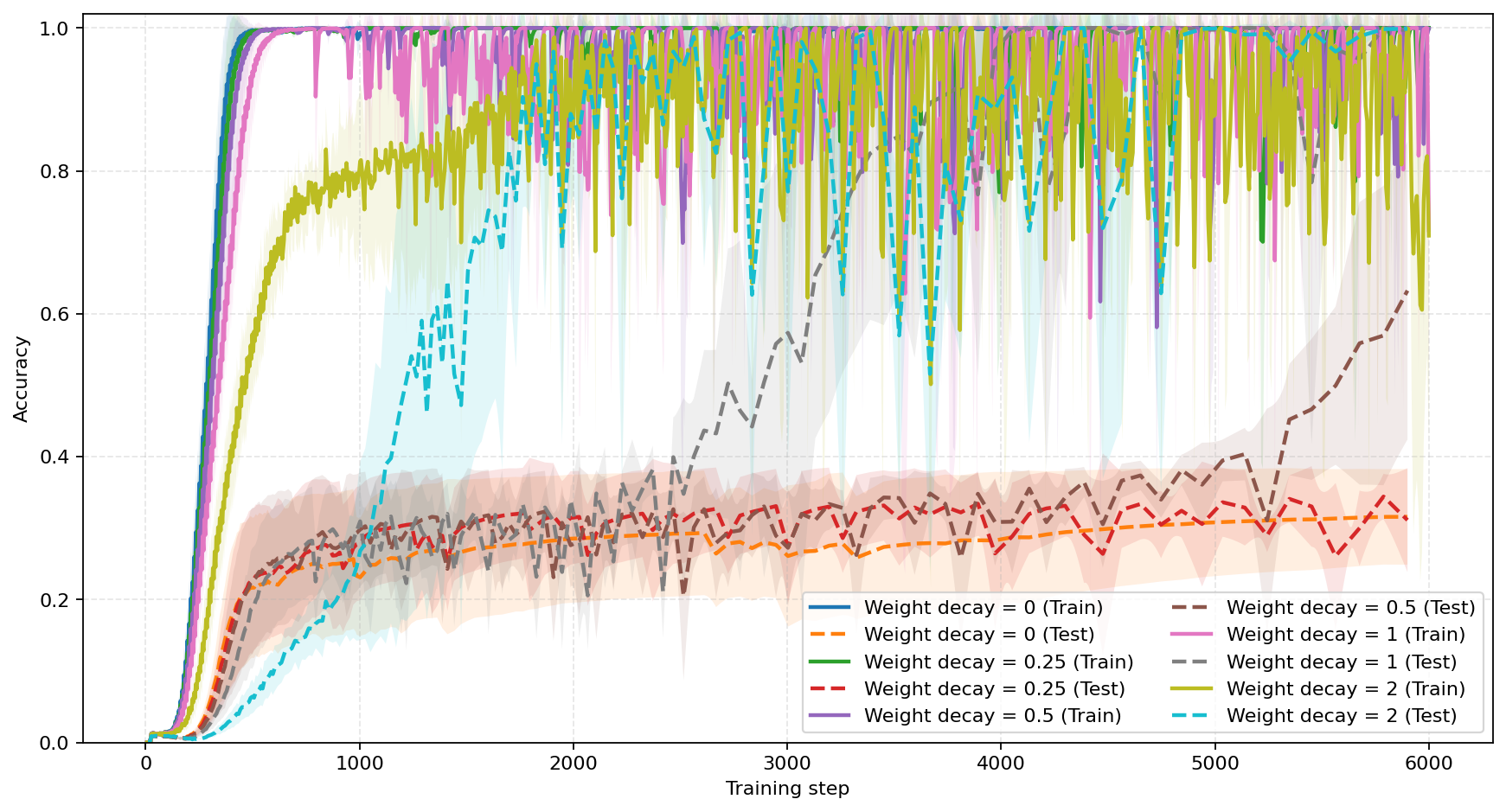}
  \caption{Test accuracy trajectories across five weight decay values (4 seeds each), with $\pm1$ std bands.}
  \label{fig:reb-acc-bands}
\end{figure}

Figure~\ref{fig:reb-acc-bands} shows the mean test accuracy trajectories with $\pm1$ standard-deviation bands. The qualitative ordering between weight decay strength and grokking onset timing observed in the original experiments remains visible across the expanded sweep: among runs that grok, higher weight decay tends to be associated with earlier generalization onset in this sweep.

\begin{figure}[h]
  \centering
  \subfigure[]{%
    \includegraphics[width=0.48\linewidth]{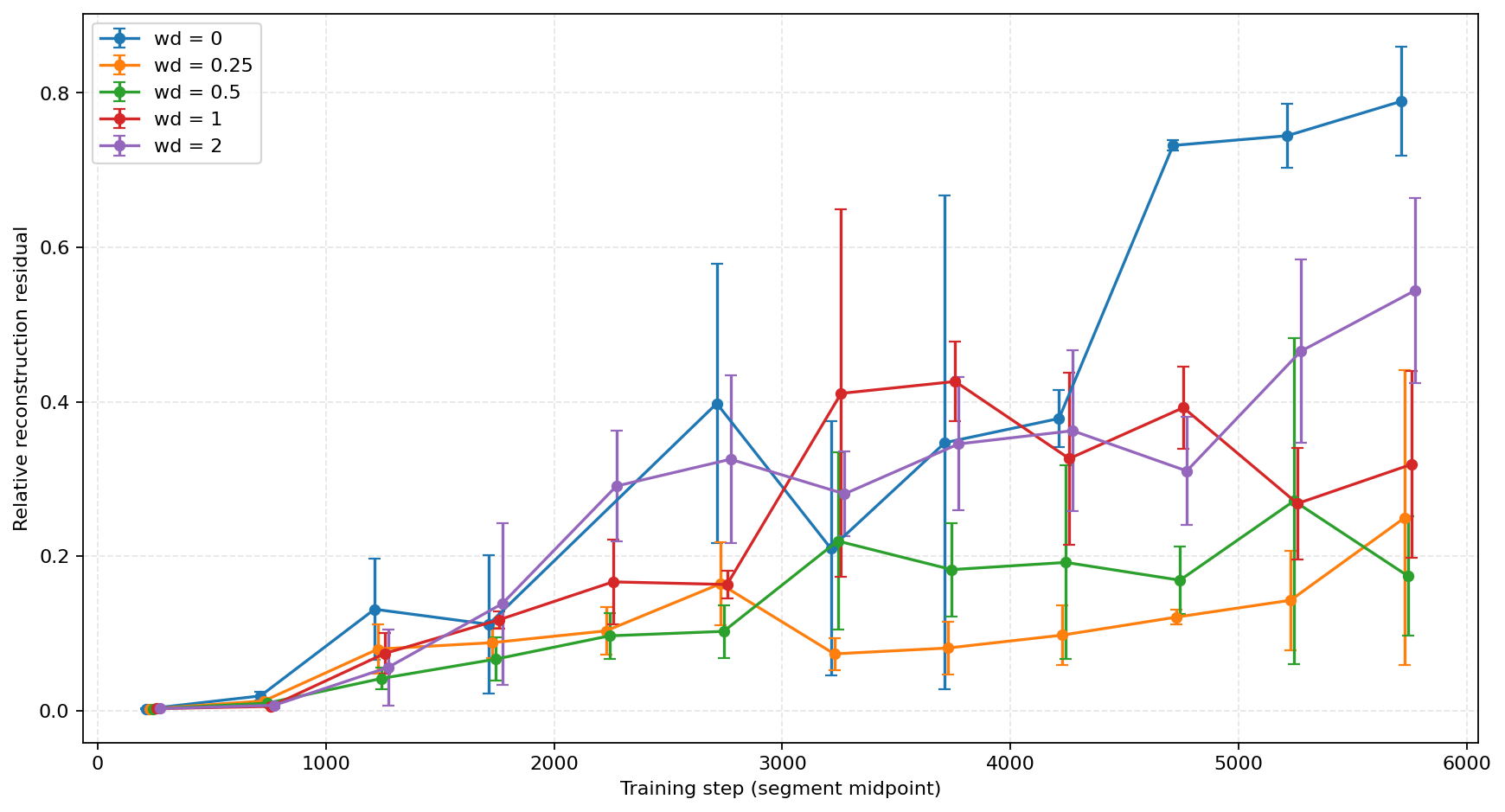}%
    \label{fig:reb-residual-bands}%
  }\hfill
  \subfigure[]{%
    \includegraphics[width=0.48\linewidth]{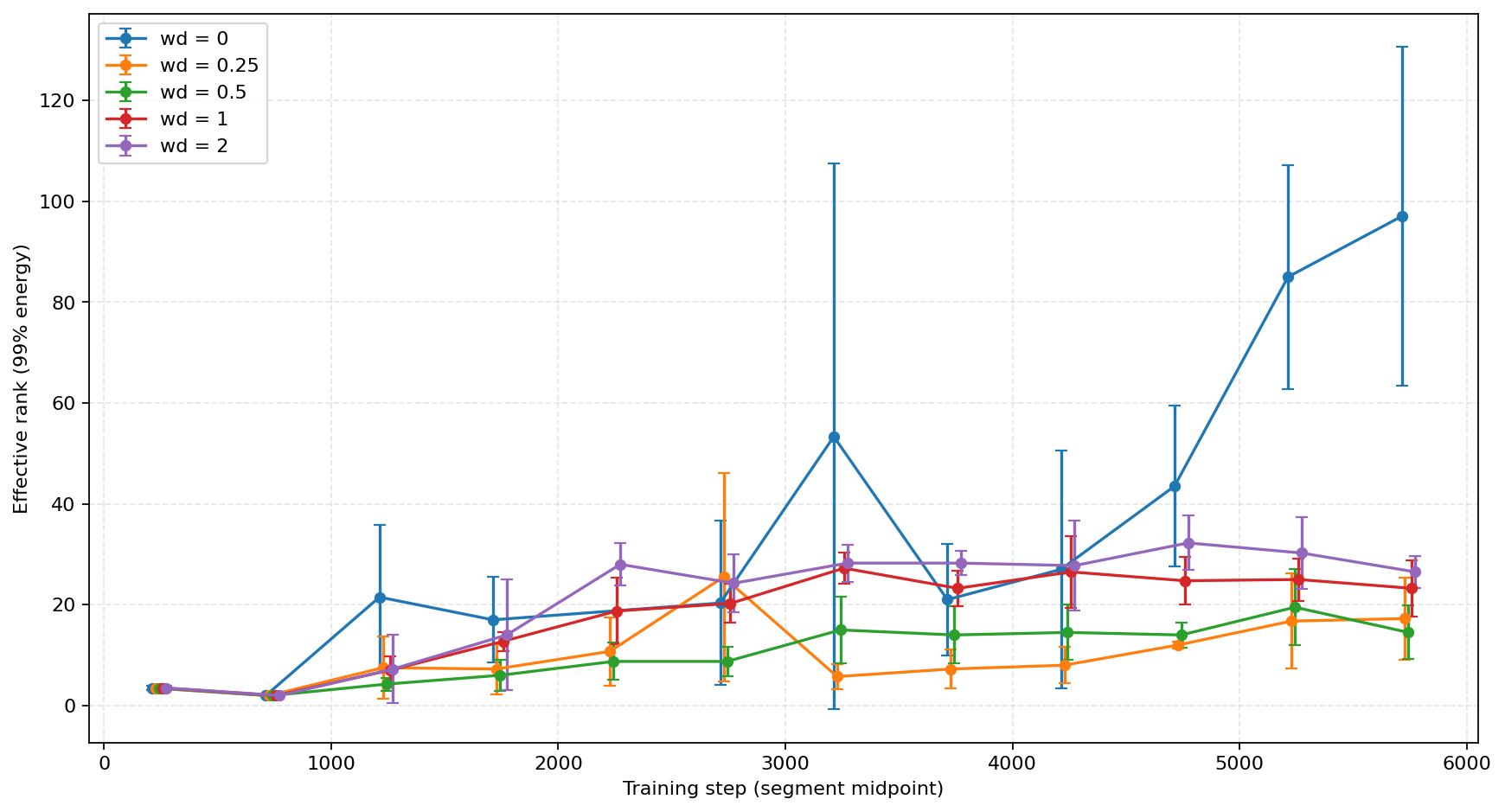}%
    \label{fig:reb-rank-bands}%
  }
  \caption{Reconstruction residual and effective rank over training for five weight decay values. Left: the reconstruction residual tracks transition windows and spikes near generalization for grokking runs. Right: the effective rank remains comparatively stable away from these transition windows.}
  \label{fig:reb-rr-rank}
\end{figure}

\paragraph{Alignment between diagnostics and accuracy.}
Figures~\ref{fig:reb-residual-bands} and~\ref{fig:reb-rank-bands} display the reconstruction residual and effective rank curves, respectively. Both diagnostics track the accuracy trajectories, with the residual peak roughly coinciding with the onset of generalization for grokking runs ($\mathrm{wd}\ge 1$).

\begin{figure}[h]
  \centering
  \subfigure[]{%
    \includegraphics[width=0.48\linewidth]{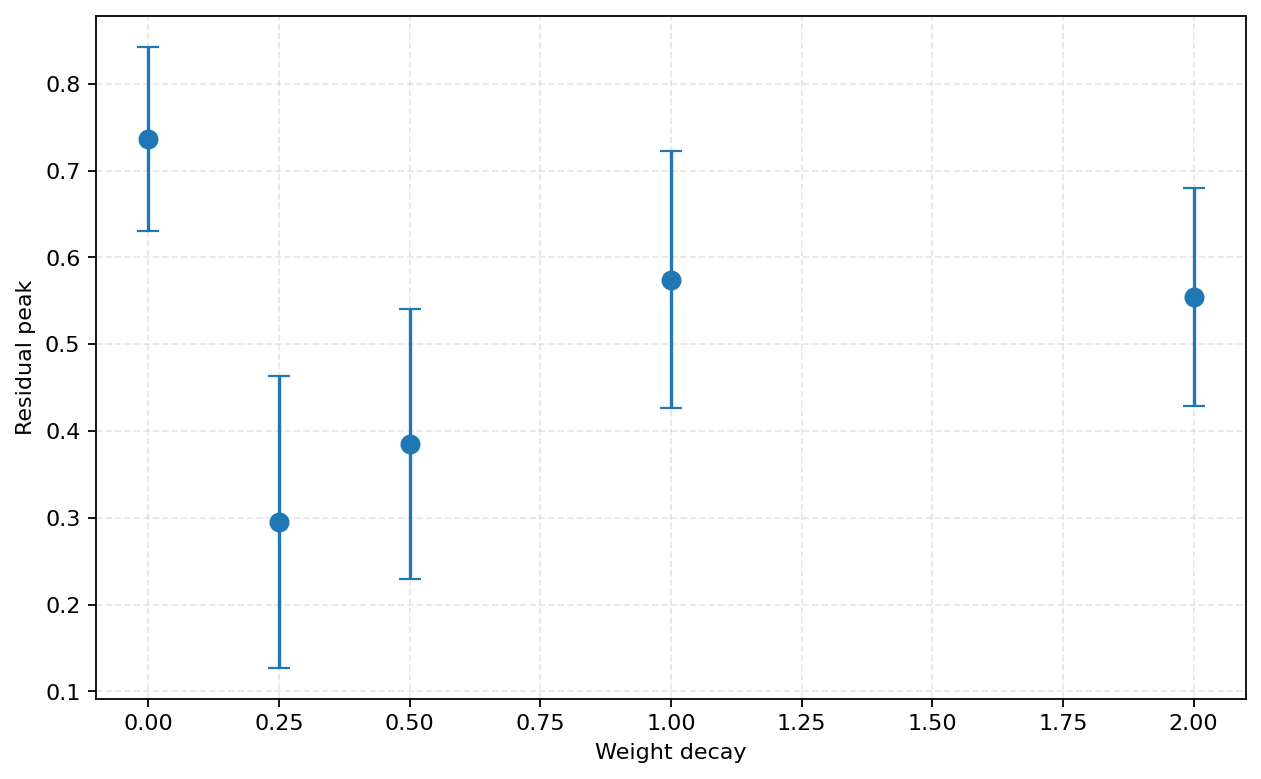}%
    \label{fig:reb-peak-wd}%
  }\hfill
  \subfigure[]{%
    \includegraphics[width=0.48\linewidth]{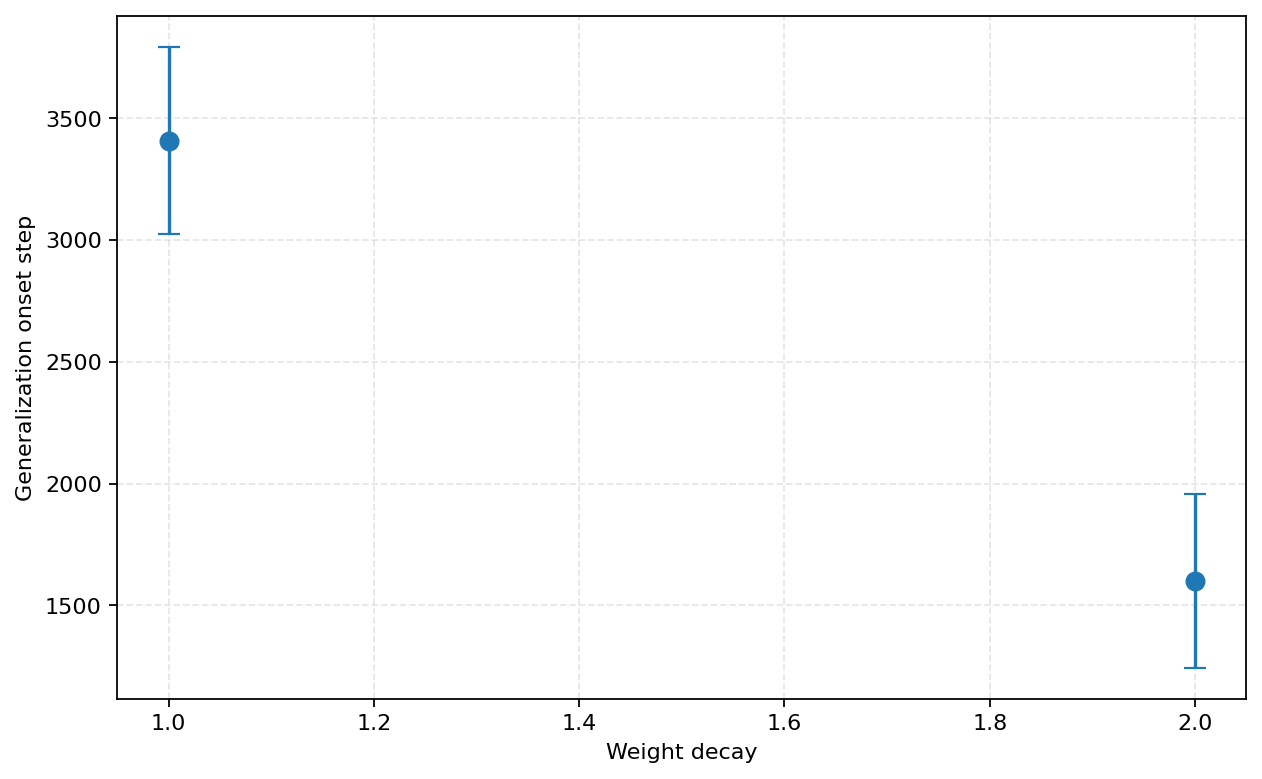}%
    \label{fig:reb-onset-wd}%
  }
  \caption{Peak reconstruction residual and grokking onset step across weight decay values. Left: peak residual varies smoothly with regularization strength. Right: grokking onset shifts earlier as weight decay increases among runs that generalize.}
  \label{fig:reb-peak-onset}
\end{figure}

\paragraph{Peak residual and onset.}
Figure~\ref{fig:reb-peak-onset} summarizes the peak residual and grokking onset step as a function of weight decay across the expanded range; both are descriptive summaries of the runs in this sweep, not a calibrated trend extrapolated to other regularization regimes.

\begin{figure}[h]
  \centering
  \includegraphics[width=0.75\linewidth]{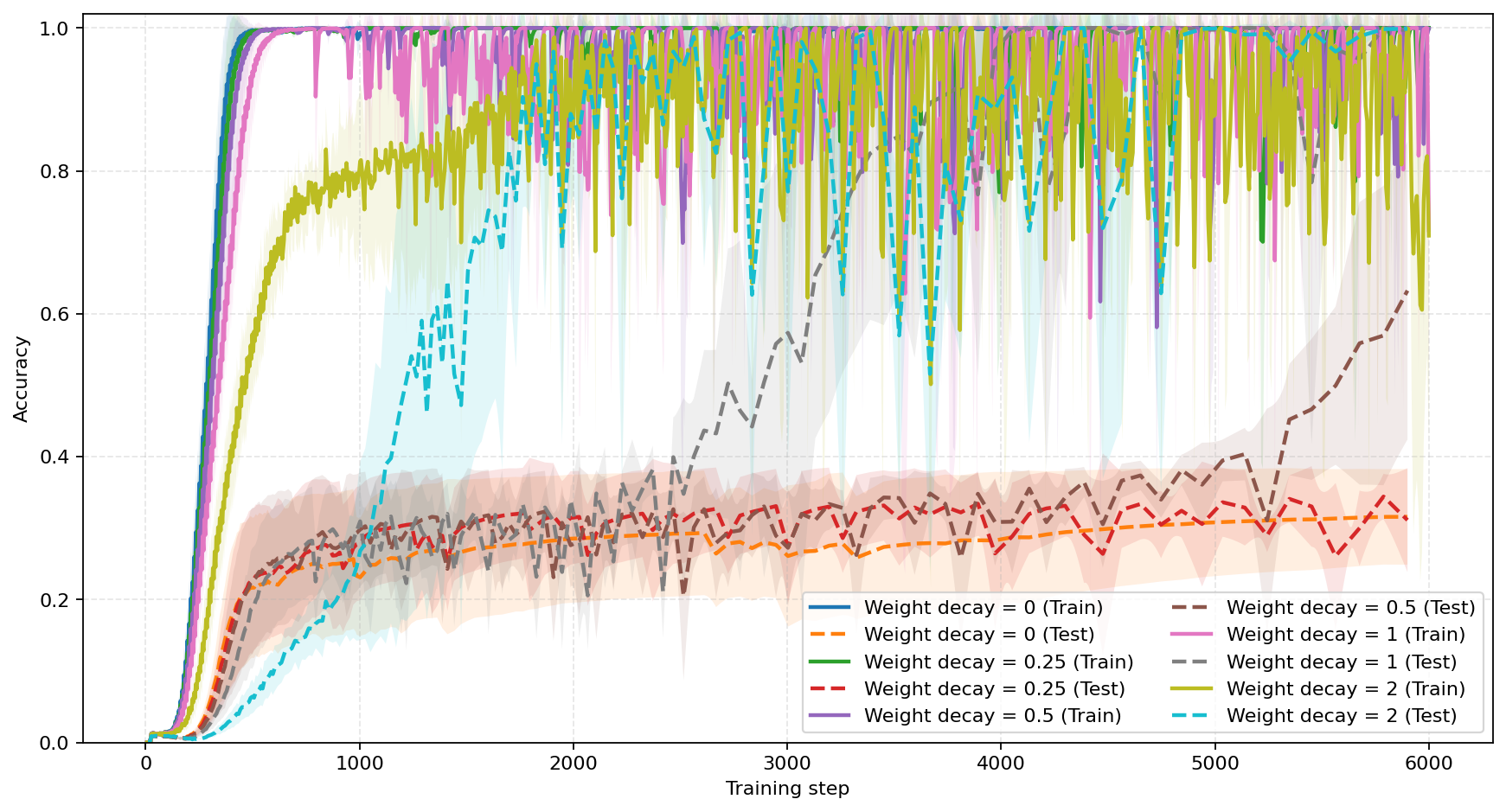}
  \caption{Train--test accuracy overlay across weight decay settings.}
  \label{fig:reb-consistency}
\end{figure}

\paragraph{Train--test consistency.}
Figure~\ref{fig:reb-consistency} overlays train and test accuracy for each weight decay value, showing that the distributional diagnostics track dynamics visible in both splits in the runs we evaluated.

\section{Architecture Ablation}\label{app:arch-ablation}

A natural concern is whether the observed patterns depend on the specific Transformer configuration ($d_{\mathrm{model}}{=}128$, $n_{\mathrm{layers}}{=}2$, $n_{\mathrm{heads}}{=}4$). We test two additional configurations while keeping the task, data split, and analysis pipeline unchanged:
\begin{itemize}
  \item \textbf{Small/shallow:} $d_{\mathrm{model}}{=}64$, $n_{\mathrm{layers}}{=}1$, $n_{\mathrm{heads}}{=}4$
  \item \textbf{Large/deep:} $d_{\mathrm{model}}{=}256$, $n_{\mathrm{layers}}{=}3$, $n_{\mathrm{heads}}{=}8$
\end{itemize}
Each variant is trained with $\mathrm{wd}\in\{0,1,2\}$ and 2 seeds.

Table~\ref{tab:arch-ablation} reports the aggregated results. The baseline observation that higher weight decay promotes generalization in this task is preserved across the three tested scales; the residual peak co-occurs with the generalization regime in all three scales, but its amplitude shifts substantially with capacity (small $\sim 0.24$, baseline $\sim 0.68$, large $\sim 0.84$ at $\mathrm{wd}=1$). At $\mathrm{wd}=2$ the small/shallow variant fails to generalize on this run pool, and the large/deep variant under-generalizes at both $\mathrm{wd}=1$ and $\mathrm{wd}=2$, indicating that the strength of the wd--generalization relationship is architecture-dependent.

\begin{table}[h]
\centering
\caption{Architecture ablation summary (modular addition, 6000 steps). Aggregates over the seeds in each cell. Onset is reported for cells where the mean test accuracy exceeds the $99\%$ criterion; ``--'' means the cell does not reach generalization on this pool.}
\label{tab:arch-ablation}
\small
\begin{tabular}{llcccc}
\toprule
Architecture & wd & Seeds & Test Acc & Peak RR & Onset \\
\midrule
baseline ($128{\times}2$) & 0 & 8 & 0.325 & 0.780 & -- \\
baseline ($128{\times}2$) & 1 & 8 & 0.999 & 0.625 & 3398 \\
baseline ($128{\times}2$) & 2 & 8 & 0.997 & 0.584 & 1528 \\
\midrule
small ($64{\times}1$) & 0 & 2 & 0.340 & 0.376 & -- \\
small ($64{\times}1$) & 1 & 5 & 0.805 & 0.238 & 3991 \\
small ($64{\times}1$) & 2 & 5 & 0.167 & 0.149 & -- \\
\midrule
large ($256{\times}3$) & 0 & 2 & 0.224 & 0.815 & -- \\
large ($256{\times}3$) & 1 & 5 & 0.610 & 0.837 & 2485 \\
large ($256{\times}3$) & 2 & 5 & 0.529 & 0.835 & 1389 \\
\bottomrule
\end{tabular}
\end{table}

\begin{figure}[h]
  \centering
  \subfigure[]{%
    \includegraphics[width=0.48\linewidth]{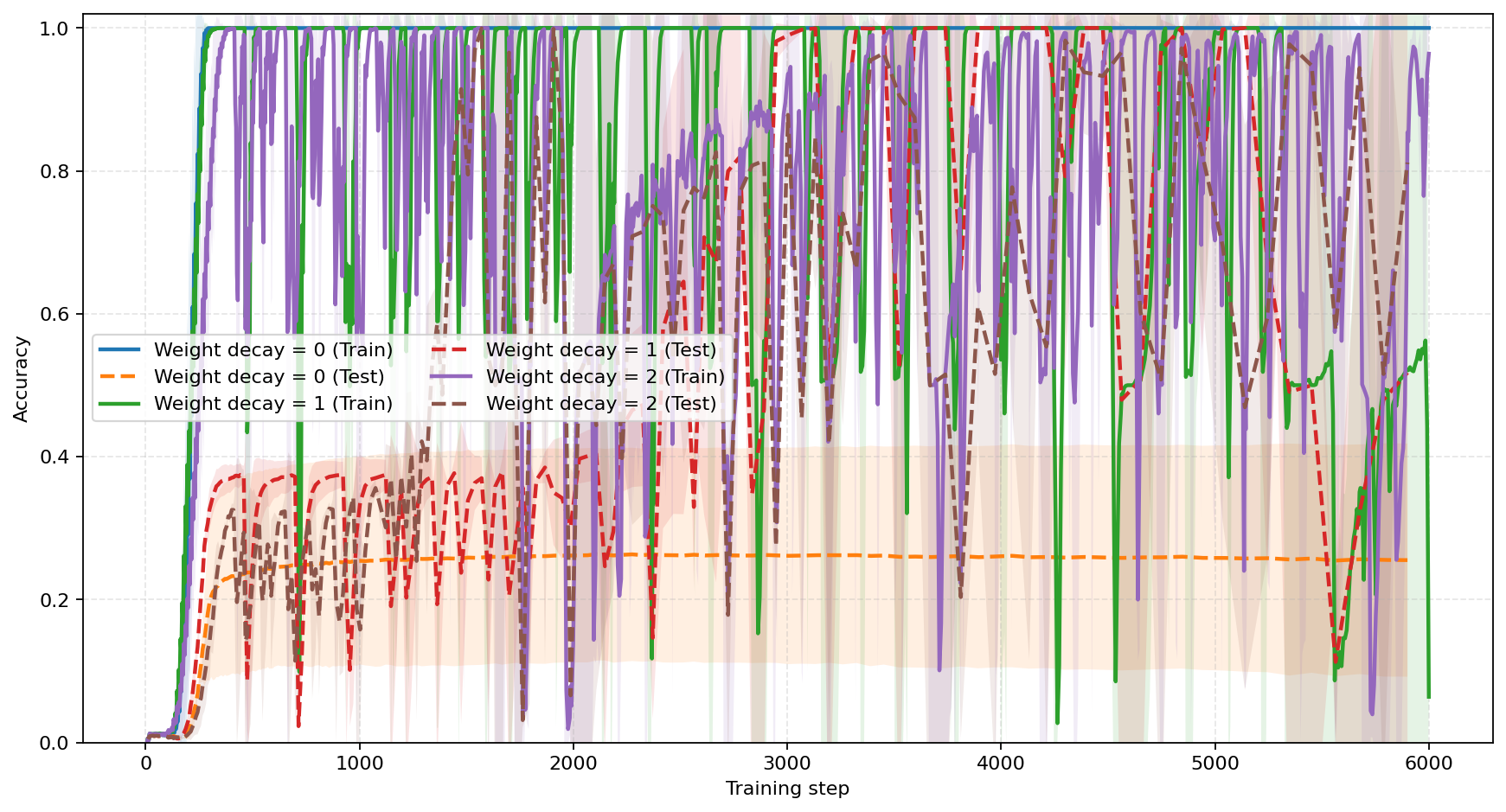}%
  }\hfill
  \subfigure[]{%
    \includegraphics[width=0.48\linewidth]{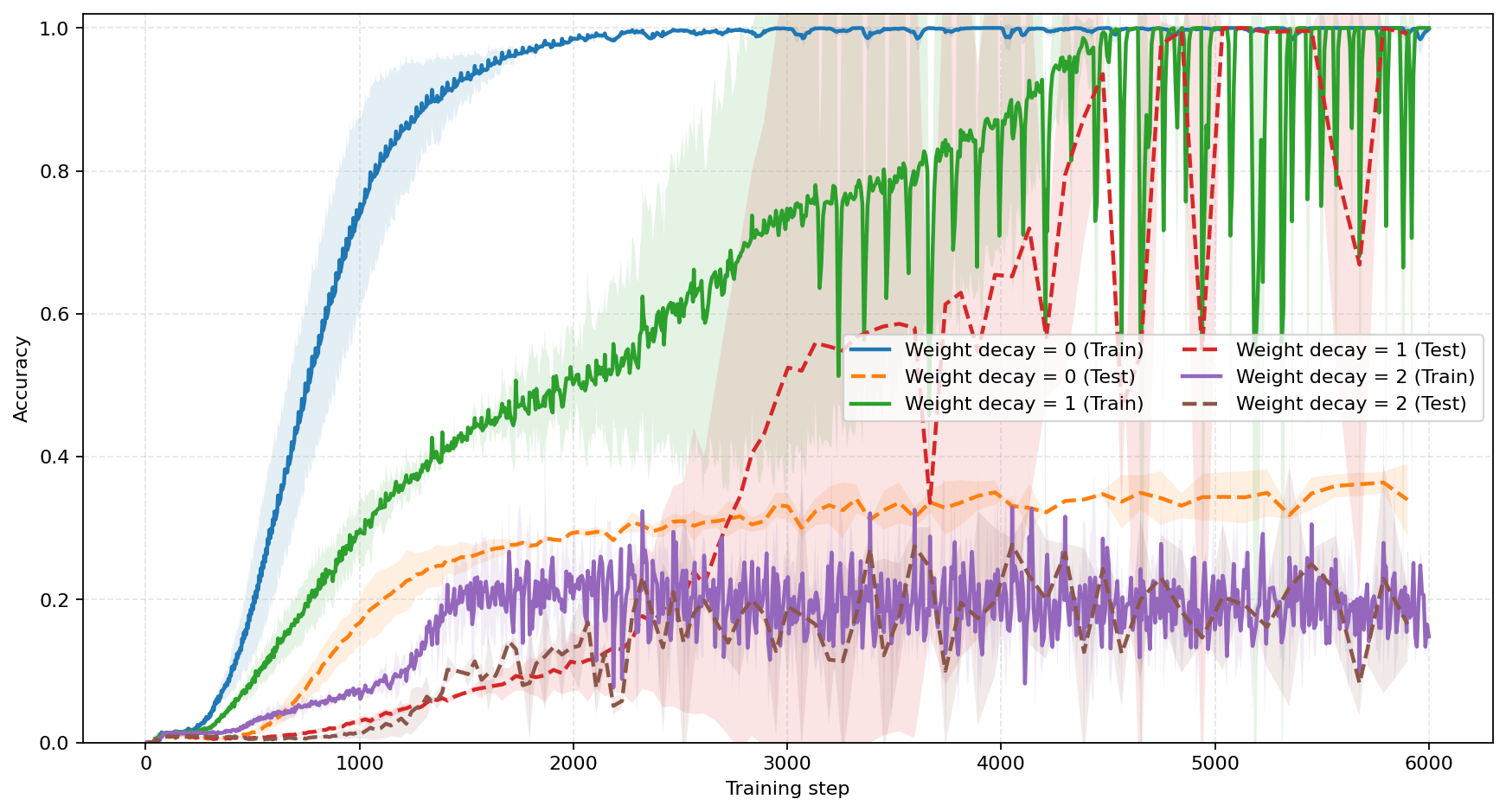}%
  }\\[2mm]
  \subfigure[]{%
    \includegraphics[width=0.48\linewidth]{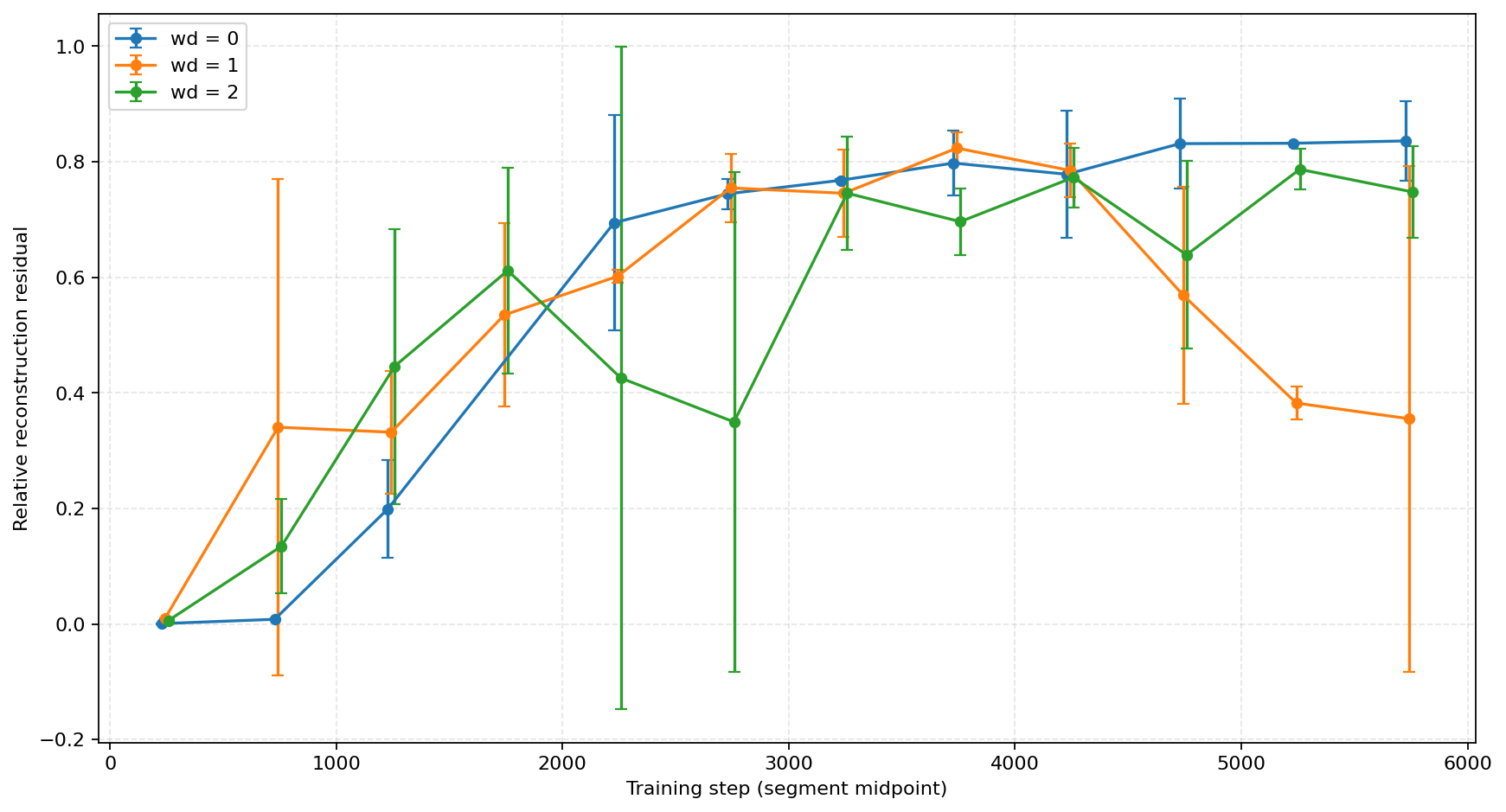}%
  }\hfill
  \subfigure[]{%
    \includegraphics[width=0.48\linewidth]{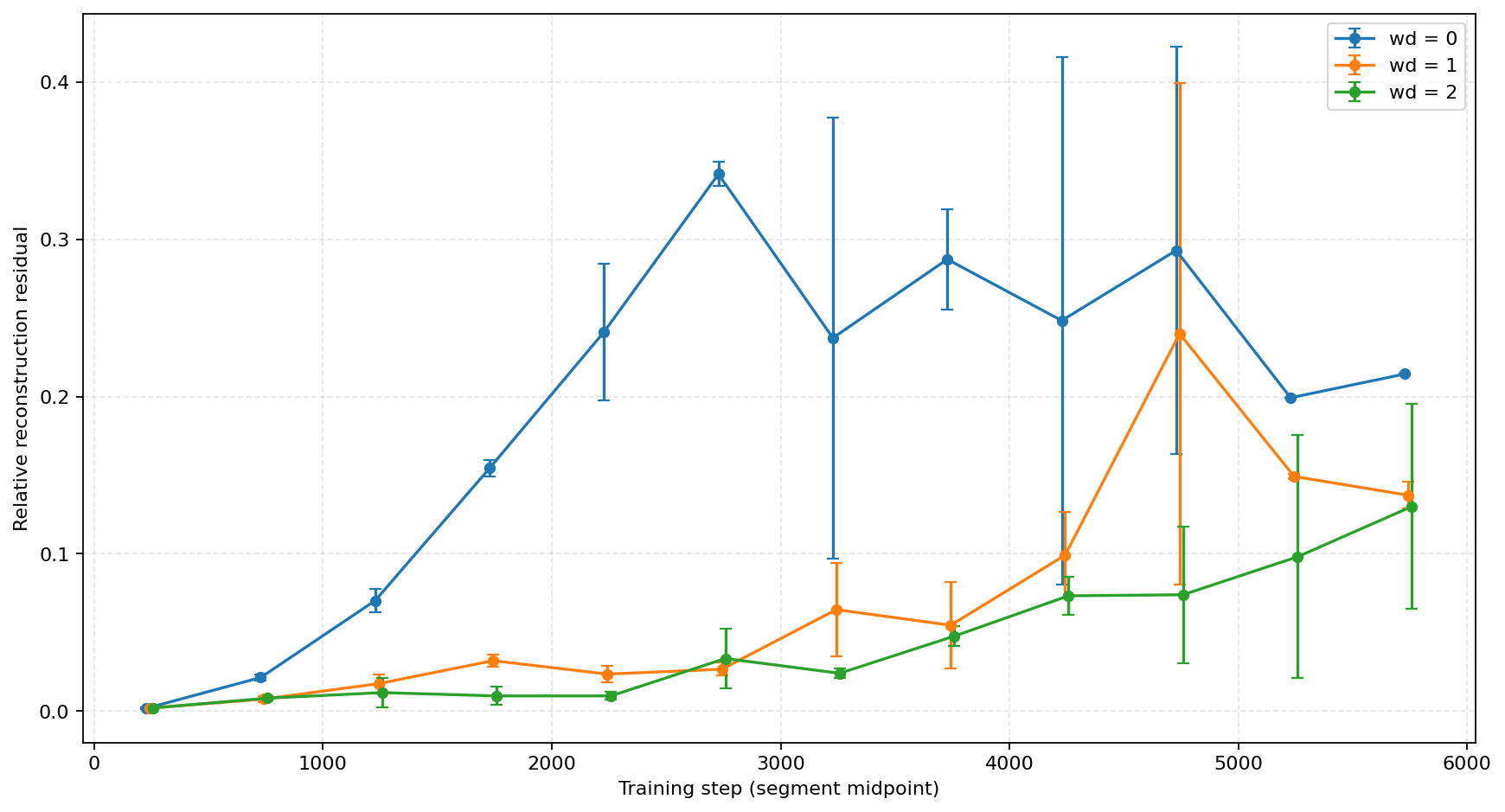}%
  }
  \caption{Test accuracy and reconstruction residual for the two architecture variants. Top left/right: accuracy bands for the large/deep and small/shallow models. Bottom left/right: the corresponding residual curves, showing that the residual--onset alignment is only partially preserved under architecture changes.}
  \label{fig:arch-ablation-plots}
\end{figure}

Figure~\ref{fig:arch-ablation-plots} compares the residual alignment and accuracy bands for both variants.

We emphasize that this ablation \emph{partially extends} the main findings to different structural hyperparameters, but it does not establish full robustness to arbitrary architecture choices.

\section{Stage Partition Sensitivity}\label{app:partition}

To examine whether the conclusions depend on the specific choice of segment size in the windowed DMD, we re-analyze a representative subset of expanded Transformer runs (2 seeds, $\mathrm{wd}\in\{0,1,2\}$) with three segment sizes: 250, 500, and 1000 steps. All other DMD hyperparameters ($n_{\mathrm{delays}}=32$, $k_{\mathrm{mode}}=10$) are held fixed.

Table~\ref{tab:partition} reports the key metrics. Across the three tested segment sizes $\{250,500,1000\}$, the qualitative observations that remain visible are:
\begin{enumerate}
  \item $\mathrm{wd}=0$ has the highest peak residual at every tested segment size;
  \item the peak-residual ordering $\mathrm{wd}=0 > \mathrm{wd}=1 > \mathrm{wd}=2$ holds at every tested segment size;
  \item generalization onset under $\mathrm{wd}=1$ is later than under $\mathrm{wd}=2$ at every tested segment size;
  \item lead-lag Spearman $\rho$ between RR and subsequent accuracy change stays above $0.76$ for the grokking regimes ($\mathrm{wd}=1,2$) at every tested segment size; for $\mathrm{wd}=0$ (no grokking) $\rho$ is lower ($\approx 0.54$--$0.60$), consistent with the weaker temporal coupling there.
\end{enumerate}
Quantitative values shift with segment size, but the orderings (Table~\ref{tab:partition-ordering}) are stable across the three sizes we tested, and the $\rho$ comparison between grokking and non-grokking regimes is preserved.

\begin{table}[h]
\centering
\caption{Partition sensitivity: peak residual, peak step, mean grokking onset, and lead-lag Spearman correlation between $\mathrm{Res}^{(r)}$ and subsequent accuracy change, under three segment sizes. Cells aggregate across $4$--$5$ seeds. ``--'' means the cell does not reach $99\%$ test accuracy.}
\label{tab:partition}
\small
\begin{tabular}{cccccc}
\toprule
Seg.\ size & wd & Peak RR & Peak step & Onset & Lead-lag $\rho$ \\
\midrule
250 & 0 & 0.760 & 4824 & -- & 0.557 \\
250 & 1 & 0.664 & 4062 & 3387 & 0.804 \\
250 & 2 & 0.627 & 5312 & 1539 & 0.767 \\
\midrule
500 & 0 & 0.780 & 4749 & -- & 0.543 \\
500 & 1 & 0.676 & 4374 & 3387 & 0.842 \\
500 & 2 & 0.611 & 5124 & 1539 & 0.762 \\
\midrule
1000 & 0 & 0.784 & 5099 & -- & 0.602 \\
1000 & 1 & 0.672 & 4249 & 3387 & 0.858 \\
1000 & 2 & 0.595 & 5249 & 1539 & 0.853 \\
\bottomrule
\end{tabular}
\end{table}

\begin{table}[h]
\centering
\caption{Ordering stability across segment sizes. ``$>$'' denotes the weight-decay value with the higher metric.}
\label{tab:partition-ordering}
\small
\begin{tabular}{lll}
\toprule
Seg.\ size & Peak RR order & Onset order \\
\midrule
250 & $0 > 1 > 2$ & $1 > 2$ \\
500 & $0 > 1 > 2$ & $1 > 2$ \\
1000 & $0 > 1 > 2$ & $1 > 2$ \\
\bottomrule
\end{tabular}
\end{table}

\section{AGOP Comparison and Joint Evidence}\label{app:agop}

The Average Gradient Outer Product (AGOP) \citep{radhakrishnan2024agop} provides a parallel route to identifying transition windows from gradient outer products. To relate it to our distributional diagnostics, we compute AGOP trace and top eigenvalue trajectories for the expanded Transformer runs and examine their joint behavior with the reconstruction residual.

\paragraph{Coverage limitation (N1).}
On the N1 baseline-battle pool, AGOP can only be evaluated at saved checkpoints; the current exponential checkpoint schedule leaves only $1$ non-grok run with full coverage, which is insufficient for a fair head-to-head AUROC comparison. We therefore do not interpret AGOP's resulting AUROC on this pool as evidence against the method, and treat AGOP as corroborative under sufficient checkpoint coverage rather than competitive evidence.

\begin{figure}[h]
  \centering
  \subfigure[]{%
    \includegraphics[width=0.32\linewidth]{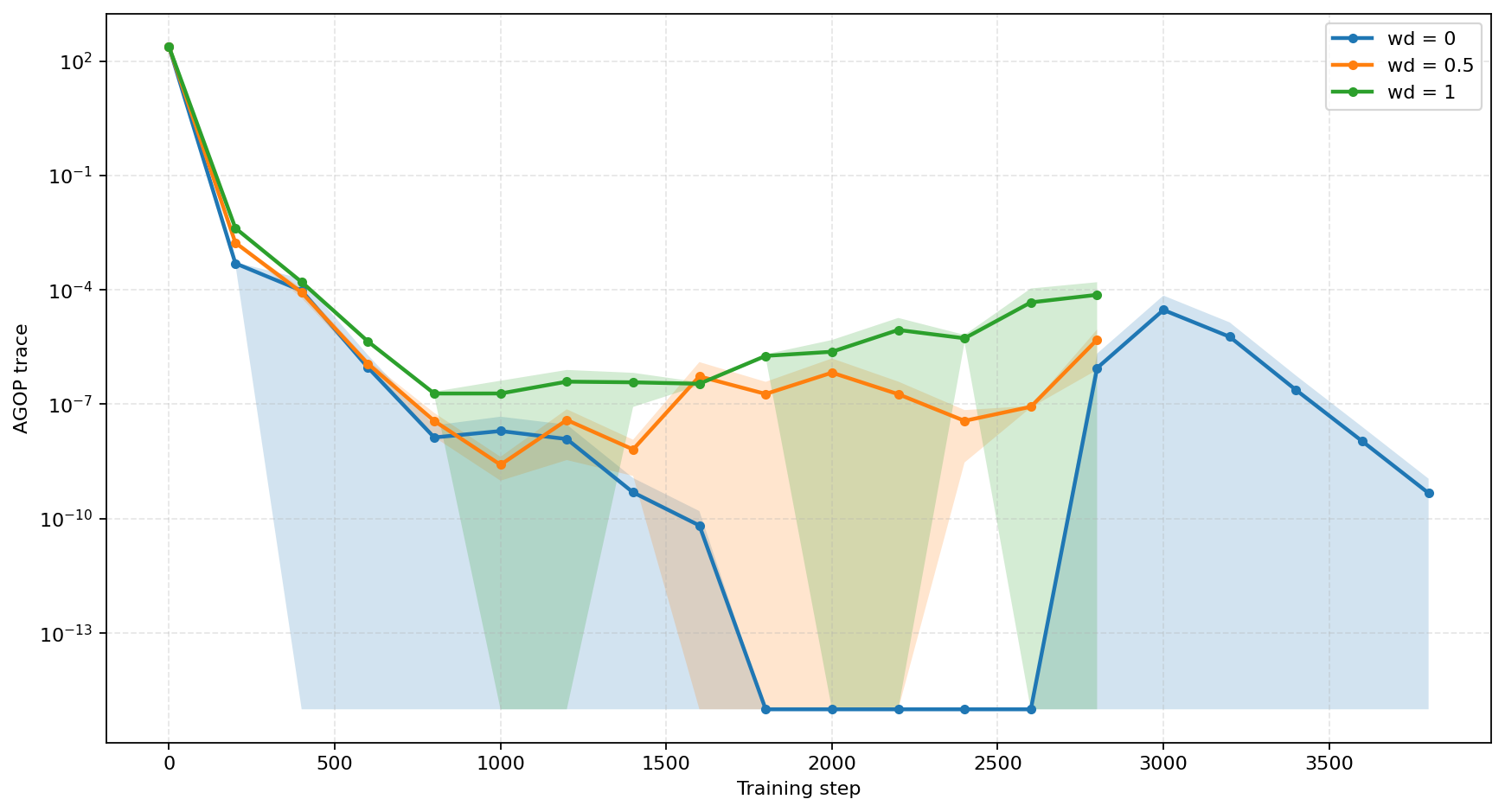}%
  }\hfill
  \subfigure[]{%
    \includegraphics[width=0.32\linewidth]{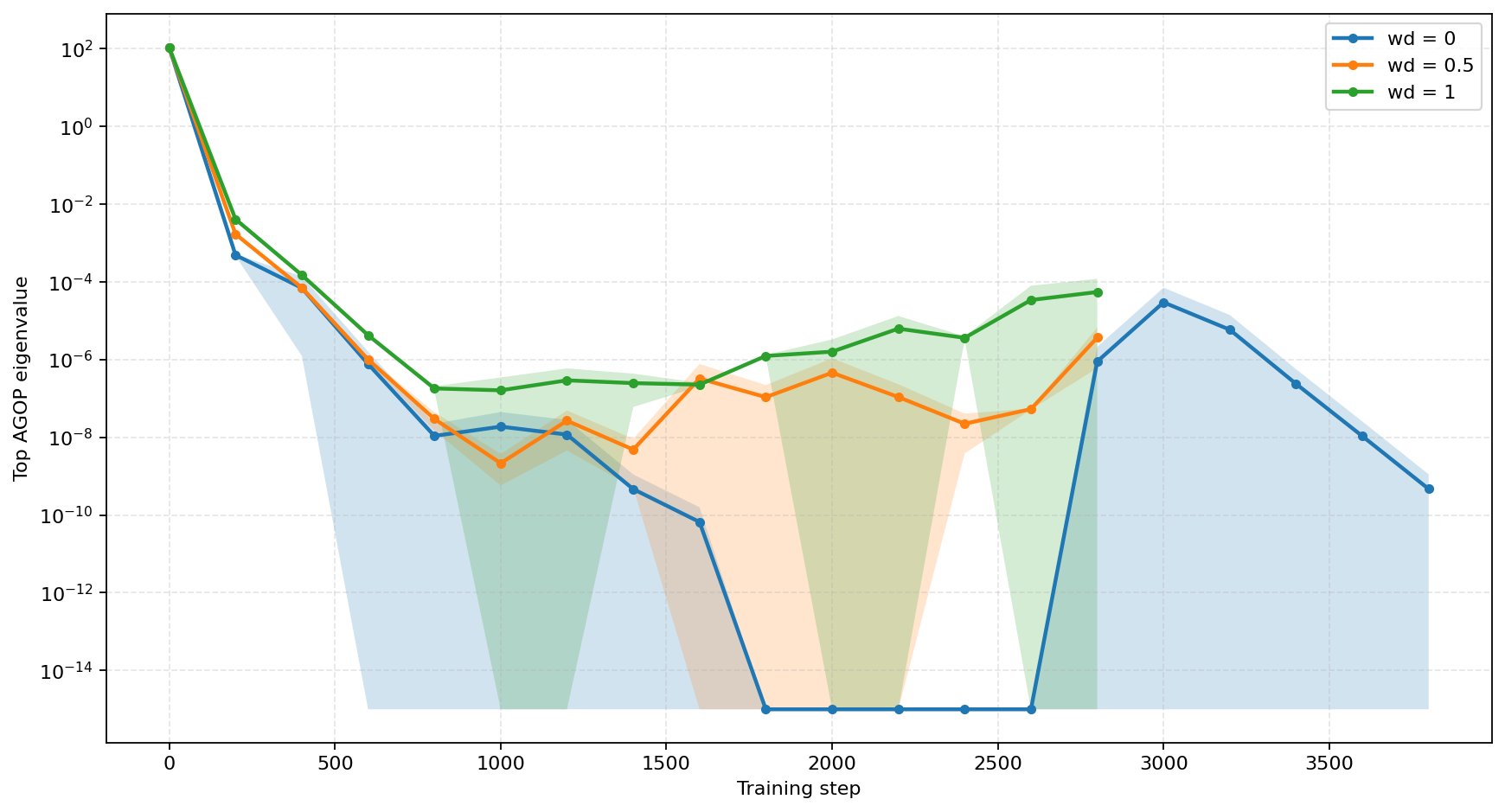}%
  }\hfill
  \subfigure[]{%
    \includegraphics[width=0.32\linewidth]{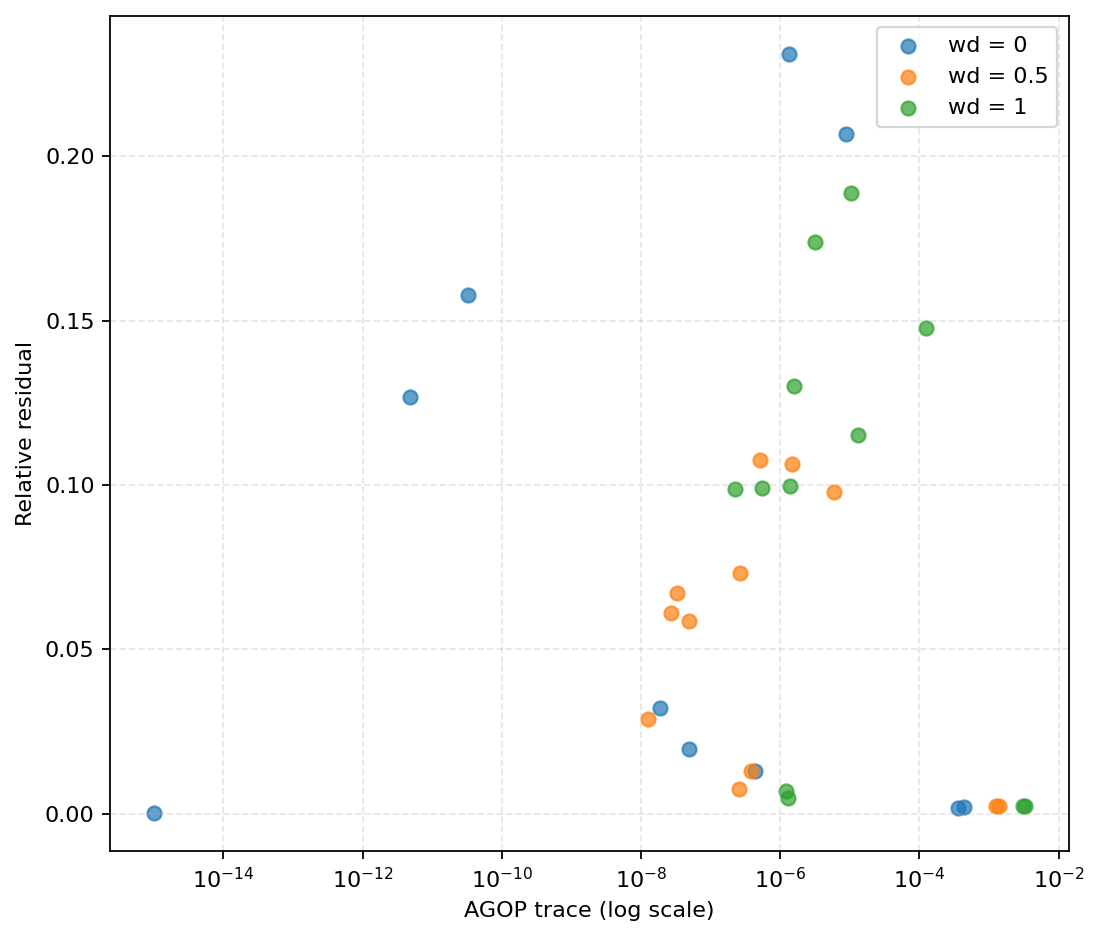}%
  }
  \caption{AGOP diagnostics and their relationship with reconstruction residual. Left: AGOP trace over training on a log scale. Middle: the top AGOP eigenvalue over training on a log scale. Right: scatter of relative residual against AGOP trace (log scale) per checkpoint. The relationship is non-monotonic on this pool: residual is elevated in the mid-AGOP regime that contains transition windows for grokking runs ($\mathrm{wd}\ge 1$), while saturated post-grokking checkpoints sit at large AGOP yet small residual. We therefore read panel (c) as qualitative co-elevation of AGOP and residual within the transition windows of the runs we could evaluate, not as a global monotone correlation between AGOP magnitude and residual.}
  \label{fig:agop-plots}
\end{figure}

Figure~\ref{fig:agop-plots} shows three complementary views: (a)~AGOP trace consistency across weight decay settings, (b)~AGOP top eigenvalue trajectories, and (c)~the joint evidence relating AGOP and reconstruction residual.

Subject to the coverage limitation noted above, the AGOP dynamics are qualitatively consistent with the distributional diagnostics in the runs we could evaluate: in those runs, grokking-regime trajectories ($\mathrm{wd}\ge 1$) show concurrent elevation in AGOP trace and reconstruction residual during the generalization transition window. We treat this as qualitative corroboration rather than independent confirmation; a fair quantitative AGOP--RR comparison would require a denser checkpoint schedule and is left for future work.

\begin{figure}[h]
  \centering
  \includegraphics[width=0.55\linewidth]{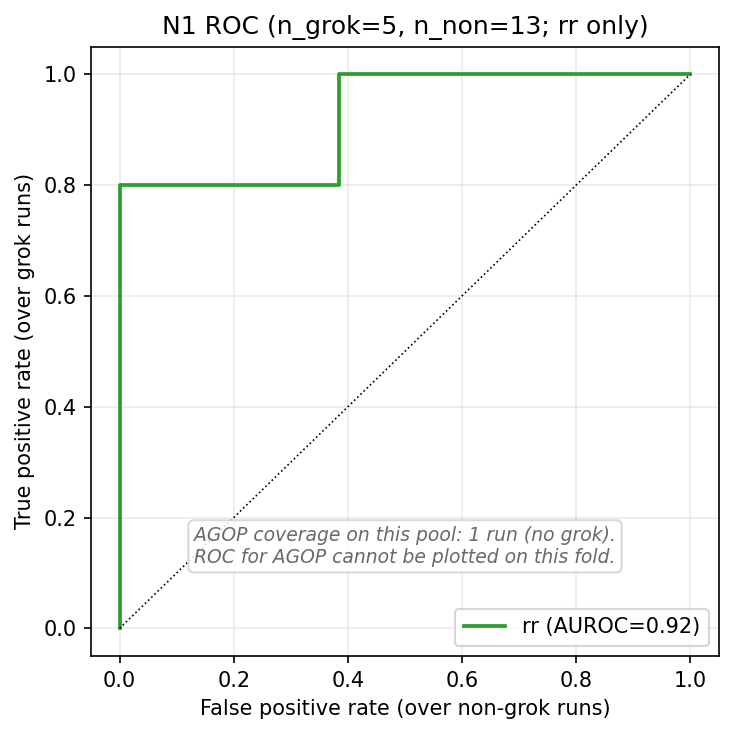}
  \caption{N1 baseline-battle pool ($18$ runs; $5$ grok / $13$ non-grok). Run-level ROC for the residual detector (AUROC $=0.9231$). AGOP cannot be plotted on this pool because its coverage is one run with no grokking, so TPR is undefined; this is annotated in-figure rather than rendered as an empty curve. We therefore do not interpret AGOP's score on this pool as evidence against the method.}
  \label{fig:n1}
\end{figure}

\paragraph{N1 detector pool.}
On the broader N1 baseline-battle pool ($18$ runs; $5$ grok / $13$ non-grok, including smoke and early-generalization border cases), the residual scores AUROC $=0.9231$ and AUPRC $=0.8944$. Figure~\ref{fig:n1} shows the run-level ROC for RR; AGOP cannot be placed in the (FPR, TPR) plane on this pool because its coverage is one non-grok run, so its TPR is undefined. We therefore do not interpret AGOP's nominal score on this pool as evidence against AGOP, and treat it only as corroborative under sufficient checkpoint coverage.

\section{DMD reconstruction quality calibration}\label{app:dmd-quality}

A natural question is whether the DMD approximation captures meaningful dynamics or merely fits noise. We address this with a holdout-based quality evaluation: for each DMD segment, we fit on a holdout split of the time steps within the segment and evaluate one-step prediction error on the held-out portion. We compare against a persistence baseline (identity transition) that predicts each next state as equal to the current state. ``Gain'' is the persistence-minus-holdout reduction in relative residual; positive values mean DMD beats persistence.

Table~\ref{tab:dmd-quality} reports the results, aggregated across seeds and segment sizes per weight-decay setting. The DMD holdout error is lower than the persistence baseline in all three regimes (positive gain), indicating that the windowed DMD fit captures nontrivial temporal structure beyond persistence. The gap is largest for $\mathrm{wd}=0$ on this pool: the wd=0 distributional trajectory is far from stationary and persistence is a particularly weak baseline there. The gap remains positive but smaller for $\mathrm{wd}=1,2$, where the trajectory is closer to a slow drift and persistence is already a strong baseline.

\begin{table}[h]
\centering
\caption{DMD quality: holdout vs.\ persistence baseline (relative residual). ``Gain'' is persistence rr minus holdout rr; positive means DMD beats persistence.}
\label{tab:dmd-quality}
\small
\begin{tabular}{ccccc}
\toprule
wd & (seed $\times$ seg-size) cells & Holdout rr & Persistence rr & Gain \\
\midrule
0 & 10 & 0.308 & 0.654 & $+0.346$ \\
1 & 8  & 0.276 & 0.378 & $+0.102$ \\
2 & 8  & 0.252 & 0.457 & $+0.205$ \\
\bottomrule
\end{tabular}
\end{table}

This calibration supports a conservative interpretation: across the regimes we tested, the DMD approximation yields a windowed prediction signal above the persistence baseline. The quality should be read alongside the regime context (the wd=0 trajectory has different structure than wd$=1,2$), and the windowed DMD fit should not be uniformly treated as a reliable predictor across all training settings.

\section{Portability to CIFAR-10 with a Tiny CNN}\label{app:cifar}

To address the concern that results are limited to MNIST MLPs and modular-addition Transformers, we apply the analysis pipeline to a Tiny CNN (two convolutional layers + global average pooling + linear head) trained on CIFAR-10. We use logits as the observable and sweep over 2 seeds $\times$ 2 channel widths $\{32,64\}$ $\times$ 2 optimizers $\{\text{AdamW},\text{SGD}\}$ for a total of 8 runs, each trained for 5 epochs.

Table~\ref{tab:cifar} reports the DMD summary. The method produces readable reconstruction residual and effective rank curves in this new setting. Wider networks ($\text{channels}=64$) exhibit lower peak residual, paralleling the width dependence observed for MNIST MLPs.

\begin{table}[h]
\centering
\caption{CIFAR-10 Tiny CNN: DMD diagnostic summary. Each cell averages over $2$ seeds.}
\label{tab:cifar}
\small
\begin{tabular}{llccc}
\toprule
Optimizer & Channels & Peak RR & Peak step & Peak rank \\
\midrule
AdamW & 32 & 0.619 & 1250 & 2.5 \\
SGD   & 32 & 0.595 & 1500 & 2.5 \\
AdamW & 64 & 0.565 & 875.5 & 3 \\
SGD   & 64 & 0.528 & 750.5 & 3 \\
\bottomrule
\end{tabular}
\end{table}

\begin{figure}[h]
  \centering
  \subfigure[]{%
    \includegraphics[width=0.48\linewidth]{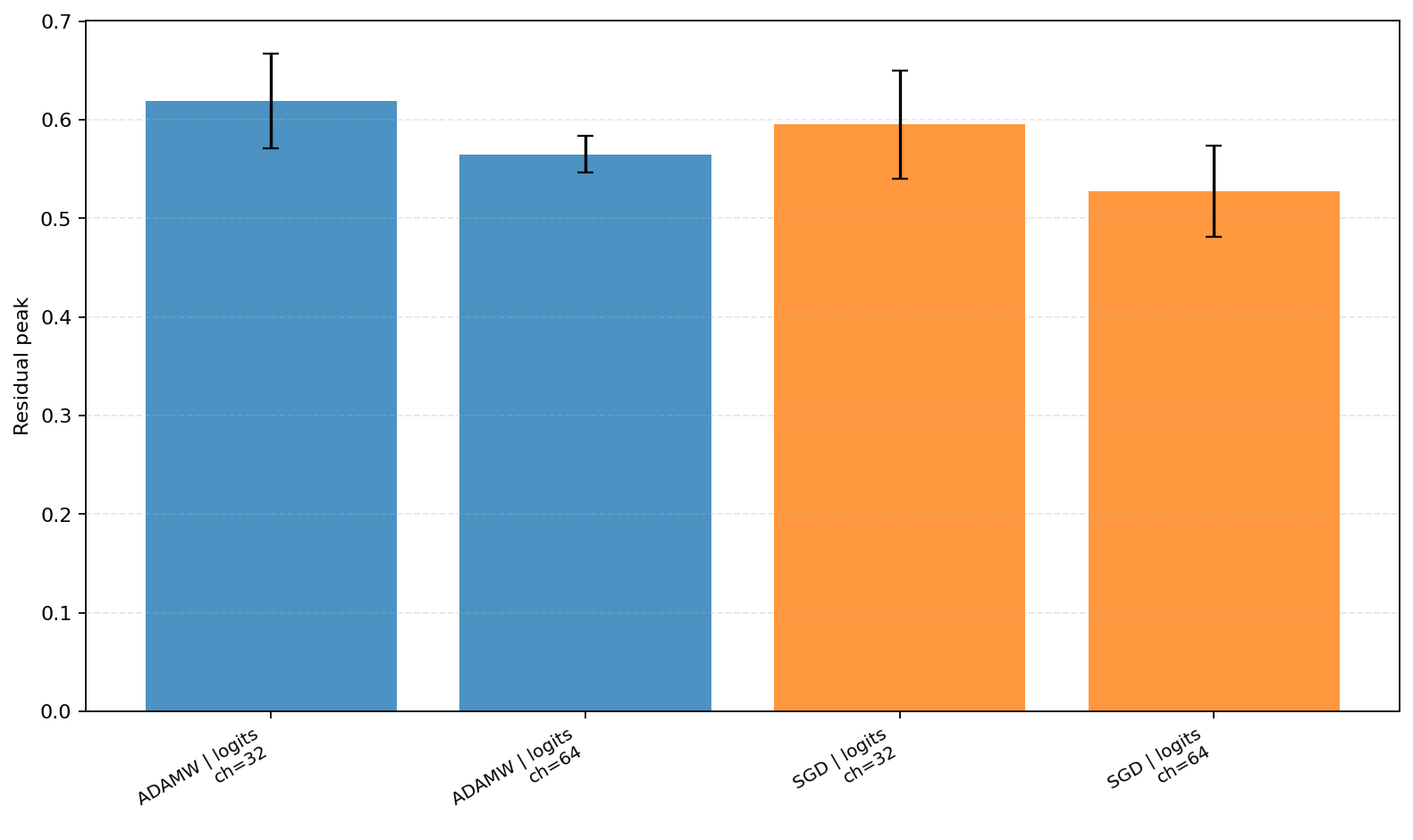}%
  }\hfill
  \subfigure[]{%
    \includegraphics[width=0.48\linewidth]{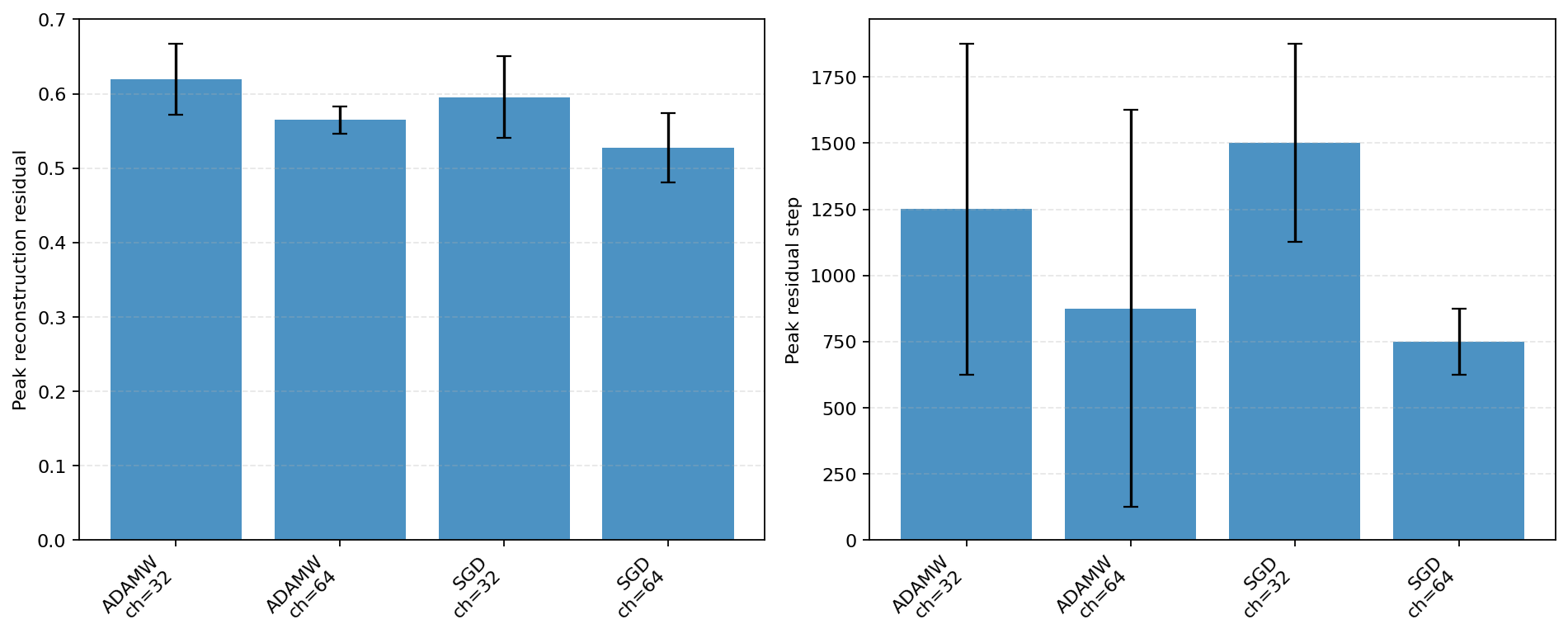}%
  }
  \caption{CIFAR-10 Tiny CNN: DMD diagnostics across channel widths and optimizers. Left: peak residual by optimizer/channel configuration. Right: peak residual and peak residual step by optimizer/channel, showing that wider channels yield lower residuals and earlier peak timing.}
  \label{fig:cifar-plots}
\end{figure}

CIFAR-10 is included as a portability check only, not as a grokking diagnostic benchmark: the run is small ($8$ runs, $5$ epochs), there is no grokking transition to localize in this setup, and no detection metric (TPR/FPR/AUROC) is reported. The pipeline runs end-to-end and yields a readable RR/effective-rank curve; we draw no detection or fragility conclusion from this section.

\section{Perturbation fragility: full pool}\label{app:sensitivity-window}

This appendix details the perturbation-fragility experiment summarized in \S\ref{subsec:fragility}.

\paragraph{Pool and protocol.}
On wd$=1$ grokking baselines, identical multiplicative perturbations of magnitude $\{0.005, 0.01\}$ are applied to model parameters at high-RR vs.\ low-RR windows ($4$ high-RR runs and $5$ low-RR runs at each scale, $21$ runs in total counting the additional low-norm seeds). For each perturbation we record short-horizon accuracy deviation (mean accuracy gap between perturbed and unperturbed trajectories over the first $1000$ steps after perturbation) and final accuracy.

\paragraph{Results.}
Mean short-horizon deviation: at scale $0.01$, $0.090$ (high-RR) vs.\ $0.029$ (low-RR), giving a high/low ratio $\approx 3.1\times$; at scale $0.005$, $0.107$ vs.\ $0.034$, ratio $\approx 3.2\times$.

\paragraph{Boundary cases.}
One unrecoverable failure occurs in a high-RR window near the transition (consistent with the high-RR-as-fragile reading). A second unrecoverable failure occurs in a low-RR window very early in training; we interpret this as a separate early-training instability unrelated to the transition window and treat it as a boundary case rather than a low-RR failure of the fragility hypothesis.

The result quantifies sensitivity under matched noise; it does not establish that the residual is a causal driver of the transition.

\section{Discussion: method positioning}\label{app:positioning}

The supplementary experiments collectively support a narrow positioning of the distributional spectral diagnostics in the studied modular-arithmetic Transformer settings:

\begin{enumerate}
  \item \textbf{Window-level transition localization.} On the held-out test fold (Appendix~\ref{app:threshold}), the \texttt{sustained\_K2\_tau10} rule attains AUROC $\approx 0.93$, TPR $=0.80$ at FPR $=0.50$, and median lead $1068$ steps on true-positive alarms ($95\%$ bootstrap CI $[142,\,2426]$; Appendix~\ref{app:uncertainty}). Lead time is threshold-dependent and is always reported jointly with FPR.

  \item \textbf{Window-level fragility under matched perturbations.} The sensitivity-window experiment (Appendix~\ref{app:sensitivity-window}) reports that high-RR windows show elevated short-horizon perturbation sensitivity relative to low-RR windows in wd$=1$ baselines under matched noise. The result quantifies sensitivity, not causal mechanism; one boundary-case low-RR early-training failure is reported alongside the main pool.

  \item \textbf{Not a total-norm proxy at the window level.} The norm-window control (Appendix~\ref{app:norm-baseline}) re-labels the same perturbation runs by total-parameter-norm percentile and reverses the fragility ordering, indicating that the residual carries window-level information not captured by the total-norm signal. Norm-derived signals nevertheless remain strong run-level regime indicators on the same pool, and we do not claim RR universally outperforms norm baselines.

  \item \textbf{AGOP as a parallel route under coverage constraints.} The AGOP comparison (Appendix~\ref{app:agop}) shows qualitative co-occurrence between AGOP elevation and RR elevation in transition windows for the runs with sufficient checkpoint coverage. AGOP coverage on the N1 baseline-battle pool is one run, so we treat AGOP as corroborative under sufficient coverage rather than a competitor; a fair head-to-head comparison would require a denser checkpoint schedule.

  \item \textbf{Scope checks: weight decay, segment size, architecture.} The weight-decay sweep (Appendix~\ref{app:wd-robust}) shows that the qualitative ordering between regularization strength and onset timing remains visible across five settings; segment-size sensitivity (Appendix~\ref{app:partition}) shows coarse conclusions stable across $\{250,500,1000\}$ steps with fine ordering varying; the architecture ablation (Appendix~\ref{app:arch-ablation}) shows the signal is observed across the three tested Transformer scales but with shrinking residual amplitude at small scale, indicating model-scale sensitivity rather than parameterization-class behavior.

  \item \textbf{Scope checks: portability and FCN.} CIFAR-10 (Appendix~\ref{app:cifar}) is a portability check that the pipeline runs end-to-end on a different task/architecture; it is not a grokking diagnostic benchmark, and no detection metric is reported there. FCN results (Appendix~\ref{app:fcn-secondary}) are a secondary low-residual regime descriptor use, not a grokking diagnostic claim.
\end{enumerate}

These results position the method as a window-level monitoring and localization signal in the studied modular-arithmetic Transformer settings, not as a universal early-warning predictor, an architecture-independent diagnostic, an automatic intervention rule, or a replacement for norm-based regime classifiers.

\section{Onset criterion and labeling protocol}\label{app:onset}

The grokking-vs-non-grokking label used by the §\ref{subsec:detection} detector requires (i) an onset definition and (ii) a step threshold separating grokking from early generalization.

\paragraph{Onset definition.}
For each run, grokking onset is the first step at which test accuracy crosses $99\%$. In the base pool ($41$ unique runs: $8$ seeds $\times$ $5$ weight-decay settings $+\,1$ smoke run), this gives a mean wd$=1$ onset of $3398$ steps with std $361$ over $9$ wd$=1$ runs; wd$=2$ runs reach the same threshold by step $1528$ on average. The released code contains scripts to reproduce these aggregates.

\paragraph{Bimodal gap.}
The empirical onset distribution is clearly bimodal between wd$=2$ early-generalization runs and wd$=1$ grokking runs: the wd$=2$ pool's maximum onset is well below the wd$=1$ pool's minimum onset, and we choose the step threshold to fall in this gap. The threshold is data-driven (post-hoc) but supported by the visible bimodal structure.

\begin{figure}[h]
  \centering
  \includegraphics[width=0.85\linewidth]{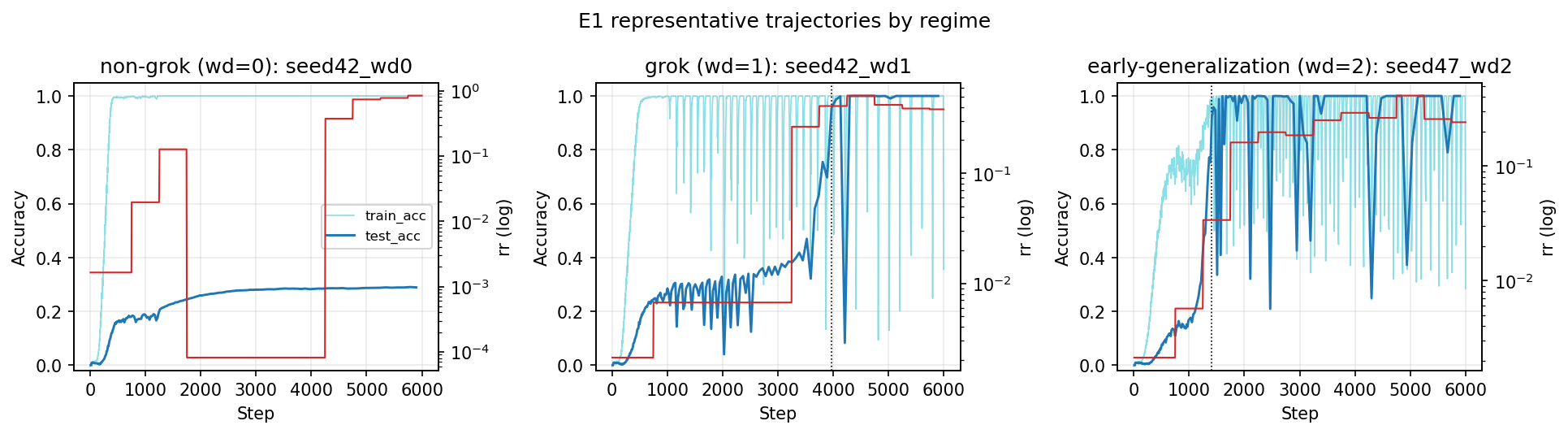}
  \caption{Representative test-accuracy and RR trajectories for non-grok, grok, and early-generalization regimes in the base pool ($41$ unique runs).}
  \label{fig:pool-trajectories}
\end{figure}

\paragraph{Pool overview.}
Figure~\ref{fig:pool-trajectories} shows representative test-accuracy and RR trajectories for non-grok, grok, and early-generalization runs in the base pool.

\section{Threshold sweep and reused-seed split behavior}\label{app:threshold}

This section reports the full threshold sweep on the held-out test fold and documents the asymmetry between the held-out test split and a reused-seed split. The reused-seed split is reported as a sensitivity check, not as a successful calibration fold; the operating rule of \S\ref{subsec:protocol} is fixed by heuristic and sustainment, not by tuning on either split.

\paragraph{Test-fold sweep.}
The test fold is fresh seeds $46$--$49$ ($n_{\mathrm{grok}}{=}5$, $n_{\mathrm{non}}{=}12$, base rate $\approx 0.29$; wd$=2$ early-generalization runs excluded). Lead and summaries are computed only over true-positive alarms.

\begin{table}[h]
\centering
\caption{Detection threshold trade-off on the held-out test fold ($n_{\mathrm{grok}}{=}5$, $n_{\mathrm{non}}{=}12$). Median lead is computed only over true-positive alarms; the bracketed range for the selected sustained rule is the $95\%$ bootstrap CI of the median (Appendix~\ref{app:uncertainty}). AUROC $\approx 0.93$ (AUPRC $\approx 0.91$) is computed on this same fold and applies to all rows (single detector at different cuts).}
\label{tab:detection-app}
\small
\begin{tabular}{lccccl}
\toprule
Rule & TPR & FPR & Median lead & $95\%$ CI & Comment \\
\midrule
\multicolumn{6}{l}{\emph{Instantaneous threshold rules}} \\
$\tau{=}5\times$              & $1.00$ & $0.917$ & $1774$  & ---            & high recall \\
$\tau{=}20\times$             & $0.40$ & $0.250$ & ---     & ---            & high specificity \\
\midrule
\multicolumn{6}{l}{\emph{Sustained operating rule}} \\
\texttt{sustained\_K2\_tau10} (selected)  & $0.80$ & $0.500$ & $1068$ & $[142,\,2426]$ & moderate recall/FPR \\
\bottomrule
\end{tabular}
\end{table}

\begin{figure}[h]
  \centering
  \subfigure[ROC on test fold]{%
    \includegraphics[width=0.45\linewidth]{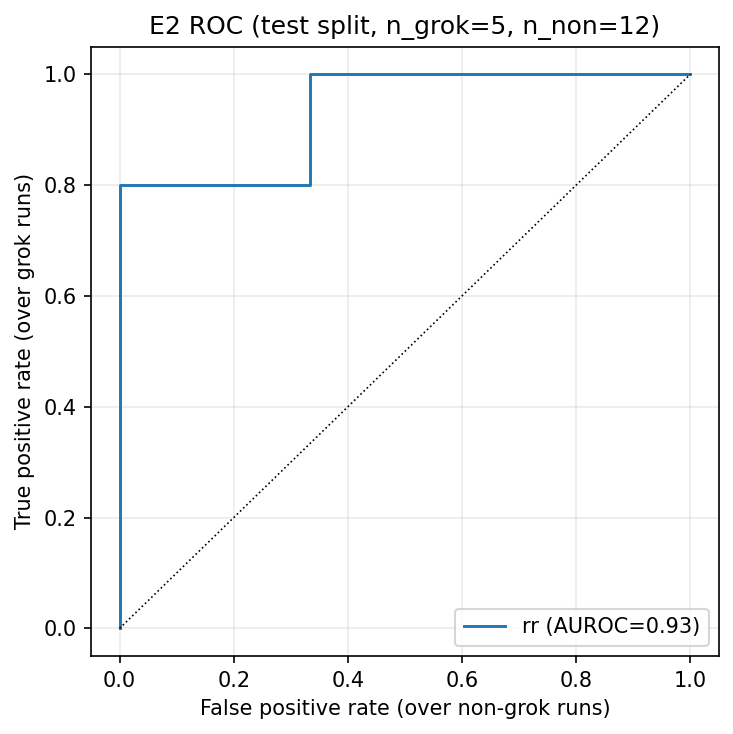}%
    \label{fig:roc}%
  }\hfill
  \subfigure[Lead-time distribution]{%
    \includegraphics[width=0.45\linewidth]{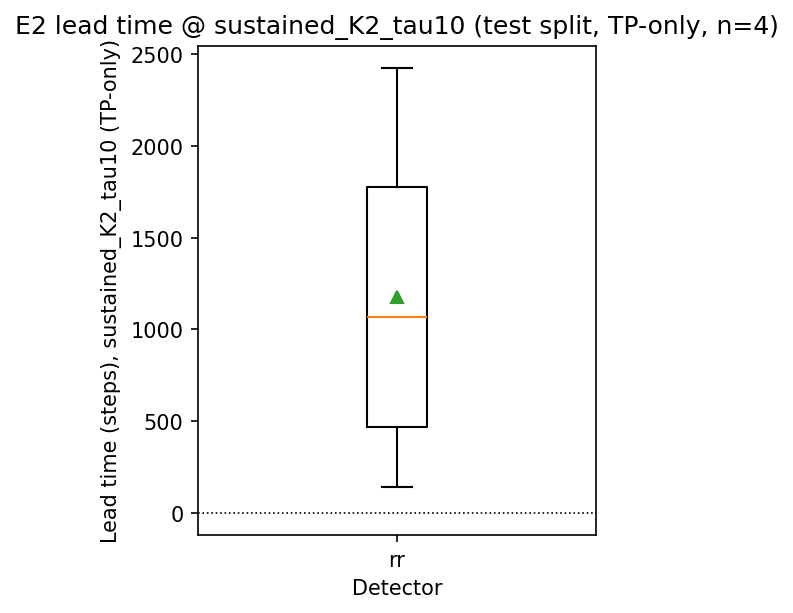}%
    \label{fig:leadbox}%
  }
  \caption{Detection on the held-out test fold ($5$ grok / $12$ non-grok). (a) ROC for the residual detector. (b) Per-threshold distribution of lead times across grok runs, computed only over true-positive alarms.}
  \label{fig:e2}
\end{figure}

\paragraph{Test-fold ROC and lead-time distribution.}
Figure~\ref{fig:e2} shows the ROC for the residual detector on the test fold and the per-threshold distribution of lead times across grok runs (computed only over true-positive alarms).

\paragraph{Reused-seed split behavior.}
On the reused-seed split (seeds $42$--$45$; $5$ grok / $11$ non-grok), the same fixed \texttt{sustained\_K2\_tau10} operating point fires no alarms, yielding TPR $=0$ and FPR $=0$. We report this as seed-split sensitivity rather than as evidence of calibrated threshold selection: the reused-seed split does not validate threshold calibration, and we therefore avoid using it to tune $K$ or $\tau$.

\begin{figure}[h]
  \centering
  \subfigure[Precision--recall on the test fold.]{%
    \includegraphics[width=0.45\linewidth]{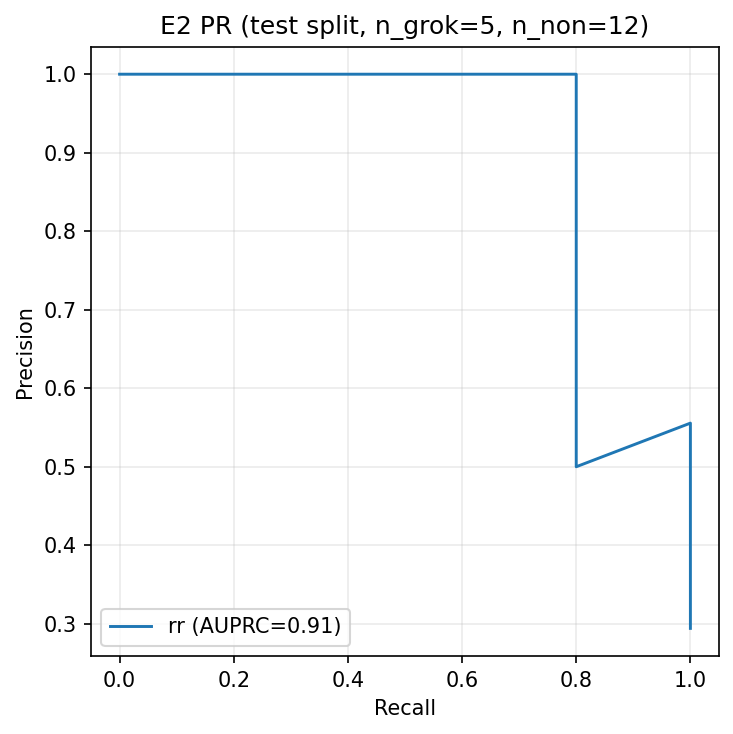}%
    \label{fig:e2-pr}%
  }\hfill
  \subfigure[Non-grok seed with rising RR (no generalization).]{%
    \includegraphics[width=0.45\linewidth]{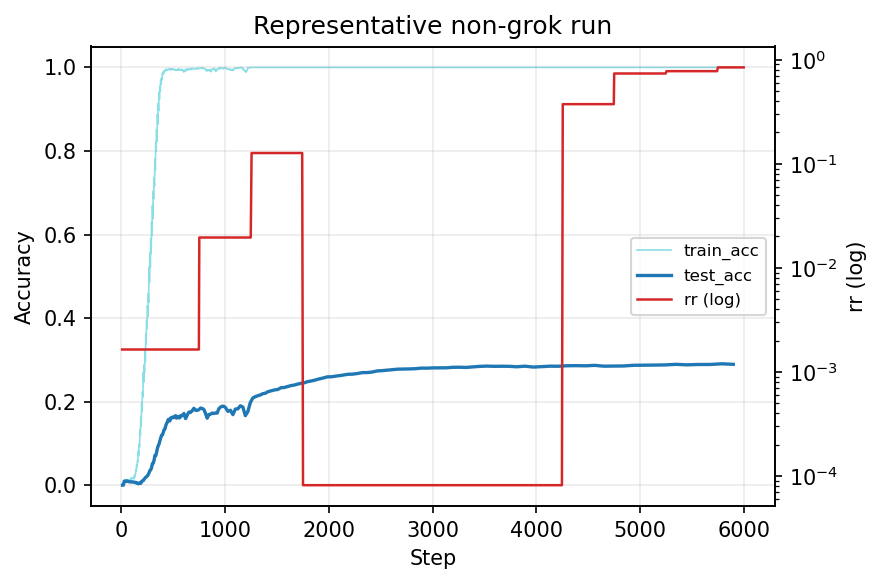}%
    \label{fig:case-failure}%
  }
  \caption{Test-fold detection diagnostics. (a) AUPRC $\approx 0.91$ on the test fold. (b) A non-grok run whose RR rises without producing generalization; sustained-threshold rules with relative baselines are intended to suppress such cases at the cost of some lead.}
  \label{fig:e2-aux}
\end{figure}

\paragraph{Precision--recall and a non-grok rising-residual case.}
Figure~\ref{fig:e2-pr} shows the precision--recall curve on the same test fold. Figure~\ref{fig:case-failure} shows a non-grok run whose RR also rises but no generalization follows; this motivates the relative sustained-threshold rule rather than an absolute residual cut.

\section{Uncertainty estimates for held-out detection}\label{app:uncertainty}

This appendix reports uncertainty estimates for the held-out detection results of \S\ref{subsec:detection}. AUROC and AUPRC intervals are computed by stratified bootstrap over runs, resampling grokking and non-grokking runs separately. TPR and FPR intervals use exact binomial intervals. Lead-time intervals are computed over true-positive alarms unless otherwise stated. The bootstrap unit is the run, not the window. These estimates are intended to quantify small-sample uncertainty, not to introduce additional claims or to reselect the operating point. The fixed operating point is not reselected by these uncertainty estimates; the intervals only quantify uncertainty for the held-out evaluation.

\begin{table}[t]
\centering
\small
\caption{Uncertainty estimates for the held-out detection results. AUROC and AUPRC intervals are computed by stratified bootstrap over runs. TPR and FPR intervals use exact binomial intervals. Lead-time intervals are computed over true-positive alarms unless otherwise stated.}
\label{tab:uncertainty_detection}
\begin{tabular}{lcc}
\toprule
Quantity & Estimate & $95\%$ interval \\
\midrule
AUROC & $0.933$ & $[0.750,\,1.000]$ \\
AUPRC & $0.906$ & $[0.642,\,0.933]$ \\
TPR   & $4/5 = 0.800$ & $[0.284,\,0.995]$ \\
FPR   & $6/12 = 0.500$ & $[0.211,\,0.789]$ \\
Lead, TP alarms only & $1068$ steps & $[142,\,2426]$ \\
Lead, all alarming grokking runs & $574$ steps & $[-226,\,2426]$ \\
\bottomrule
\end{tabular}
\end{table}

The wide intervals reflect the small held-out split. The AUROC/AUPRC intervals support the run-level ranking value of RR, while the TPR/FPR intervals show that the fixed operating point has substantial uncertainty. The TP-only lead summarizes alarms that occur before grokking onset; the all-alarming lead additionally includes grokking runs whose alarms occur after onset, which explains the negative lower endpoint. We report these intervals as small-sample uncertainty quantification and not as evidence that the fixed operating point is statistically calibrated or validated.

\section{Observable ablation}\label{app:n3-obs}

This appendix reports the per-observable detection results referenced in \S\ref{subsec:protocol}.

For the $4$ grok runs in the observable-ablation sub-pool, log-probability ($19$-dim) fires \texttt{sustained\_K2\_tau10} alarms on $4/4$ runs but only $2/4$ before onset (true positive); the median true-positive lead is $\approx 601$ steps. Logits ($19$-dim) fire alarms on $3/4$ runs but $0/3$ before onset; top-$k$ ($\sim 95$-dim) and hidden ($\sim 1900$-dim) fire $0/4$ alarms. The matched FPR for these lead numbers is the rule's test-fold value $0.50$ from \S\ref{subsec:detection}; per-observable FPR on this sub-pool is not separately reported. As in \S\ref{subsec:protocol} and the Discussion, the present implementation is best suited to scalar empirical observables represented by one-dimensional quantile coordinates, so the failure of top-$k$ and hidden-state variants under the same DMD configuration should be interpreted as a limitation of this scalar-distribution implementation and fixed DMD setup, not as evidence that those observables lack useful information.

\begin{figure}[h]
  \centering
  \includegraphics[width=0.95\linewidth]{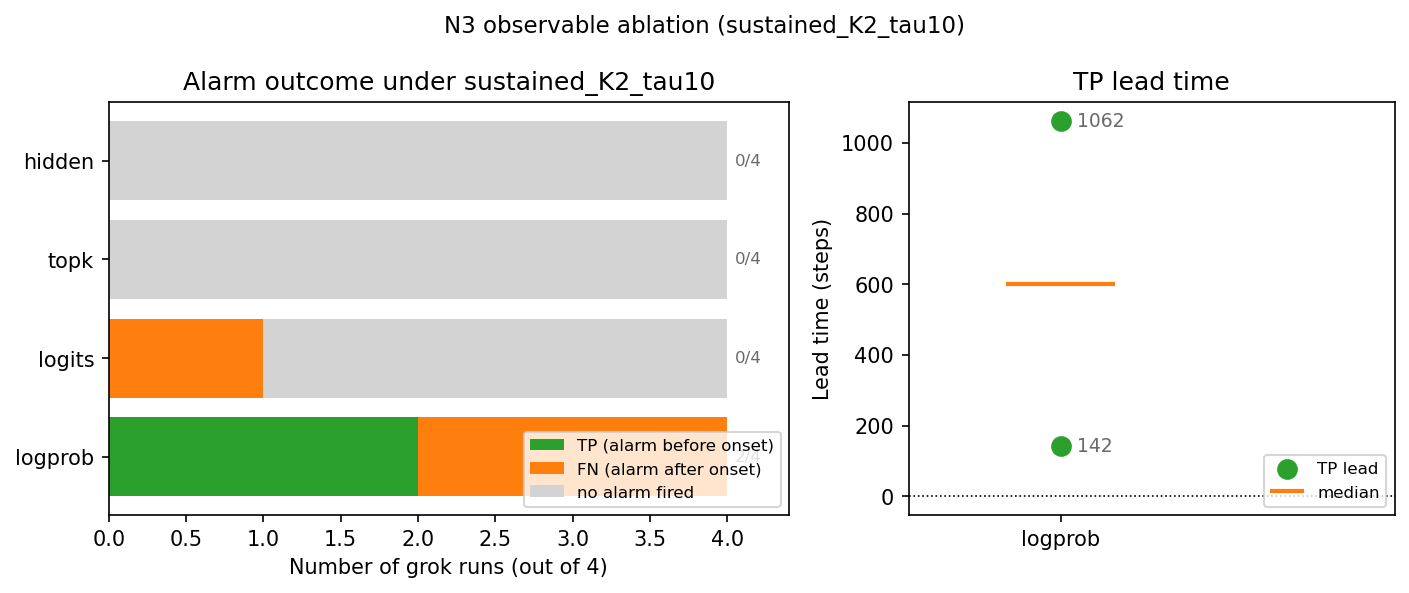}
  \caption{N3 observable ablation under the selected \texttt{sustained\_K2\_tau10} rule, on the four grok runs of the sub-pool. \textbf{Left:} alarm outcome per observable, stacked over the four runs: green = TP (alarm before onset), orange = FN (alarm after onset), gray = no alarm fired. Log-probability fires alarms on all four runs and lands two of them before onset (TP); logits fires once and lands after onset (FN); top-$k$ and hidden produce no alarms under the current DMD configuration. \textbf{Right:} lead time of the TP alarms — only log-probability has TPs, with leads of $142$ and $1062$ steps (median $\approx 601$). The figure shows that, under this rule and DMD setup, log-probability is the only scalar-distribution observable that yields useful pre-onset signal on this sub-pool; we read this as a limitation of the current scalar-quantile implementation, not as evidence that the other observables lack information.}
  \label{fig:n3-lead}
\end{figure}

\section{Norm and logit-scale baselines}\label{app:norm-baseline}

We compare the residual against norm-derived signals at two scales: run-level (max-of-trajectory) discrimination and window-level fragility under matched perturbations.

\paragraph{Pool.}
$23$ runs enter the pipeline; $22$ have finite ROC scores under all signals on the shared pool.

\paragraph{Run-level AUROC.}
Under run-level max-of-trajectory scoring on this pool, total parameter norm gives AUROC $=0.9829$, $\lvert\Delta\mathrm{norm}\rvert$ gives $0.9915$, and RR (run-level max) gives $0.4103$. Norm-derived signals are therefore strong run-level regime indicators on this pool, while RR's strength is at the window level (next paragraph and Table~\ref{tab:e5p-app}).

\paragraph{Fair-FPR temporal-alarm protocol.}
For the fair-FPR norm comparison, each signal's threshold is selected separately to match the target FPR on the reused-seed split; these thresholds are not the \texttt{sustained\_K2\_tau10} operating rule used in the main RR experiment of \S\ref{subsec:detection}. This is a baseline-comparison protocol, not the selected operating rule, and the reused-seed split is used here only to set comparison thresholds, not as a successful calibration fold. With FPR target $0.50$ on the reused-seed split and reported on the test fold ($n_{\mathrm{test\_grok}}{=}4$, $n_{\mathrm{test\_non\_grok}}{=}5$): under this fair-FPR protocol, RR triggers TPR $=0.75$ with median lead $273.5$ steps; \texttt{norm\_N\_total} triggers TPR $=0$ (no alarms fire) under the same fair-FPR protocol. A fully fair head-to-head detection comparison would require per-signal threshold recalibration tailored to each signal's dynamic range and is left for future work.

\paragraph{Window-control contrast.}
The same $21$ perturbation runs ($17$ reused from \S\ref{subsec:fragility} plus $4$ new low-norm $s{=}5250$ perturbations on seeds $46$/$49$) are re-labeled by total-parameter-norm percentile rather than RR percentile. The aggregator below reports short-horizon deviation in accuracy on a $0$--$100$ scale, so values are $100\times$ those in the \S\ref{subsec:fragility} reporting.

\begin{table}[h]
\centering
\caption{Window-control contrast ($21$ runs, perturbation scale $0.01$). Re-labeling the same pool by RR percentile vs.\ total-parameter-norm percentile reverses the fragility ordering. Aggregator: short-horizon deviation in accuracy on a $0$--$100$ scale.}
\label{tab:e5p-app}
\small
\begin{tabular}{lccc}
\toprule
Framing & High-window deviation & Low-window deviation & Ratio (H/L) \\
\midrule
RR-window      & $11.43$ & $2.995$ & $\approx 3.82\times$ \\
Norm-window    & $4.622$ & $11.16$ & $\approx 0.41\times$ \\
\bottomrule
\end{tabular}
\end{table}

\paragraph{Interpretation.}
Norm-derived signals are strong run-level regime indicators, confirming that scale dynamics are informative for grokking. In the same-data window-control experiment, perturbation sensitivity aligns with the residual ordering rather than total-norm ordering, suggesting that RR is not merely a total-norm proxy at the window level in the studied wd$=1$ dynamics. RR therefore provides a complementary window-level fragility diagnostic on this pool. We do not claim RR universally outperforms norm baselines, and the fair-FPR comparison above is a baseline-comparison protocol distinct from the selected operating rule of \S\ref{subsec:detection}.

\section{Triggered intervention}\label{app:intervention}

We test whether RR-triggered control improves training beyond passive monitoring.

Three strategies apply a learning-rate halving event: (a) at a fixed step ($4000$); (b) at a uniformly random step within a fixed window; (c) when the RR alarm fires (\texttt{rr\_trigger}, using the \texttt{sustained\_K2\_tau10} rule). On wd$=1$ runs ($10$ runs per strategy), the \texttt{rr\_trigger} strategy prevents grokking in $0.10$ of runs; the \texttt{fixed} strategy lands an average of $773$ steps after onset (negative $=$ too late). The fixed and random strategies produce a $0$ failure rate but do not reliably accelerate grokking either; the \texttt{rr\_trigger} variant gives at most a marginal speedup at the cost of a non-zero failure rate. Monitoring does not directly imply beneficial control.

\begin{figure}[h]
  \centering
  \includegraphics[width=0.7\linewidth]{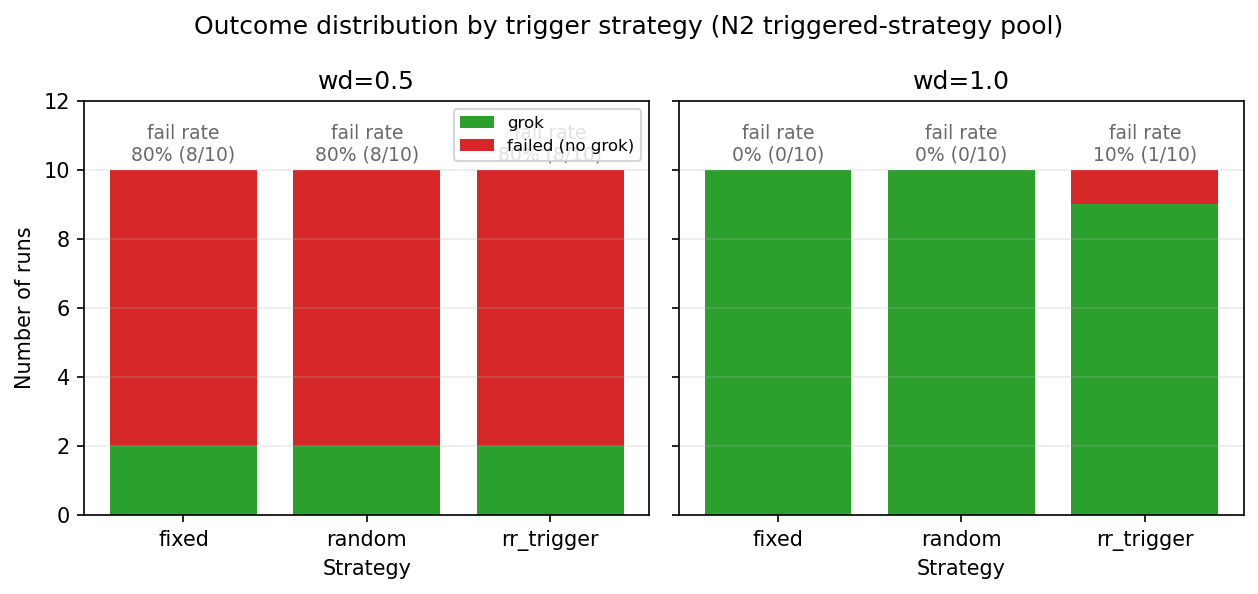}
  \caption{Outcome distribution by trigger strategy (\texttt{fixed} / \texttt{random} / \texttt{rr\_trigger}). RR-triggered intervention introduces a $10\%$ failure mode that the fixed and random strategies do not exhibit.}
  \label{fig:n2-outcomes}
\end{figure}

\section{Task-family transfer}\label{app:task-transfer}

The signal is evaluated on related modular-arithmetic settings as a transfer check.

\paragraph{Modular-addition fraction/prime sweep.}
Sweeping the training fraction and modulus prime size yields partial transfer with explicit failure cases under the same operating rule.

\paragraph{Modular multiplication.}
The same \texttt{sustained\_K2\_tau10} rule triggers on wd$=1$ modular-multiplication runs but with shorter lead than mod-addition.

We emphasize that this is preliminary task-family transfer evidence, not clean cross-task robustness, and pair every transfer claim with FPR before re-stating lead time once the relevant rows are added to the index.

\section{FCN secondary validation}\label{app:fcn-secondary}

\begin{figure*}[t]
  \centering
  \subfigure[Reconstruction residuals for FCNs]{%
    \includegraphics[width=0.45\linewidth]{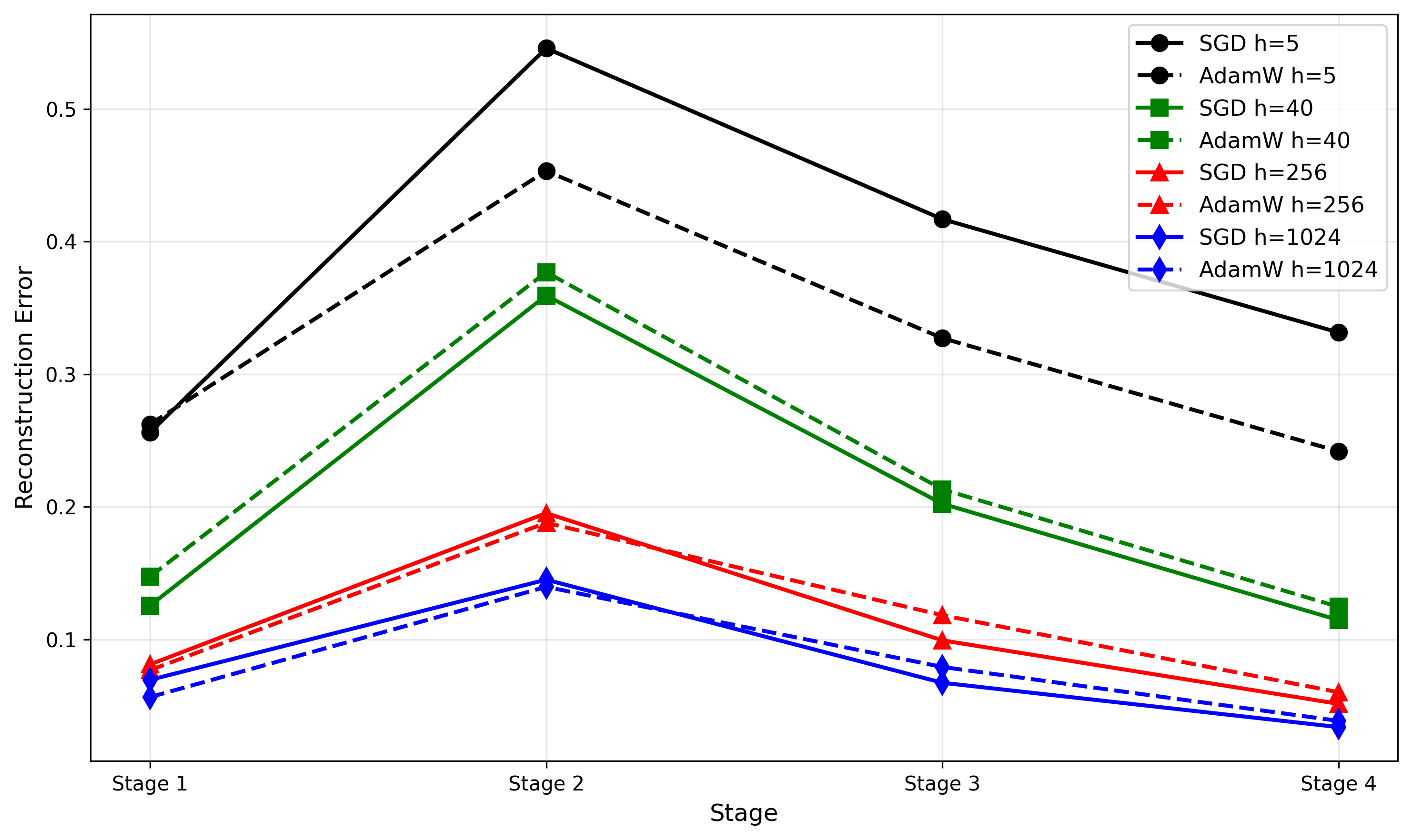}%
    \label{fig:tri:a}%
  }\hfill
  \subfigure[$99\%$ effective rank for FCNs]{%
    \includegraphics[width=0.45\linewidth]{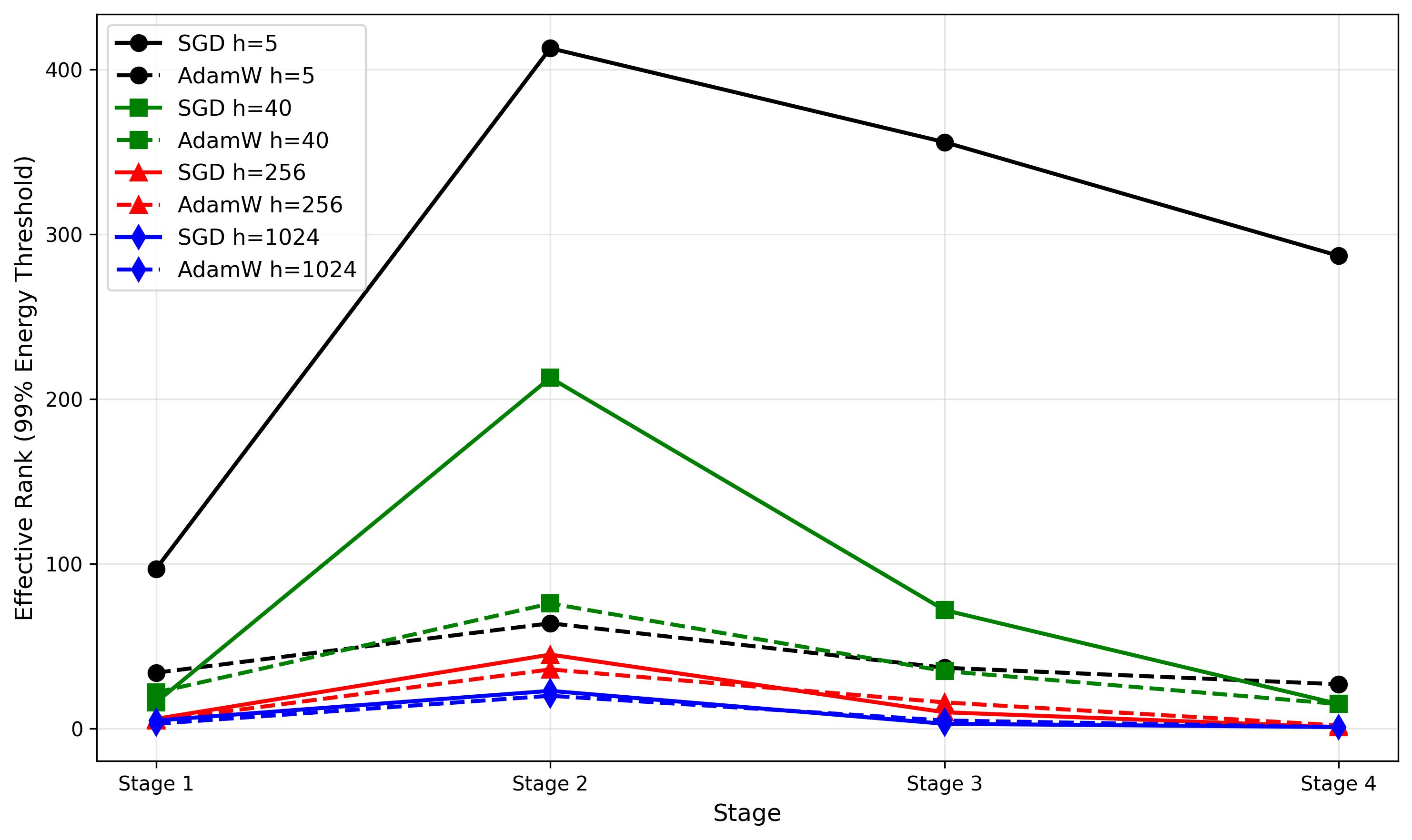}%
    \label{fig:tri:b}%
  }
  \caption{FCN secondary check: stage-wise reconstruction residuals and $99\%$ effective rank for FCNs of widths $h\in\{5,40,256,1024\}$ under SGD (solid) and AdamW (dashed). Wider networks tend to give smaller residuals and lower effective rank in the chosen observation space; AdamW tends to give lower effective rank than SGD at matched settings.}
  \label{fig:tri}
\end{figure*}

For FCNs of varying width on MNIST, we treat residual and effective rank as low-residual regime descriptors (a secondary use), not as the primary detection signal. The FCN training details (Table~\ref{tab:fcn_hparams}), randomized shuffle control, and qualitative seed/initialization/activation robustness checks are reported as part of Section~\ref{app:exp} above. Figure~\ref{fig:tri} reports the stage-wise reconstruction residual and effective rank across widths under SGD and AdamW.

The qualitative observations are: spectral structure varies with width under both SGD and AdamW; wider networks tend to exhibit smaller reconstruction residuals and lower effective rank across stages; AdamW tends to yield lower effective rank than SGD under matched settings; and the spectra are stable under small multiplicative initialization perturbations and across seeds. Among the tested factors in this FCN setting, width produces the clearest observed differences in training dynamics. We do not interpret these FCN results as a grokking diagnostic claim; they are included as a low-residual regime check and as a comparison point with prior Koopman-spectral analyses on FCN training \citep{redman2024identifyingequivalenttrainingdynamics}.

\section{Extended related work and positioning}\label{app:related}

This appendix expands the brief positioning given in the introduction.

\paragraph{Mechanistic and norm-based accounts of grokking.}
Since the original observation of delayed generalization on algorithmic tasks \citep{power2022grokkinggeneralizationoverfittingsmall}, several lines of work have sought to explain grokking. Mechanistic interpretability identifies the emergence of specific computational circuits---Fourier features for modular addition \citep{nanda2023progressmeasuresgrokkingmechanistic} and circuit-efficiency tradeoffs \citep{varma2023explaininggrokkingcircuitefficiency}---and yields progress measures that track circuit formation. Implicit-bias accounts attribute grokking to late-phase norm minimization on the zero-loss manifold \citep{liu2022understandinggrokkingeffectivetheory,lyu2024dichotomyearlylatephase,musat2026geometrygrokkingnormminimization}; stability-based accounts link it to logit scaling and softmax collapse \citep{thilak2022slingshotmechanismempiricalstudy,prieto2025grokkingedgenumericalstability}. These works primarily address why grokking occurs and yield signals tied to specific circuits or mechanisms. Our reconstruction residual addresses a different question---window-level transition localization from chosen task-dependent distributional observables---and is intended to complement rather than substitute for these mechanistic accounts.

\paragraph{Gradient-based diagnostics.}
A separate line of work tracks training through gradient-derived quantities. The Average Gradient Outer Product (AGOP) \citep{radhakrishnan2024agop} has been used to study feature emergence and grokking-related transitions, providing another route to identifying transition windows. We view our residual as complementary: it is derived from windowed distributional dynamics rather than from gradient outer products, and is intended as a window-level monitoring signal rather than a mechanism-specific progress measure. In our setup, sparse checkpoint coverage prevents a fair head-to-head comparison; AGOP appears in Appendix~\ref{app:agop} as corroborative rather than competitive evidence.

\paragraph{Spectral and geometric diagnostics of training.}
A large body of work characterizes training through curvature spectra \citep{sagun2018empiricalanalysishessianoverparametrized,papyan2019measurementsthreelevelhierarchicalstructure,ghorbani2019investigationneuralnetoptimization}, gradient confinement to top Hessian subspaces \citep{gurari2018gradientdescenthappenstiny}, and the edge-of-stability phenomenon in which the top Hessian eigenvalue saturates near $2/\eta$ \citep{cohen2022gradientdescentneuralnetworks,damian2023selfstabilizationimplicitbiasgradient}. Weight-matrix spectra have been studied via random-matrix theory and heavy-tailed self-regularization \citep{pmlr-v70-pennington17a,martin2018implicitselfregularizationdeepneural}. Trajectory-based analyses report low-dimensional structure of parameter evolution \citep{li2018measuringintrinsicdimensionobjective,aghajanyan2020intrinsicdimensionalityexplainseffectiveness,Mao_2024,li2021lowdimensionallandscapehypothesis}, correlated dynamics \citep{brokman2024enhancingneuraltrainingcorrelated}, and distinct lazy/rich regimes \citep{jacot2020neuraltangentkernelconvergence,chizat2020lazytrainingdifferentiableprogramming,Lee_2020,woodworth2020kernelrichregimesoverparametrized}, with the precise scaling determining whether feature learning occurs \citep{yang2022featurelearninginfinitewidthneural}. Our diagnostic operates on the empirical distribution of a chosen observable rather than on raw parameters or curvature, and is computed over windows rather than as instantaneous summaries; for output-distribution observables the construction does not depend on hidden-unit indexing.

\paragraph{Koopman and DMD approaches to neural training.}
Koopman-operator methods \citep{SCHMID_2010,rowley2009,H_Tu_2014,Arbabi_2017,Brunton_2017,drmač2017datadrivenmodaldecompositions} have been applied to analyze and accelerate neural training \citep{dogra2020optimizingneuralnetworkskoopman,tano2020acceleratingtrainingartificialneural,luo2024quackacceleratinggradientbasedquantum} and to detect equivalent training dynamics through Koopman conjugacy \citep{redman2024identifyingequivalenttrainingdynamics}. The most closely related of these, \citet{redman2024identifyingequivalenttrainingdynamics}, uses Koopman spectra to test global, full-trajectory equivalence between training runs. Our use of windowed Hankel-DMD \citep{2019koopmanrds} on distributional observables targets a different scale of analysis: window-level transition localization within a single trajectory, rather than equivalence testing across full trajectories.

\paragraph{Distributional view and Wasserstein geometry.}
Mean-field analyses model training as a Wasserstein gradient flow on the empirical parameter distribution \citep{chizat2018globalconvergencegradientdescent,Mei_2018,Rotskoff_2022,sirignano2019meanfieldanalysisneural,gess2023stochasticmodifiedflowsmeanfield}. These results require specific parameterizations and asymptotic regimes that do not directly apply to our finite-width, finite-step setting. We therefore use Wasserstein geometry and the tangent-space identification at a reference measure \citep{villani2009optimal} as a descriptive coordinate system for distribution-valued observables, and treat the choice of observable as task-dependent. We make no asymptotic claim, and our framework does not depend on a mean-field limit being attained.